\documentclass{article}

\usepackage[preprint]{neurips_2023}

\usepackage[utf8]{inputenc}      
\usepackage[T1]{fontenc}         
\usepackage[hidelinks]{hyperref} 
\usepackage{url}                 
\usepackage{booktabs}            
\usepackage{amsfonts}            
\usepackage{nicefrac}            
\usepackage{microtype}           
\usepackage{xcolor}              
\usepackage{natbib}
\usepackage{amsmath}
\usepackage{graphicx}
\usepackage{wrapfig}
\usepackage[page]{appendix}
\usepackage{siunitx}
\usepackage{fixme}
\usepackage{tabularx}
\usepackage{caption}
\usepackage[labelformat=simple]{subcaption}

\makeatletter
\newenvironment{captionalign*}{\ifvmode\else\hfil\null\linebreak\fi
  \minipage\textwidth\scriptsize
  \setlength{\abovedisplayskip}{0pt}%
  \setlength{\abovedisplayshortskip}{\abovedisplayskip}%
  \start@align\@ne\st@rredtrue\m@ne}%
{\endalign\endminipage\linebreak}
\makeatother

\DeclareMathOperator*{\argmax}{argmax}

\newcommand{\norm}[1]{\left\lVert#1\right\rVert}
\newcommand{\ode}{\log p_\theta^\textrm{ODE}}

\renewcommand{\citet}{\citep}

\title{Investigating the Adversarial Robustness of Density Estimation Using the Probability Flow ODE}

\author{%
  Marius Arvinte \quad Cory Cornelius \quad Jason Martin \quad Nageen Himayat \\
  Intel Labs, USA
}

\begin{document}

\maketitle

\begin{abstract}
  Beyond their impressive sampling capabilities, score-based diffusion models offer a powerful analysis tool in the form of unbiased density estimation of a query sample under the training data distribution.
  In this work, we investigate the robustness of density estimation using the probability flow (PF) neural ordinary differential equation (ODE) model against gradient-based likelihood maximization attacks and the relation to sample complexity, where the compressed size of a sample is used as a measure of its complexity.
  We introduce and evaluate six gradient-based log-likelihood maximization attacks, including a novel reverse integration attack. Our experimental evaluations on CIFAR-10 show that density estimation using the PF ODE is robust against high-complexity, high-likelihood attacks, and that in some cases adversarial samples are semantically meaningful, as expected from a robust estimator.
\end{abstract}

\section{Introduction}
Score-based diffusion models \citep{song2019generative,ho2020denoising,song2020score} are a powerful class of generative models that have been shown to excel at learning the score function of real-world, high-dimensional distributions of many signal modalities, such as images \citep{rombach2022high,saharia2022photorealistic}, audio \citep{kong2020diffwave}, video \citep{ho2022imagen}, text \citep{li2022diffusion}, and combinations thereof. Score-based diffusion models are also highly modular and can be used in numerous downstream applications, such as image editing \citep{meng2021sdedit}, inverse problem solving \citep{chung2022improving}, or generative classification \citep{chen2023robust}.

One downstream use for a score-based model is its use as a gradient oracle in the PF ODE for unbiased density estimation \citep{song2020score}. Density estimation is a central part of machine learning and has numerous downstream applications, such as out-of-distribution detection \citep{ren2019likelihood,wang2020further,schmier2023positive}, generative classification \citep{grathwohl2019your,chen2023robust}, and uncertainty estimation \citep{jalal2021robust,song2021solving,han2022card,luo2023bayesian}. We investigate the robustness of density estimation using the PF ODE without tying it to a specific application. Evaluating the standalone robustness of a model is highly relevant from a security perspective, where it is well-known that the robustness of the entire system depends on the weakest link \citep{carlini2017adversarial,carlini2019evaluating,tramer2020adaptive}, as well as from an interpretability perspective \citep{casper2022robust}, where it is desired to show that robust models produce semantically meaningful samples when the attacker attempts to increase confidence of the prediction \citep{ilyas2019adversarial}.

In this work, we investigate the outcome of end-to-end gradient-based attacks that target maximization of the PF ODE density estimate.
We leverage adversarial attacks as an explainability tool \citep{augustin2020adversarial,xu2020adversarial}: probing a learned density model against input-space attacks is useful for understanding what the model considers to be in-distribution. Our contributions are:
\begin{itemize}
    \item We investigate the relation between the estimated log-likelihood using the PF ODE and sample complexity and find a negative correlation even in the adversarial setting, and that no samples with simultaneously high-likelihood and high-complexity are found, where we use the notion of \textit{compressibility} as an empirical measure of sample complexity \citep{serra2019input}.
    \item We introduce a reverse integration attack designed for the PF ODE and experimentally show that this attack successfully finds high-likelihood and semantically meaningful samples, known to be a feature of robust discriminative models \citep{schott2018towards,ilyas2019adversarial}.
\end{itemize}

\section{Density Estimation using the Probability Flow ODE}
Let $\mathbf{x} \in \mathbb{R}^D$ denote a high-dimensional sample, and $t \in [0, 1]$ be a scalar time index. We consider the setting in \citet{song2020score}, which characterizes the forward diffusion process by an SDE with time- and sample-dependent drift and diffusion coefficients $\mathbf{f}$ and $g$, respectively:
\begin{equation}
    \label{eq:sde_formulation}
    \mathrm{d}\mathbf{x} = \mathbf{f}(\mathbf{x}, t) \mathrm{d}t + g(t) \mathrm{d}\mathbf{w},
\end{equation}
\noindent where $\mathbf{w}$ denotes a Wiener process. A central construct in our paper is the probability flow ODE \citet{maoutsa2020interacting,song2020score} shown to have the same marginal densities $p_t(\mathbf{x})$ for its solution at all time indices $t$ as the SDE in \eqref{eq:sde_formulation}:
\begin{equation}
    \label{eq:ode_formulation}
    \mathrm{d}\mathbf{x} = \left[ \mathbf{f}(\mathbf{x}, t) - \frac{1}{2} g(t)^2 \nabla_{\mathbf{x}} \log p_t(\mathbf{x)} \right] \mathrm{d}t \overset{\textnormal{def}}{=} \tilde{\mathbf{f}}(\mathbf{x}, t) \mathrm{d}t.
\end{equation}
We consider the sub-variance preserving formulation in \citet{song2020score}, which leads to specific choices for $\mathbf{f}$ and $g$. We discuss this choice in Appendix~\ref{app:choosing_sde}. Given a pre-trained score-based diffusion model $s_\theta(\mathbf{x}, t) \approx \nabla_\mathbf{x} \log p_t(\mathbf{x})$, the gradient of the flow function $\log p_t(\mathbf{x}(t))$ with respect to $t$ is given by the neural ODE change of variable as \citep{chen2018neural}:
\begin{equation}
\label{eq:neural_ode_change}
    \nabla_t \log p_{\theta, t}(\mathbf{x}(t)) = -\nabla \cdot \tilde{\mathbf{f}}_\theta(\mathbf{x}, t) = -\sum_i \frac{\partial [\tilde{\mathbf{f}}_\theta(\mathbf{x}, t)]_i}{\partial x_i},
\end{equation}
\noindent where the notation $\theta$ subscript indicates the dependence on the score-based model. Because exactly evaluating the sum in \eqref{eq:neural_ode_change} requires $\mathcal{O}(D)$ function evaluations, we use the Skilling-Hutchinson trace estimator in \citet{song2020score} with one Monte Carlo sample using a single backward pass and a Rademacher random vector $\mathbf{z} \in [-1, 1]^D$ as:
\begin{equation}
    \nabla \cdot \tilde{\mathbf{f}}_\theta(\mathbf{x}, t) \approx \mathbf{z}^T\nabla_\mathbf{x} \tilde{\mathbf{f}}_\theta(\mathbf{x}, t) \mathbf{z}.
\end{equation}
Given a test sample $\mathbf{x}_0$ at time $t=0$ serving as an initial condition for \eqref{eq:ode_formulation}, off-the-shelf numerical ODE solvers can be used to solve \eqref{eq:ode_formulation} and increment the argument of the flow function up to $t=1$, where the latent representation $\mathbf{x}_1$ of the sample is found. Additionally, integration of the scalar function in \eqref{eq:neural_ode_change} followed by basic manipulations yields the per-sample log-likelihood estimator:
\begin{alignat}{3}
\label{eq:integral_basic}
    \log p_\theta^{\textrm{ODE}}(\mathbf{x}_0)  & = \int_0^1 \nabla \cdot \tilde{\mathbf{f}}_\theta(\mathbf{x}, t) \ \mathrm{d}t + \log p_1(\texttt{solver}(\mathbf{x}_0)) &=  I_\theta^{\textnormal{fw}}(\mathbf{x}_0) + P_\theta^{\textnormal{fw}}(\mathbf{x}_0), \\
\label{eq:integral_reverse}
    \log p_\theta^{\textrm{ODE}}(\texttt{solver}(\mathbf{x}_1)) & = -\int_1^0 \nabla \cdot \tilde{\mathbf{f}}_\theta(\mathbf{x}, t) \ \mathrm{d}t + \log p_1(\mathbf{x}_1) & = I_\theta^{\textnormal{rev}}(\mathbf{x}_1) + P_\theta^{\textnormal{rev}}(\mathbf{x}_1).
\end{alignat}
Modern ODE solvers can be used to simultaneously obtain the value of the integral in \eqref{eq:integral_basic} and yield $\mathbf{x}_1$ as the solution to \eqref{eq:ode_formulation} when the test sample $\mathbf{x}_0$ is the initial condition. The log-likelihood of the latent representation can be evaluated in closed-form. Both terms in the right-hand side of \eqref{eq:integral_basic} can be viewed as functions of the input $\mathbf{x}_0$, and are denoted by $I^{\textnormal{fw}}_\theta(\mathbf{x}_0)$ and $P^{\textnormal{fw}}_\theta(\mathbf{x}_0)$, respectively. The solver can also be used in reverse-time to recover the original sample $\mathbf{x}_0$ and $\log p_\theta^\textrm{ODE} (\mathbf{x}_0)$ from a latent representation, yielding $I_{\theta}^{\textnormal{rev}} (\mathbf{x}_1)$ and $P_{\theta}^{\textnormal{rev}} (\mathbf{x}_1)$ that solely depend on $\mathbf{x}_1$ as in \eqref{eq:integral_reverse}.

\vspace{-2mm}
\paragraph{Fast vs. accurate solvers} When running end-to-end gradient-based optimization, we use a \textit{single} and \textit{fixed} $\mathbf{z}$ vector shared across a relatively low number of solver iterations (e.g., $21$ steps in a fixed grid Runge-Kutta $4$th order solver), giving a lightweight, deterministic, and differentiable implementation. When measuring the outcome of gradient-based optimization, we \textit{always} use a much more expensive solver routine (e.g., $1001$ steps in the same solver). We denote the fast solver by $\log \tilde{p}_\theta^\textrm{ODE}$, and as the sequel shows, it is a successful approximation of the accurate solver.

\section{Adversarial Attacks on \texorpdfstring{$\boldmath\ode$}{PF ODE}}
Our motivation is to provide an \textit{adversarial} verification of the hypothesis that deep density estimation models tend to assign higher densities to \textit{less complex} samples, and that high-likelihood, high-complexity samples do not exist. This has been first observed for auto-regressive and normalizing flows in \citep{serra2019input}, but not in an adversarial setting -- this does not preclude the existence of such samples.
In the context of images, we use the quantitative approximation of image complexity from \citet{serra2019input} defined as the size of the image output when losslessly compressed using \texttt{PNG} defined as:
\begin{equation}
    \mathcal{C}(\mathbf{x}) = \texttt{size}(\texttt{PNG}(\mathbf{x})) / D.
\end{equation}

The idea behind our attacks is to find an additive perturbation $\delta$ applied to an initial sample $\mathbf{x}$, and constrained to lie in a set $\mathcal{S}$, such that it approximately solves:
\begin{equation}
    \argmax_{\delta} \enskip \ode (\mathbf{x} + \delta) \quad \text{s.t.} \enskip \delta \in \mathcal{S}.
\end{equation}

\begin{figure}
  \centering
  \begin{subfigure}[t]{0.137\textwidth}
    \centering
    \caption{Benign}%
    \label{fig:end2end_attacks_test_sample}
    \includegraphics[width=\textwidth]{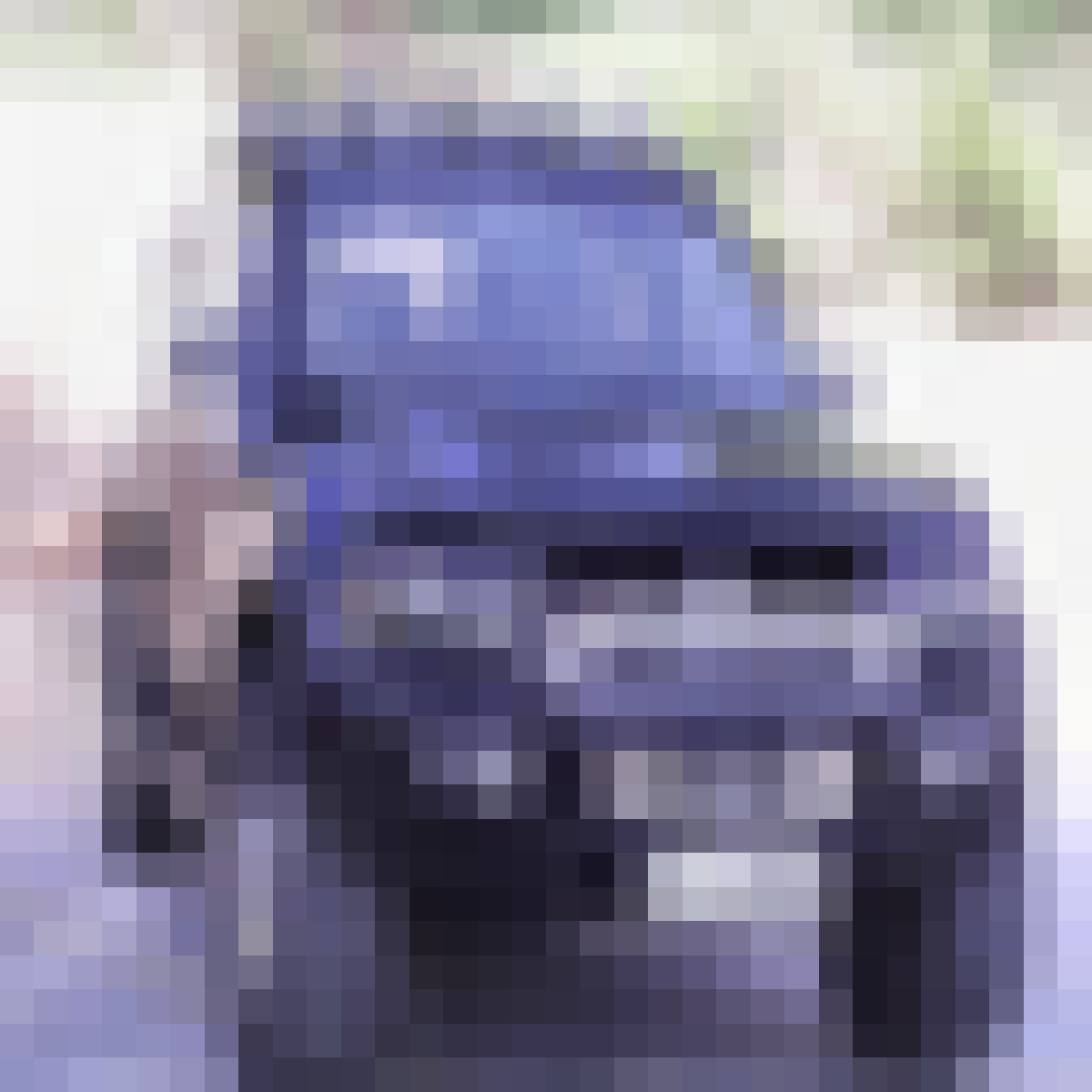}%
    \begin{captionalign*}
      \log \tilde{p}_\theta &\textnormal{ = } \num{2.38} \\
      \log p_\theta &\textnormal{ = } \num{2.33} \\
      \mathcal{C} &\textnormal{ = } \num{0.802}
    \end{captionalign*}
  \end{subfigure}
  \hfill
  \begin{subfigure}[t]{0.137\textwidth}
    \centering
    \caption{Near Sample}%
    \label{fig:end2end_attacks_near_sample}
    \includegraphics[width=\textwidth]{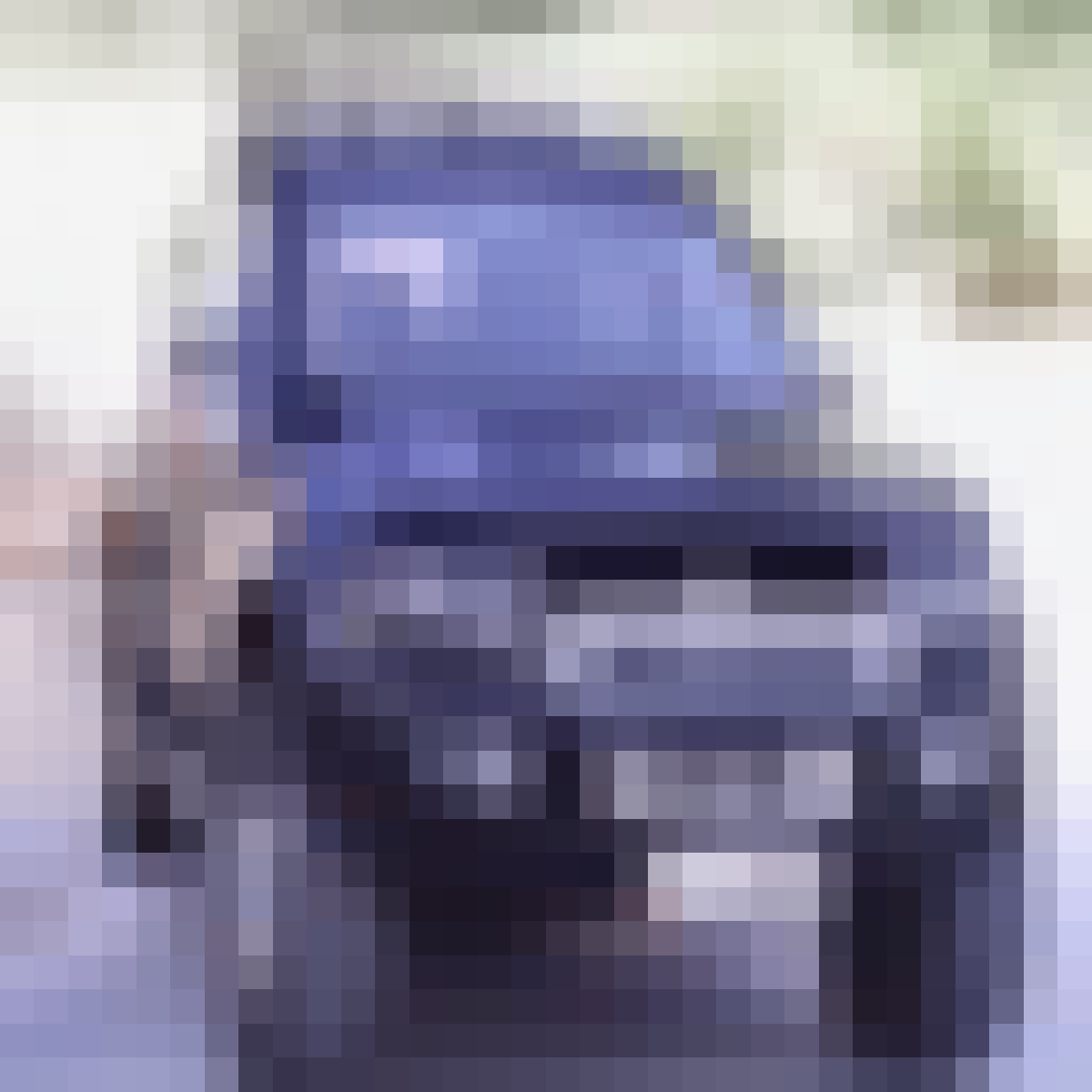}%
    \begin{captionalign*}
      \log \tilde{p}_\theta &\textnormal{ = } \num{3.77} \\
      \log p_\theta &\textnormal{ = } \num{3.49} \\
      \mathcal{C} &\textnormal{ = } \num{0.783}
    \end{captionalign*}
  \end{subfigure}
  \hfill
  \begin{subfigure}[t]{0.137\textwidth}
    \centering
    \caption{High Complex.}%
    \label{fig:end2end_attacks_dct}
    \includegraphics[width=\textwidth]{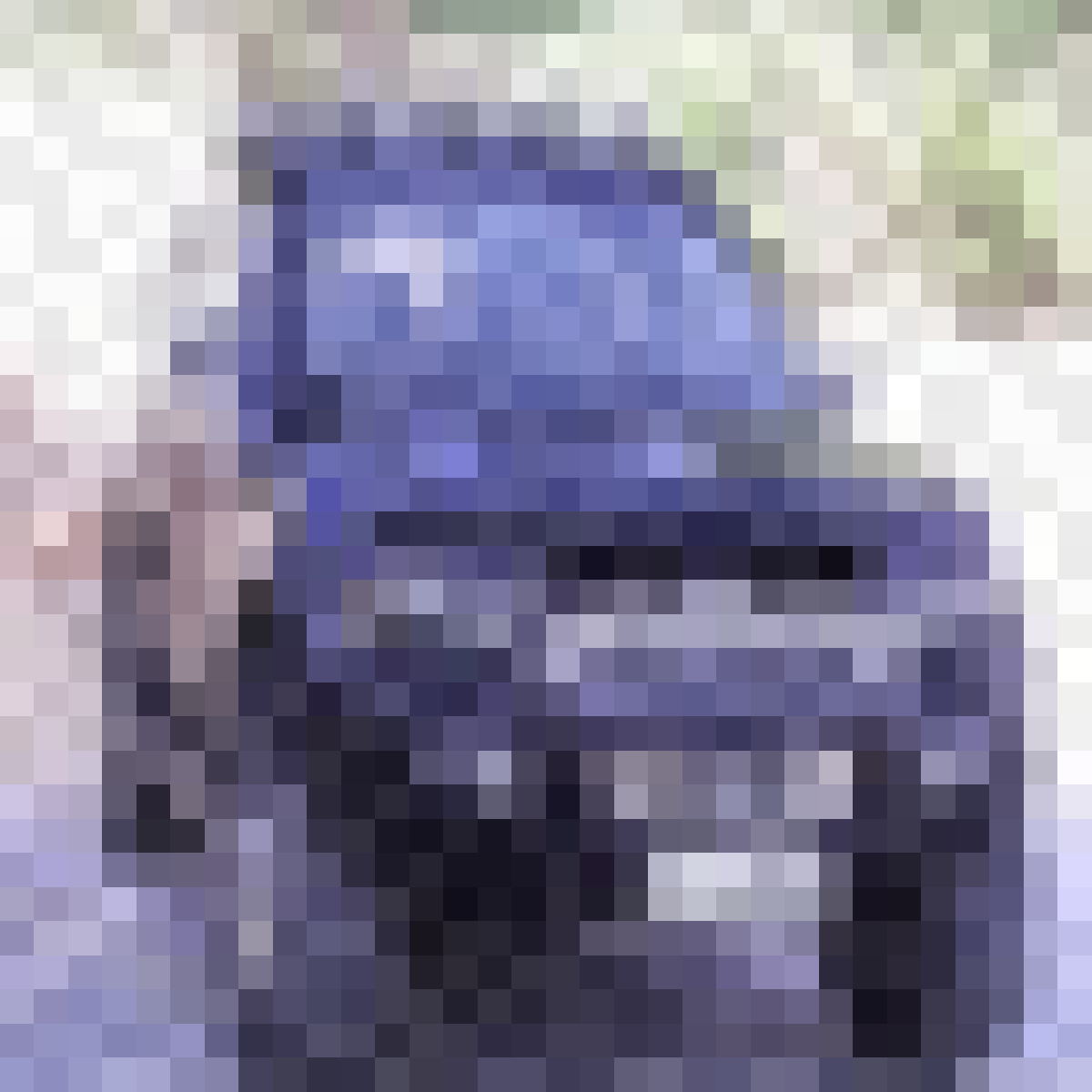}%
    \begin{captionalign*}
      \log \tilde{p}_\theta &\textnormal{ = } \num{2.41} \\
      \log p_\theta &\textnormal{ = } \num{2.18} \\
      \mathcal{C} &\textnormal{ = } \num{0.876}
    \end{captionalign*}
  \end{subfigure}
  \hfill
  \begin{subfigure}[t]{0.137\textwidth}
    \centering
    \caption{Rev. Integration}%
    \label{fig:end2end_attacks_reverse}
    \includegraphics[width=\textwidth]{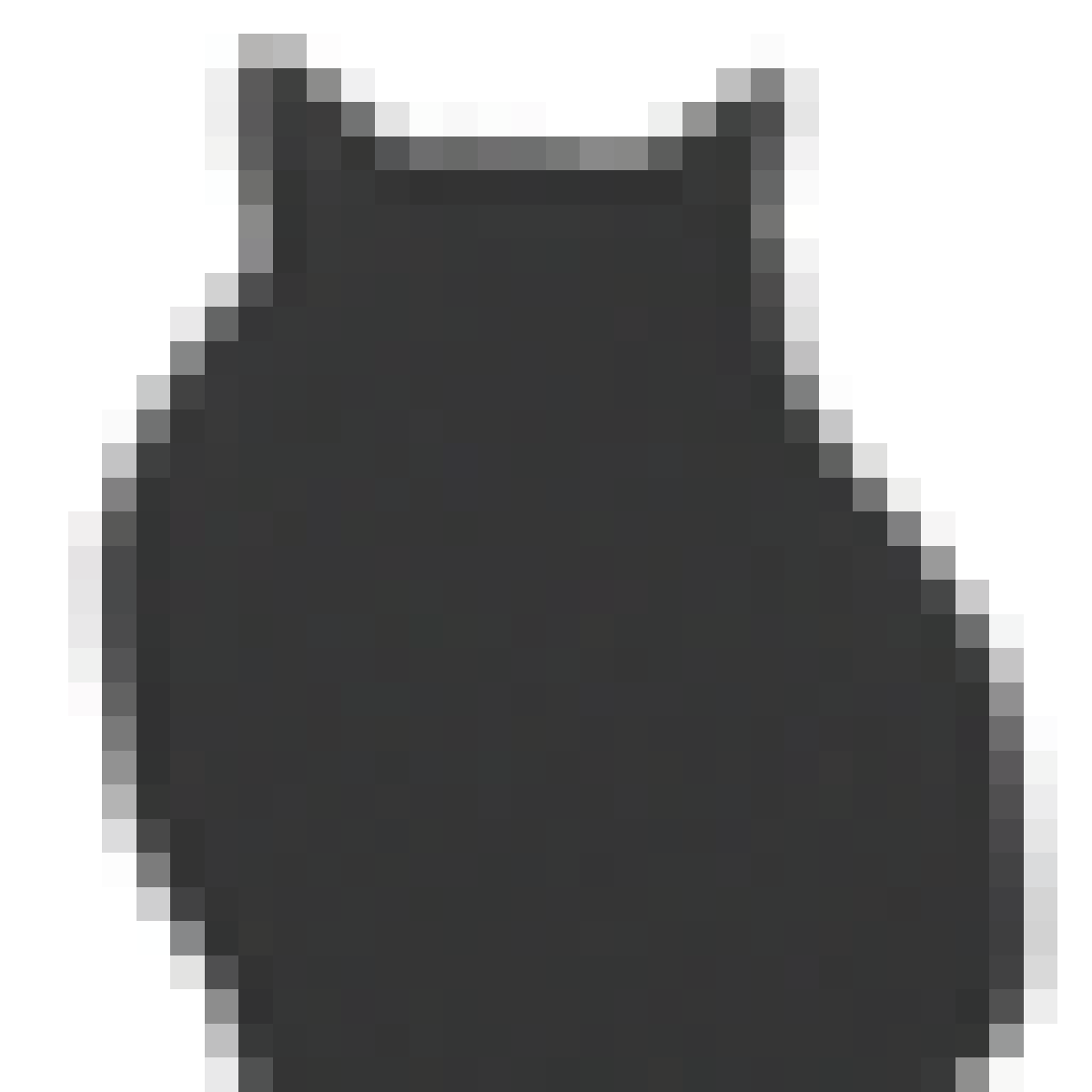}%
    \begin{captionalign*}
      \log \tilde{p}_\theta &\textnormal{ = } \num{4.67}  \\
      \log p_\theta &\textnormal{ = } \num{6.69}  \\
      \mathcal{C} &\textnormal{ = } \num{0.400}
    \end{captionalign*}
  \end{subfigure}
  \hfill
  \begin{subfigure}[t]{0.137\textwidth}
    \centering
    \caption{Random Reg.}%
    \label{fig:end2end_attacks_random_region}
    \includegraphics[width=\textwidth]{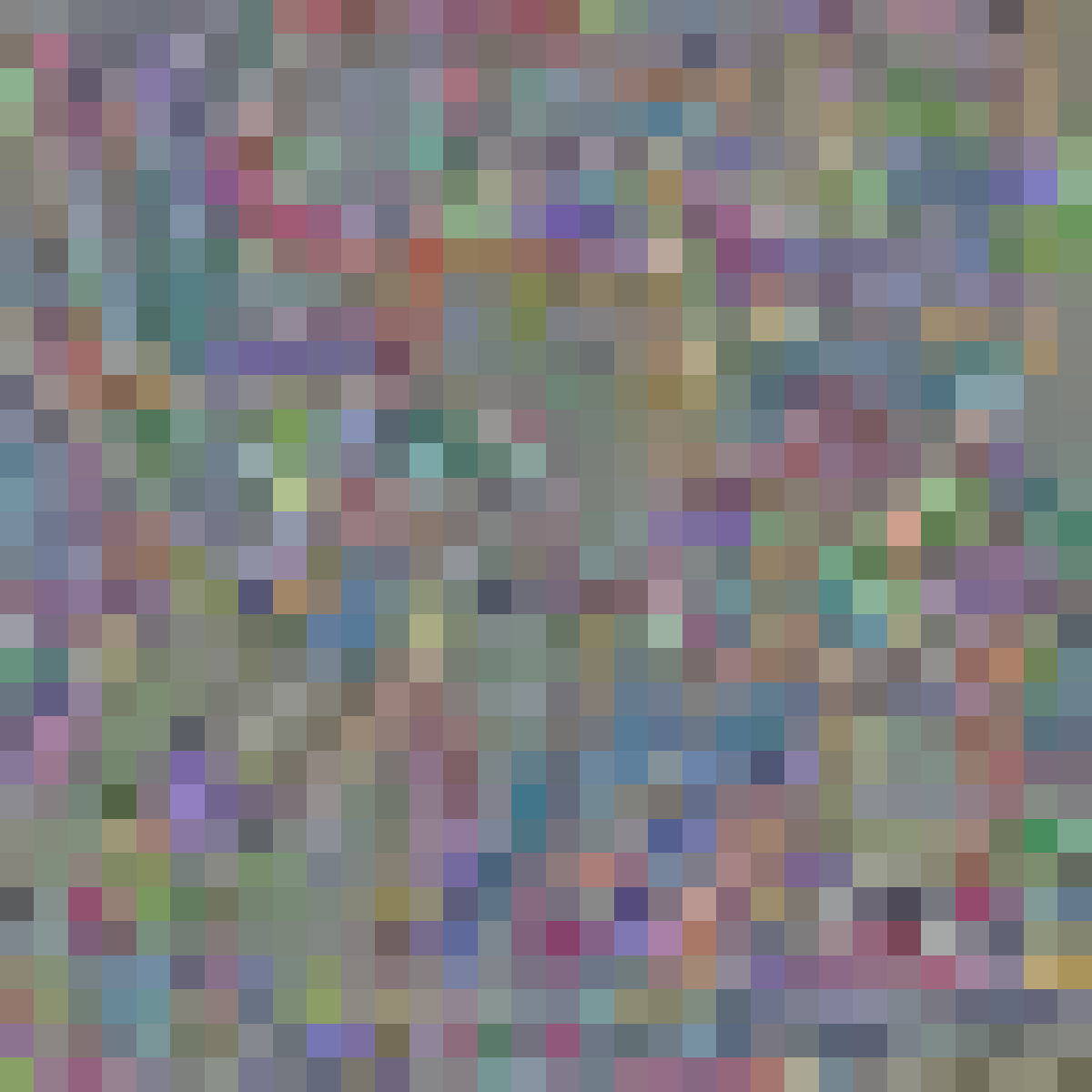}%
    \begin{captionalign*}
      \log \tilde{p}_\theta &\textnormal{ = } \num{1.37}  \\
      \log p_\theta &\textnormal{ = } \num{1.27}  \\
      \mathcal{C} &\textnormal{ = } \num{0.831}
    \end{captionalign*}
  \end{subfigure}
  \hfill
  \begin{subfigure}[t]{0.137\textwidth}
    \centering
    \caption{Prior-Only}%
    \label{fig:end2end_attacks_prior}
    \includegraphics[width=\textwidth]{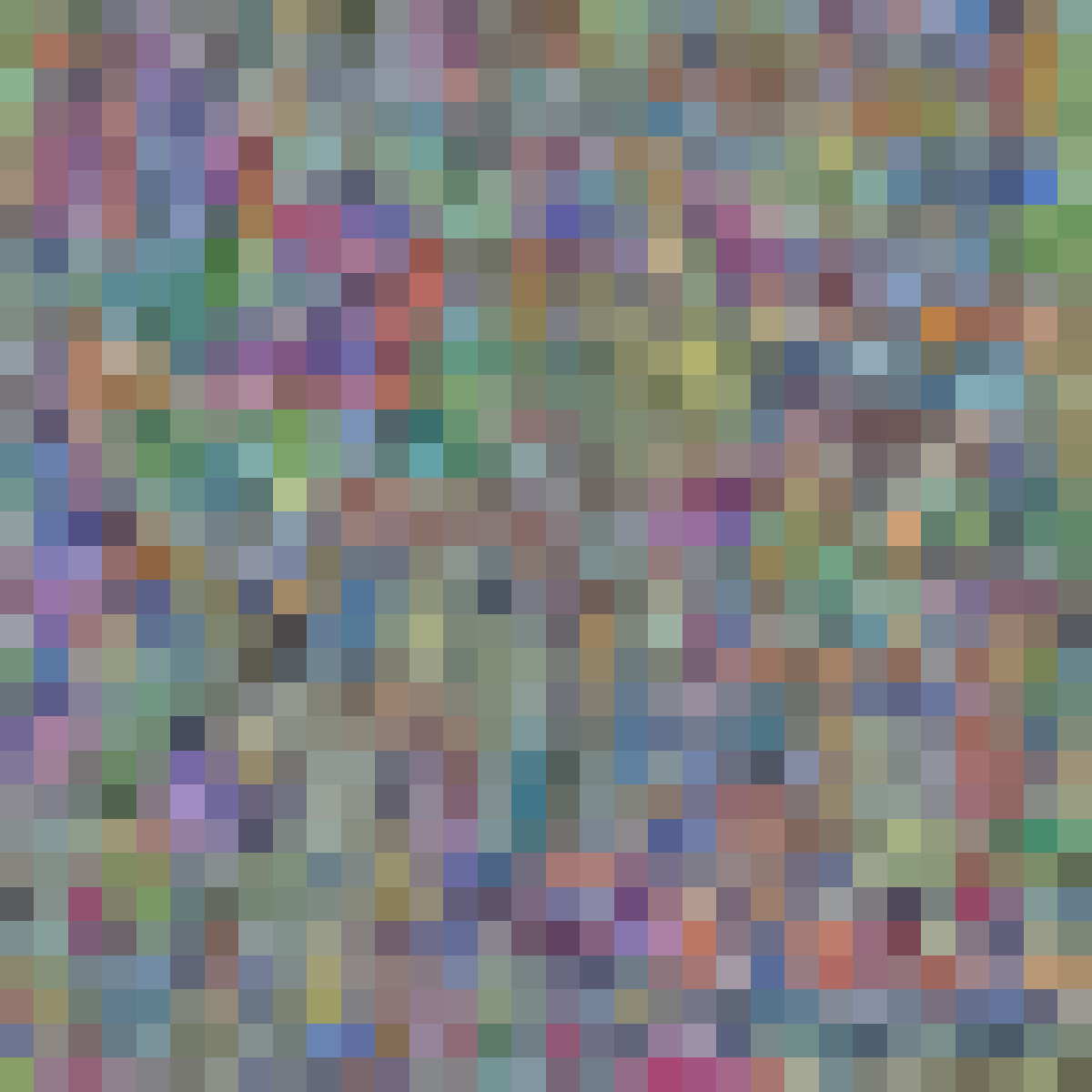}%
    \begin{captionalign*}
      \log \tilde{p}_\theta &\textnormal{ = } \num{0.91} \\
      \log p_\theta &\textnormal{ = } \num{0.92} \\
      \mathcal{C} &\textnormal{ = } \num{0.858}
    \end{captionalign*}
  \end{subfigure}
  \hfill
  \begin{subfigure}[t]{0.137\textwidth}
    \centering
    \caption{Unrestricted}%
    \label{fig:end2end_attacks_unrestricted}
    \includegraphics[width=\textwidth]{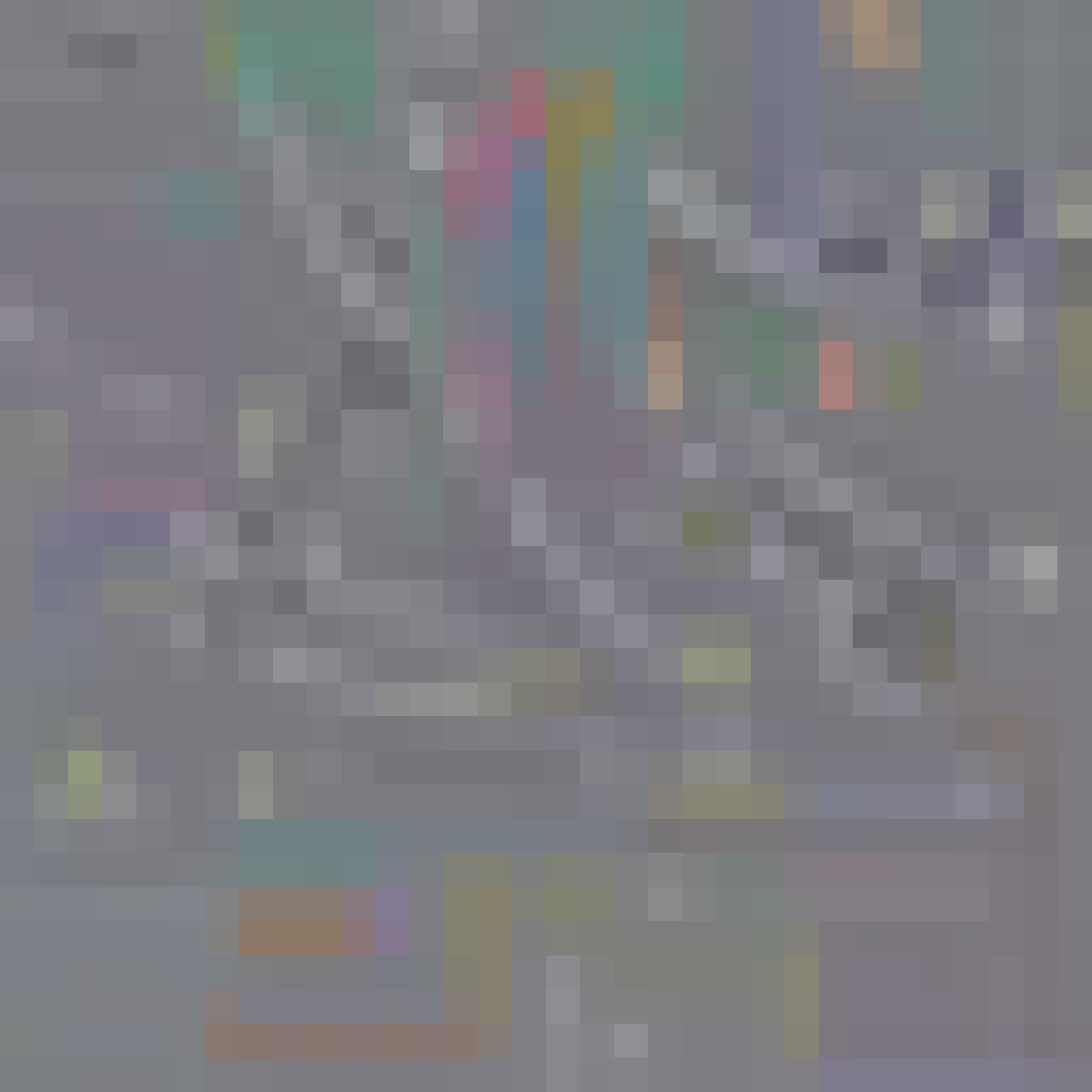}%
    \begin{captionalign*}
      \log \tilde{p}_\theta &\textnormal{ = } \num{4.88} \\
      \log p_\theta &\textnormal{ = } \num{3.72} \\
      \mathcal{C} &\textnormal{ = } \num{0.603}
    \end{captionalign*}
  \end{subfigure}
  \caption{Sample images with fast $\log \tilde{p}_\theta$ and accurate $\log p_\theta$ estimates of $\ode$ and complexity $\mathcal{C}$.
  We measure \subref{fig:end2end_attacks_test_sample}~CIFAR-10 test image and \subref{fig:end2end_attacks_near_sample}-\subref{fig:end2end_attacks_unrestricted} end-to-end attacks on $\ode$. We compute attacks \subref{fig:end2end_attacks_near_sample}~bounded to be near the benign sample, \subref{fig:end2end_attacks_dct}~targeting high complexity, \subref{fig:end2end_attacks_reverse}~via reverse integration, \subref{fig:end2end_attacks_random_region}~on random regions of input space, \subref{fig:end2end_attacks_prior}~using a prior-only objective, and \subref{fig:end2end_attacks_unrestricted}~with no restrictions.}%
  \label{fig:end2end_attacks}
\end{figure}

\begin{wrapfigure}{R}{7.0cm}
  \centering
  \vspace{-5pt}
  \includegraphics[width=\linewidth]{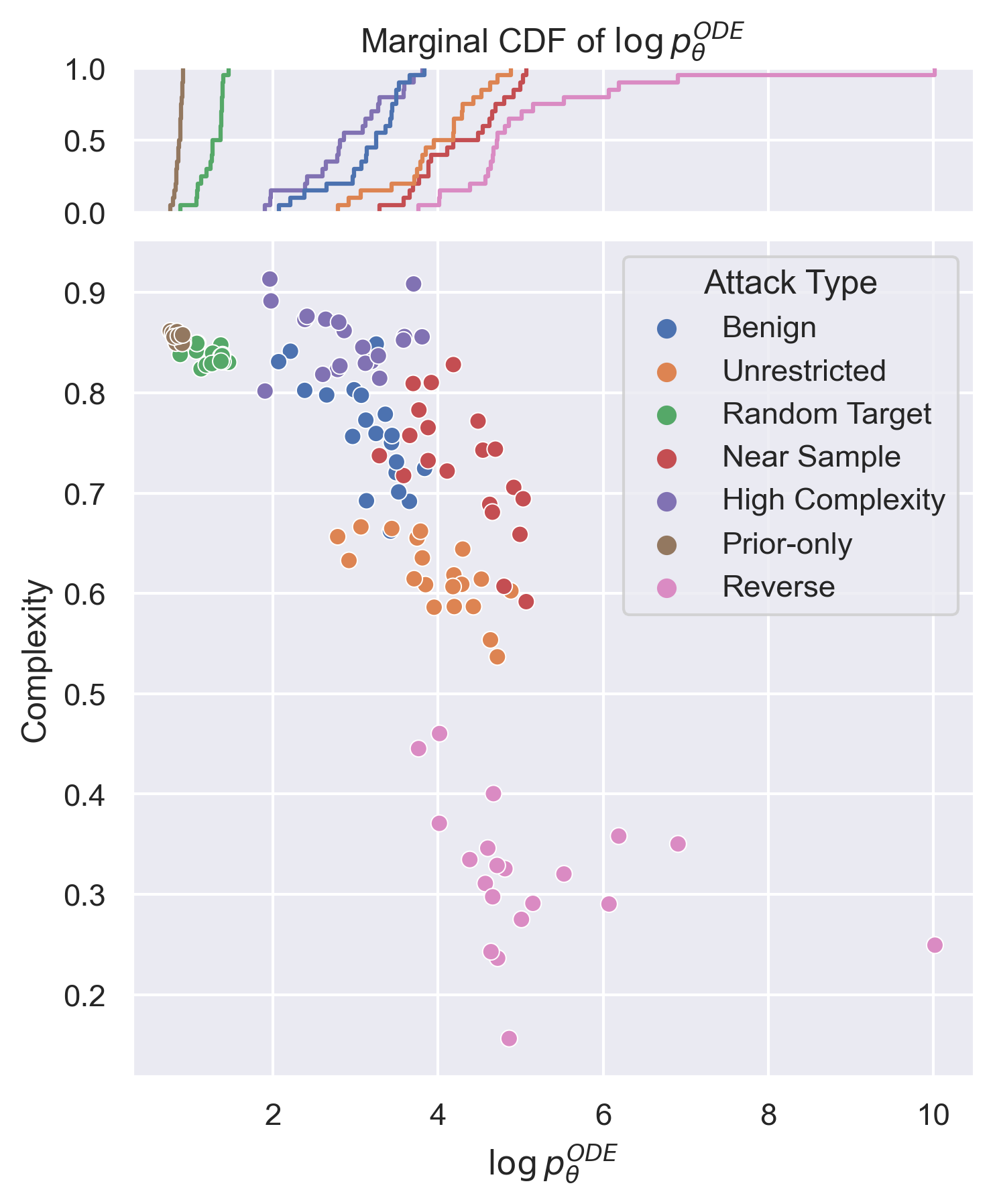}
  \caption{$\mathcal{C}$ as a function of $\ode$ for the benign data and the considered attacks. The top plot shows the marginal cumulative distribution function for $\ode$ on the benign data and each attack type.}
  \vspace{-20pt}
  \label{fig:scatter_plot}
\end{wrapfigure}

The outcome of the attacks applied to a set of $20$ samples are shown in Figure~\ref{fig:scatter_plot}, where it can be noticed that no attack was able to find samples that simultaneously have high values of $\ode$ and high complexity. Unless specified, we report the raw value of $\ode / D$ using the accurate solver.

\vspace{-3mm}
\paragraph{Unrestricted attacks}
The adversary optimizes $\delta$ in image space, and is only restricted by the global pixel bounds. The attack uses $\mathbf{x} = \mathbf{0}$ and $\mathcal{S} = [-1, 1]^D$, and clips after every gradient step to enforce the constraint. A set of outcomes is shown in Figure~\ref{fig:end2end_attacks_unrestricted}. Unrestricted adversaries find monochrome images with higher likelihoods than the CIFAR-10 test set, and lower complexity. This has been previously observed for normalizing flows and auto-regressive models \citet{serra2019input}, and shows that this bias is also present for $\ode$. This result also indicates that there is a strong mode of the estimated density around the all-zeroes sample, with additional results in Appendix~\ref{app:more_examples}.

\vspace{-3mm}
\paragraph{Random region attacks}
The adversary is restricted to optimizing $\delta$ in a random region of image space bounded by an $\ell_\infty$-ball of radius $\epsilon$. The attack uses a random, but fixed $\mathbf{x} \sim \mathcal{N}(0, 0.2)$ and $\mathcal{S} = [-\epsilon, \epsilon]^D$ with $\epsilon = 0.16$, and clips after every gradient step to enforce the constraint. We chose $\mathbf{x}$ and $\epsilon$ such that clipping occurs infrequently. Results are shown in Figure~\ref{fig:end2end_attacks_random_region}, where it can be seen that this attack cannot find any sample with values of $\ode$ above the average benign test set value. The best the attacker can do is to reduce image complexity as much as possible, indicating a negative correlation between $\ode$ and $\mathcal{C}$.

\vspace{-3mm}
\paragraph{Near-sample attacks}
This attack is similar to the random region attack. The perturbation $\delta$ is added to an in-distribution test sample. We use a random $\mathbf{x}$ from the CIFAR-10 test set and $\mathcal{S} = [-\epsilon, \epsilon]^D$ with $\epsilon = 0.06$. Results are shown in Figure~\ref{fig:end2end_attacks_near_sample}, where it can be noticed that the attacks successfully increase $\ode$ by reducing $\mathcal{C}$ and making the images appeared qualitatively smoother.

\vspace{-3mm}
\paragraph{High-complexity attacks}
The adversary uses the high frequencies in the Discrete Cosine Transform of $\mathbf{x}$ as a differentiable approximation to $\mathcal{C}$, and maximizes the regularized objective $\ode (x + \delta) + \lambda \cdot \norm{\texttt{DCT}(\mathbf{x})_{W/2:W, H/2:H}}_2^2$, where $W$ and $H$ are the image dimensions. We use the same setting as the near-sample attacks, and show results in Figure~\ref{fig:end2end_attacks_dct} for $\lambda = 1$. We notice that our term is a successful proxy for image complexity, and that there is always a trade-off between inducing high frequencies of the perturbation and $\ode$, with the attacks failing to increase $\ode$ for most samples.

\vspace{-3mm}
\paragraph{Random region prior-only attacks}
This formulation uses only the prior-related term $P_\theta^\textnormal{fw}(\mathbf{x})$ in the maximization objective. Figure~\ref{fig:end2end_attacks_prior} shows the outcome in the same setting as the random region attack. Surprisingly, we find that these attacks are as efficient as optimizing the complete expression of $\ode$ itself, and have more stable convergence than when $I_\theta^\textnormal{fw}(\mathbf{x})$ is included. This same attack could be applied in all other settings, and Appendix~\ref{app:attack_conv} discusses its effects further.

\vspace{-3mm}
\paragraph{Reverse integration attacks}
The attack optimizes the latent code $\mathbf{x}_1$ of a sample $\mathbf{x}_0$ using \textit{reverse} integration of the PF ODE via the objective $\argmax_\delta I_{\theta}^{\textnormal{rev}} (\mathbf{x}_1 + \delta) + P_{\theta}^{\textnormal{rev}} (\mathbf{x}_1 + \delta)$ and $\mathcal{S} = [-\epsilon, \epsilon]^D$. While the identities $I_\theta^{\textnormal{fw}}(\mathbf{x}_0) = I_{\theta}^{\textnormal{rev}} (\mathbf{x}_1)$ and $P_\theta^\textnormal{fw}(\mathbf{x}_0) = P_{\theta}^{\textnormal{rev}} (\mathbf{x}_1)$ should hold when $\mathbf{x}_0$ and $\mathbf{x}_1$ are paired in the PF ODE, they may not hold in practice. This choice may affect the gradients of these functions with respect to their inputs, ultimately affecting the convergence of a gradient-based adversary. These attacks are bounded in latent space to implicitly preserve high values of $P_{\theta}^{\textnormal{rev}}$, but unrestricted in $L_\infty$-norm in sample space.

Results for $\epsilon = 0.16$ are shown in Figure~\ref{fig:end2end_attacks_reverse} in the form of samples decoded from the resulting latent codes. The resulting samples have higher likelihoods than the unrestricted attack and were found without the need of learning rate tuning. Further, the optimized samples preserve semantic meaning in some cases, e.g., the vehicle is transformed into a cat silhouette with a large value of $\ode$, an effect similar to the interpretability of adversarial samples designed for robust discriminative models \citep{ilyas2019adversarial}.

\section{Conclusion}
Our findings have shown that the estimator $\ode$ is robust against high-likelihood, high-complexity attacks and is generally inversely correlated with the estimated complexity of a sample. Using this observation, a corrected density estimate formed by scaling and subtracting the complexity can be formed as per \citep{serra2019input}, allowing $\ode$ to be safely used in downstream applications, such as adversarially robust out-of-distribution detection. We found that gradient-based attacks transfer from fast to accurate ODE solvers, and that certain types of well-crafted white-box attacks are successful at producing semantically meaningful samples. This is not surprising in theory, due to the adversary essentially performing gradient ascent on the \textit{approximate} log-density of the training distribution, but is extremely relevant in practice for qualitatively measuring how well this density is learned.

\vspace{-2mm}
\subsection{Related Work}
Several recent works have used forward and reverse integration of the SDE in \eqref{eq:sde_formulation} as adversarial pre-processing defenses for downstream discriminative models \citep{yoon2021adversarial,nie2022diffusion,carlini2022certified}. While these methods are robust against off-the-shelf attackers, other works have discussed the limitations and challenges of correctly evaluating stochastic defenses \citep{gao2022limitations,lee2023robust}, as well as adversarially evaluating numerical ODE solvers \citep{huang2022adversarial}. The attacks we introduced deal with these challenges via a fully deterministic model and exact gradients. 

The recent work in \citep{chen2023robust} has used the upper bound for $\ode$ derived from the SDE formulation in a robust generative classification framework, and demonstrated robustness against adaptive attacks on SDE solvers. On a different line, the recent works in \citep{carlini2022certified,zhang2023diffsmooth} have derived certified robustness guarantees for discriminative models by leveraging single or multi-step sampling from the SDE.

Adversarial verification of other neural ODE models, such as flow matching \citep{lipman2022flow} or maximum likelihood trained score-based diffusion models \citep{song2021maximum} is a promising area of future research, as is formally connecting any of the training objectives with those in robust discriminative learning, such as the well-known adversarial training \citep{goodfellow2014explaining,madry2017towards}.

\section*{Acknowledgements}
This publication includes work that was funded by the Government under the Defense Advanced Research Projects Agency (DARPA) Guaranteeing AI Robustness against Deception (GARD) Program, Agreement \#HR00112030001.

\bibliography{references}


\begin{appendices}
    
\section{Implementation Details}
\label{app:implementation}
All our experiments have used a pre-trained CIFAR-10 score-based diffusion model from \citep{song2020score}, as well as building on top of their open-sourced library\footnote{\href{https://github.com/yang-song/score_sde_pytorch}{https://github.com/yang-song/score\_sde\_pytorch}}. More concretely, we used the subVP-SDE checkpoint file \texttt{checkpoint\_19.pth} with the model configuration \texttt{cifar10\_ddpmpp\_deep\_continuous}. Publicly available source code for reproducing our experiments can be found at \texttt{anonymous}.

\subsection{Choosing the subVP-SDE Model}
\label{app:choosing_sde}
Compared to its VP- and VE- counterparts, the subVP-SDE formulation for the forward diffusion process is a lesser popular choice. However, as the work in \citep{song2020score} has shown, this is the configuration that produces the highest average test set likelihoods and is shown to perform best at estimating likelihoods even without maximum likelihood training \citep{song2021maximum}.

When density estimation is desired, the VE-SDE process can be excluded because of not having a closed-form prior at $T=1$ under which $\log p_1 (\mathbf{x}_1)$ can be evaluated, and ultimately leads to lower test set likelihoods. This leaves us with a choice between the VP- and subVP-SDE formulations. To validate the claim in \citep{song2020score}, we evaluated $\ode$ using checkpoints from both models and found that the fast likelihood estimate is much more accurate for the subVP-SDE, hence decided to use this model in all experiments. The subVP-SDE formulation uses the functions \citep{song2020score}:
\begin{align}
    \mathbf{f}(\mathbf{x}, t) & = -\frac{1}{2}\beta(t) \mathbf{x}, \\
    g(t) & = \sqrt{\beta(t) \left( 1 - e^{-2\int^t_0 \beta(s) ds} \right)},
\end{align}
\noindent where $\beta(t) = (1 - t) \beta_0 + t \beta_1$ and $\beta_0 = 0.1$, $\beta_1 = 20$.

\subsection{ODE Solver Parameters}
All our experiments use a differentiable PyTorch implementation of a fourth order fixed Runge-Kutta ODE solver (RK4), using the publicly available \texttt{torchdiffeq} library \citep{torchdiffeq}.

For the fast solver that is being back-propagated throughout we used a step size of $0.05$, leading to $21$ unrolled steps, and an initial time of $t_0 = 0.00001$ to avoid numerical issues at $t_0=0$. The accurate solver for which we always report the value of $\ode$ for uses the same solver and starting time, but a step size of $0.001$, leading to $1001$ steps. Additionally, the accurate solver uses a different random $\mathbf{z}$ at each solver step and across samples, to rule out the possibility of attacks overfitting to the fixed $\mathbf{z}$ used with the fast solver.

\begin{figure}[t]
  \centering
  \includegraphics[width=0.6\linewidth]{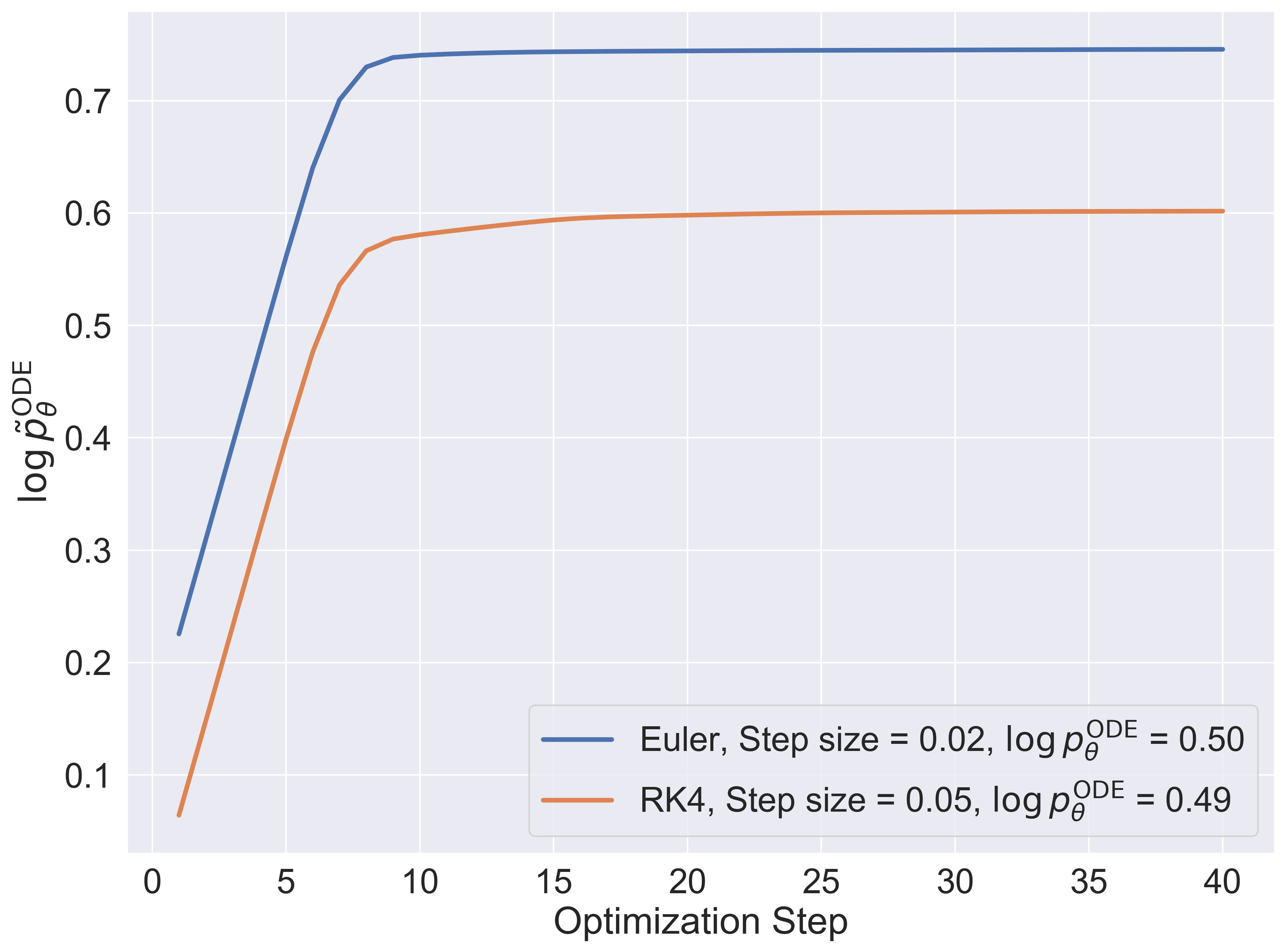}
  \caption{Convergence of gradient-based attacks using the RK4 and Euler solvers with different step sizes, but same overall complexity, respectively.}
  \label{fig:rk4_vs_euler}
\end{figure}

We chose the RK4 solver with fewer steps as opposed to a lower order solver with more steps to trade-off some computational latency for memory, as well as to avoid potential issues caused by vanishing gradients when back-propagating through a large number of sequential function calls.

Figure~\ref{fig:rk4_vs_euler} shows the results of end-to-end likelihood maximization through two different solvers: RK4 with step size of $0.05$ and the Euler solver with step size of $0.02$, having the same overall latency. The optimization targets the prior-only attack using the Adam optimizer \citep{kingma2014adam} with a learning rate of $0.01$ and default PyTorch parameters. While both attacks converge quickly, the Euler solver over-estimates the true $\ode$ twice as much as the RK4 solver, indicating its poorer quality as an approximation when used in end-to-end gradient-based attacks.

\section{Attack hyper-parameters and convergence}
\label{app:attack_conv}
Unless otherwise specified, all attacks are implemented using the Adam optimizer with a learning rate of $0.003$, and ran for a maximum number of $500$ steps. For all attacks, we have used the \textit{exact} gradients when back-propagating through the ODE solver. Running one iteration of the fixed grid RK4 step solver with a step size of $0.05$ takes approximately $34$ seconds and $48$ GB of memory on a single NVIDIA RTX A6000 card.

\begin{figure}[t]
  \centering
  \includegraphics[width=0.6\linewidth]{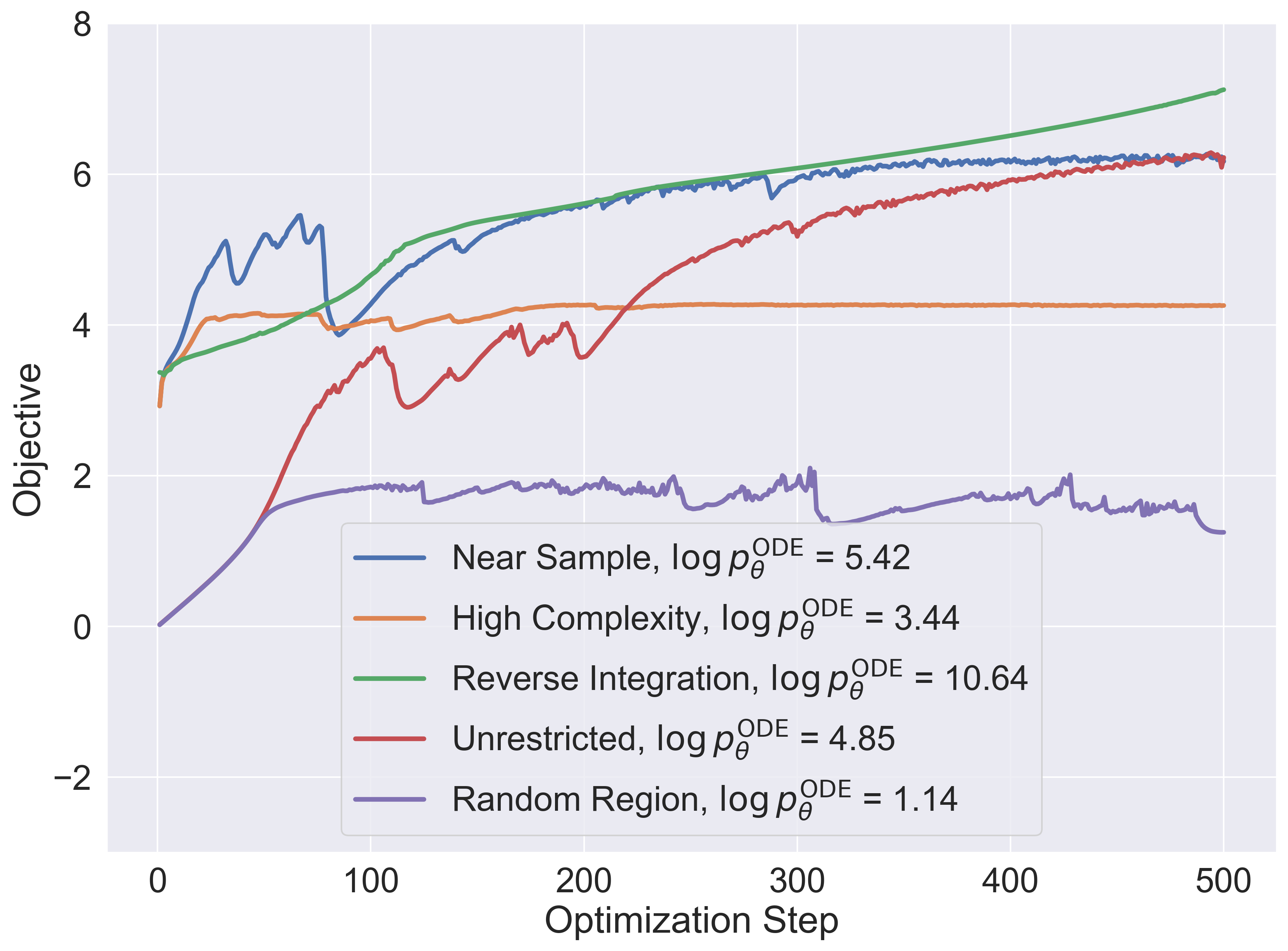}
  \caption{Convergence of gradient-based attacks on a single run. The near sample, high complexity and reverse integration attacks were all initialized using the same benign sample.}
  \label{fig:attack_convergence}
\end{figure}

Figure~\ref{fig:attack_convergence} shows the convergence of the optimization objective for all attacks except the prior-only attack:
\begin{itemize}
    \item The high complexity attack is generally the most stable and fastest to converge, indicative of the strict trade-off that has to be made between $\mathcal{C}$ and $\ode$. 
    \item The near sample is unstable in the beginning but ultimately converges after $400$ steps. 
    \item The reverse integration and unrestricted attacks stabilize but do not fully converge. 
    \item The random region attack is the most unstable, and in some cases diverges after a number of optimization steps. This is a sign of gradient masking -- intentional or not -- present in random regions of space.
\end{itemize}

Overall, Figure~\ref{fig:attack_convergence} shows some of the challenges of properly tuning the considered end-to-end optimization: even with a relatively low learning rate of $0.003$, some attacks are not stable, and additionally do not completely converge to their asymptotic values of $\log \tilde{p}_\theta^{\textrm{ODE}}$.

\begin{figure}[t]
  \centering
  \includegraphics[width=0.75\linewidth]{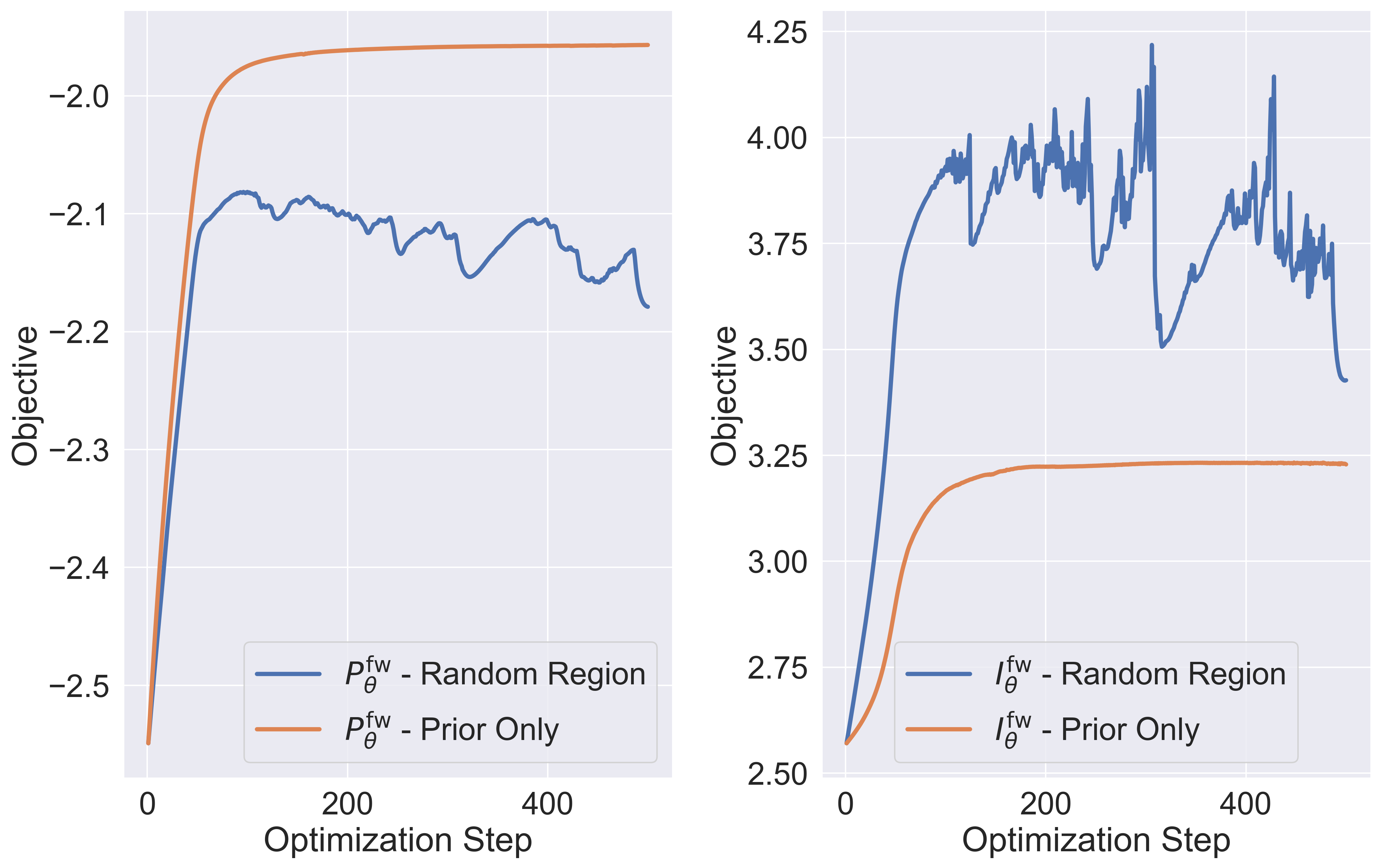}
  \caption{Impact of discarding the integral term in the random region attack, and only using the prior term in the prior-only attack. Left: The prior term across optimization. Right: The integral term across optimization.}
  \label{fig:prior_only_effect}
\end{figure}

Some of these issues with the convergence of the random region attack are alleviated by the prior-only attack, for which the convergence is shown in Figure~\ref{fig:prior_only_effect}. In this case, convergence is drastically improved for \textit{both} terms, and the value of $\ode$ at the end of optimization is higher, indicating that the attack is more successful. This also shows that a correlation between $I_\theta^{\textnormal{rev}}(\mathbf{x}_1)$ and $ P_\theta^{\textnormal{rev}}(\mathbf{x}_1)$ exists, given that $I_\theta^{\textnormal{rev}}(\mathbf{x}_1)$ rapidly increases in the initial steps of optimization in the prior-only attack.

\section{Additional Examples of Attack Outcomes}
\label{app:more_examples}
\begin{figure}[tp]
  \centering
  \setlength{\tabcolsep}{2pt}
  \renewcommand\tabularxcolumn[1]{m{#1}}
  \newcolumntype{R}{>{\raggedleft\arraybackslash}X}
  \newcolumntype{C}{>{\centering\arraybackslash}X}
  \begin{tabularx}{0.8\textwidth}{CCCC}
    $\lambda = 0.1$ \break\
      $\log \tilde{p}_\theta \textnormal{ = } \num{3.65}$  \break\
      $\log p_\theta \textnormal{ = } \num{3.10}$  \break\
      $\mathcal{C} \textnormal{ = } \num{0.794}$
     & $\lambda = 0.3$ \break\
      $\log \tilde{p}_\theta \textnormal{ = } \num{3.01}$  \break\
      $\log p_\theta \textnormal{ = } \num{2.49}$  \break\
      $\mathcal{C} \textnormal{ = } \num{0.837}$
     & $\lambda = 1$ \break\
      $\log \tilde{p}_\theta \textnormal{ = } \num{2.91}$  \break\
      $\log p_\theta \textnormal{ = } \num{2.64}$  \break\
      $\mathcal{C} \textnormal{ = } \num{0.853}$
     & $\lambda = 3$ \break\
      $\log \tilde{p}_\theta \textnormal{ = } \num{2.07}$  \break\
      $\log p_\theta \textnormal{ = } \num{1.96}$  \break\
      $\mathcal{C} \textnormal{ = } \num{0.857}$ \\
   \includegraphics[width=\linewidth]{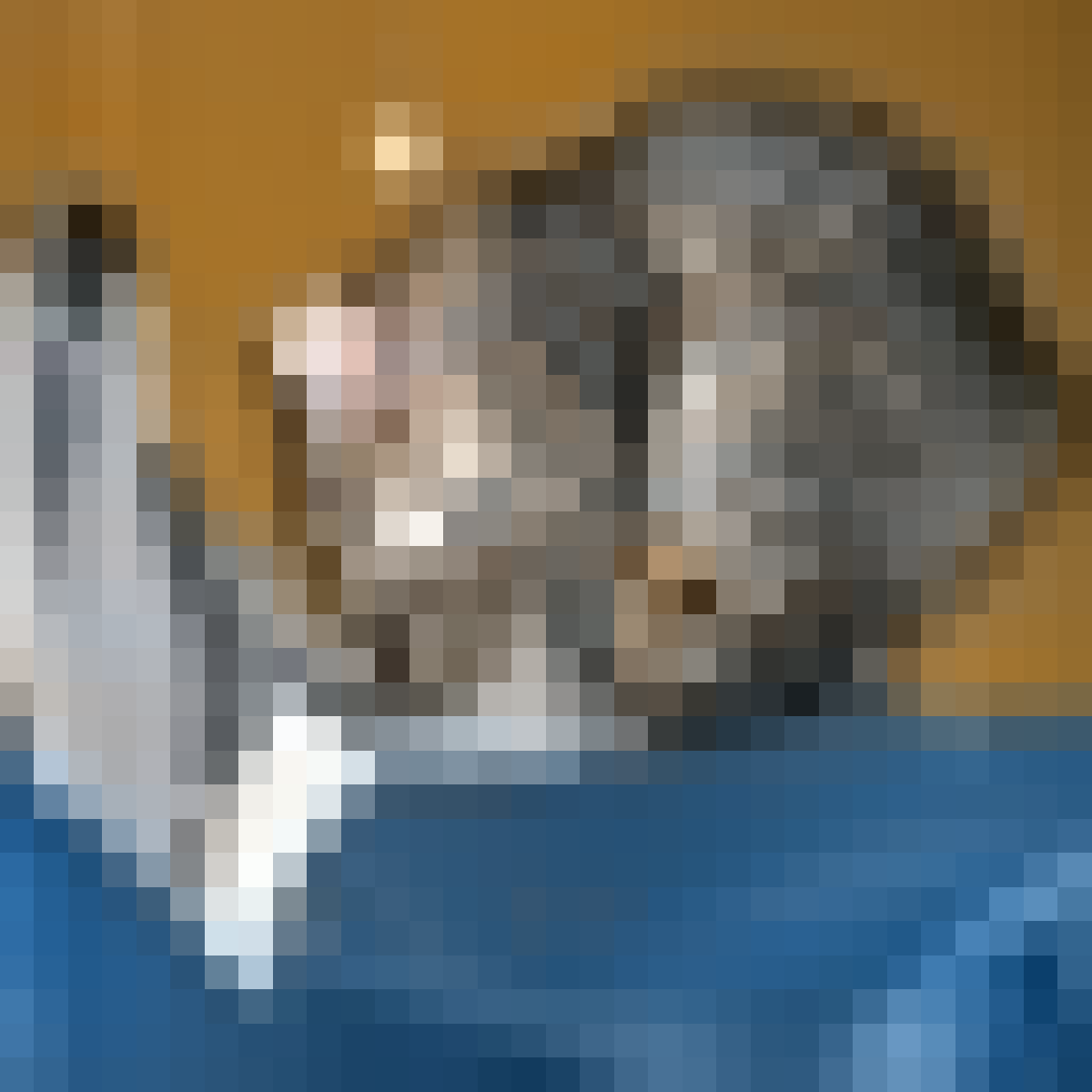} & \includegraphics[width=\linewidth]{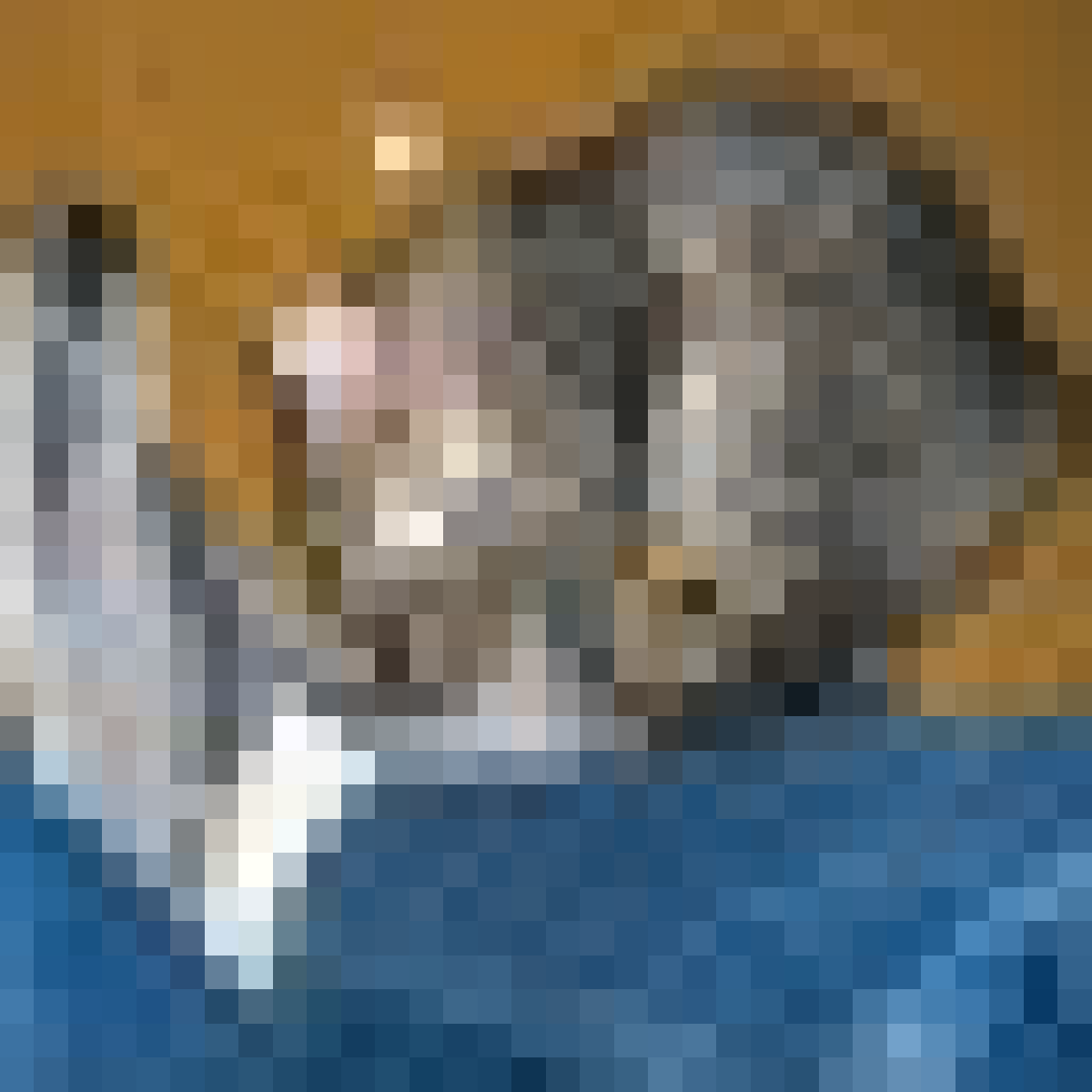} & \includegraphics[width=\linewidth]{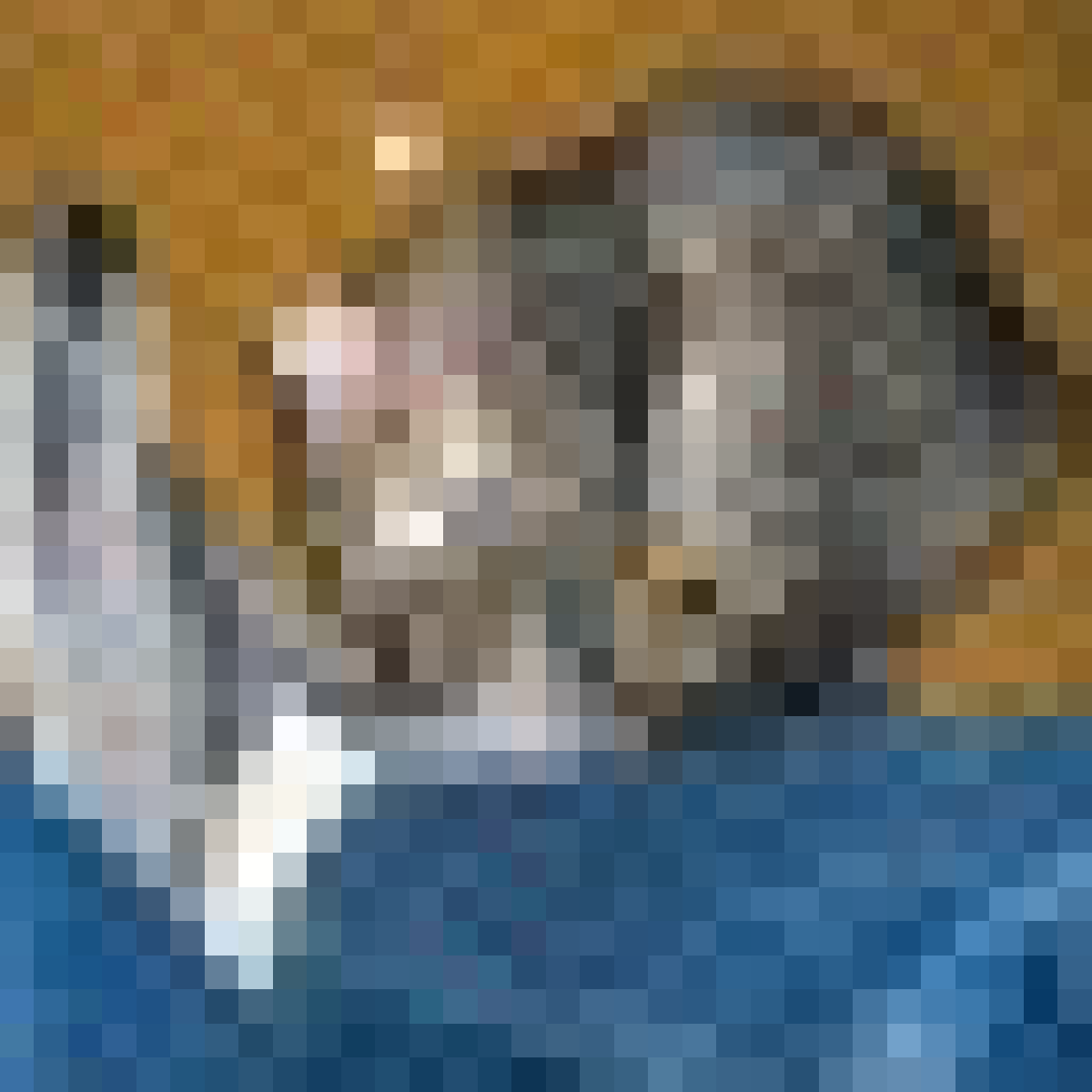} & \includegraphics[width=\linewidth]{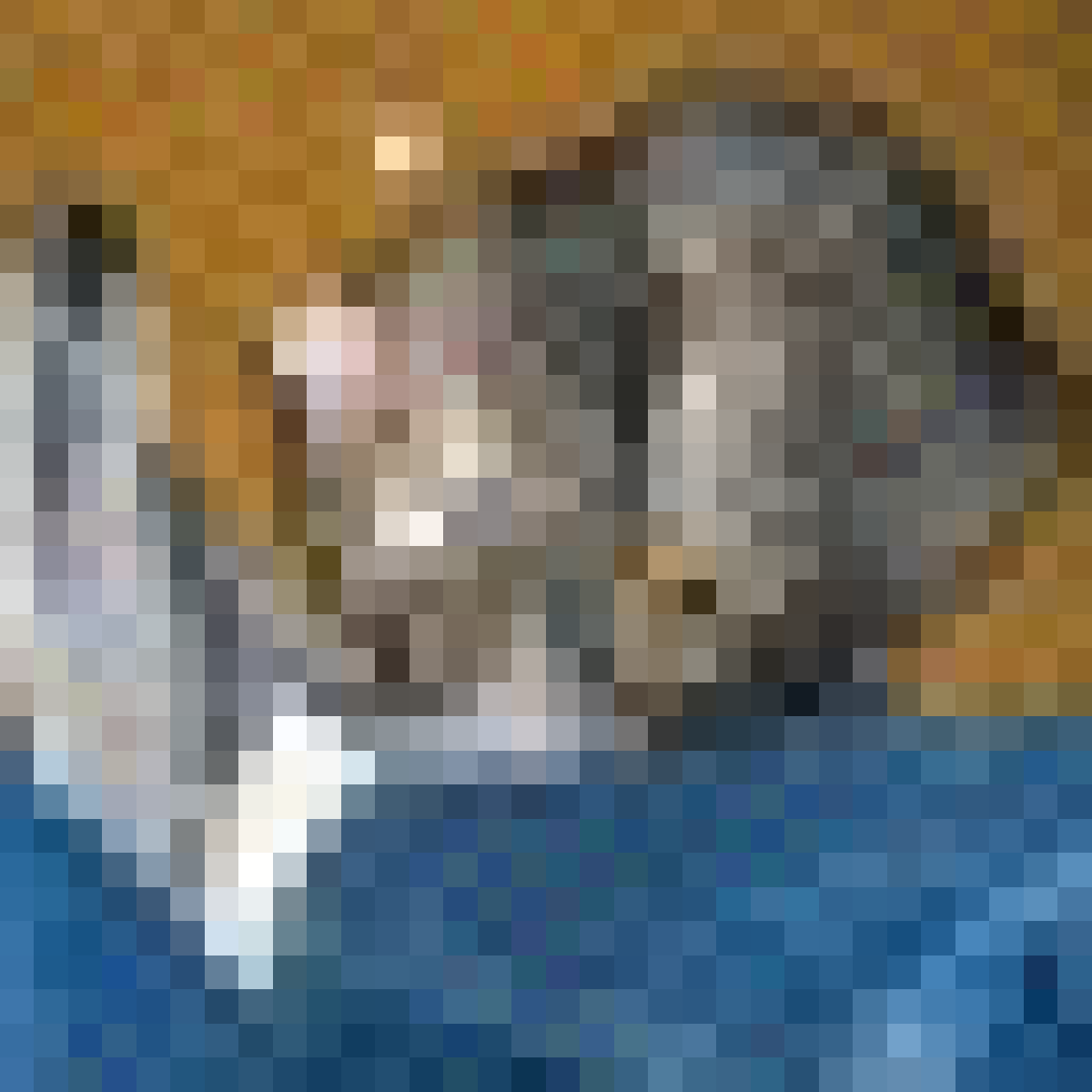} \\
  \end{tabularx}
  \caption{Outcomes of running the high-complexity attack for the same sample using different values of the hyper-parameter $\lambda$.}%
  \label{fig:dct_sweep}
\end{figure}

Figure~\ref{fig:dct_sweep} shows the outcome of the high-complexity attack when the value of the $\lambda$ term that enforces the high-complexity of the sample is varied. This reinforces the conclusion that there is a trade-off between $\ode$ and $\mathcal{C}$ even in the adversarial setting, and also further shows that our differentiable approximation to $\mathcal{C}$ is accurate. Increasing $\lambda$ leads to increased sample complexity, and also generally a lower estimated log-likelihood. Furthermore, in the case $\lambda=0.3$ it can be noticed that the optimized high-frequency perturbation itself is semantic, and is mostly located around the object of interest in the sample (in this case, a cat).

\begin{figure}[p]
  \centering
  \setlength{\tabcolsep}{2pt}
  \renewcommand\tabularxcolumn[1]{m{#1}}
  \newcolumntype{R}{>{\raggedleft\arraybackslash}X}
  \newcolumntype{C}{>{\centering\arraybackslash}X}
  \begin{tabularx}{\textwidth}{CCCCCCC}
   Benign & Near Sample & High Complexity & Reverse Integration & Random Region & Prior-Only & Unrestricted \\
   \includegraphics[width=\linewidth]{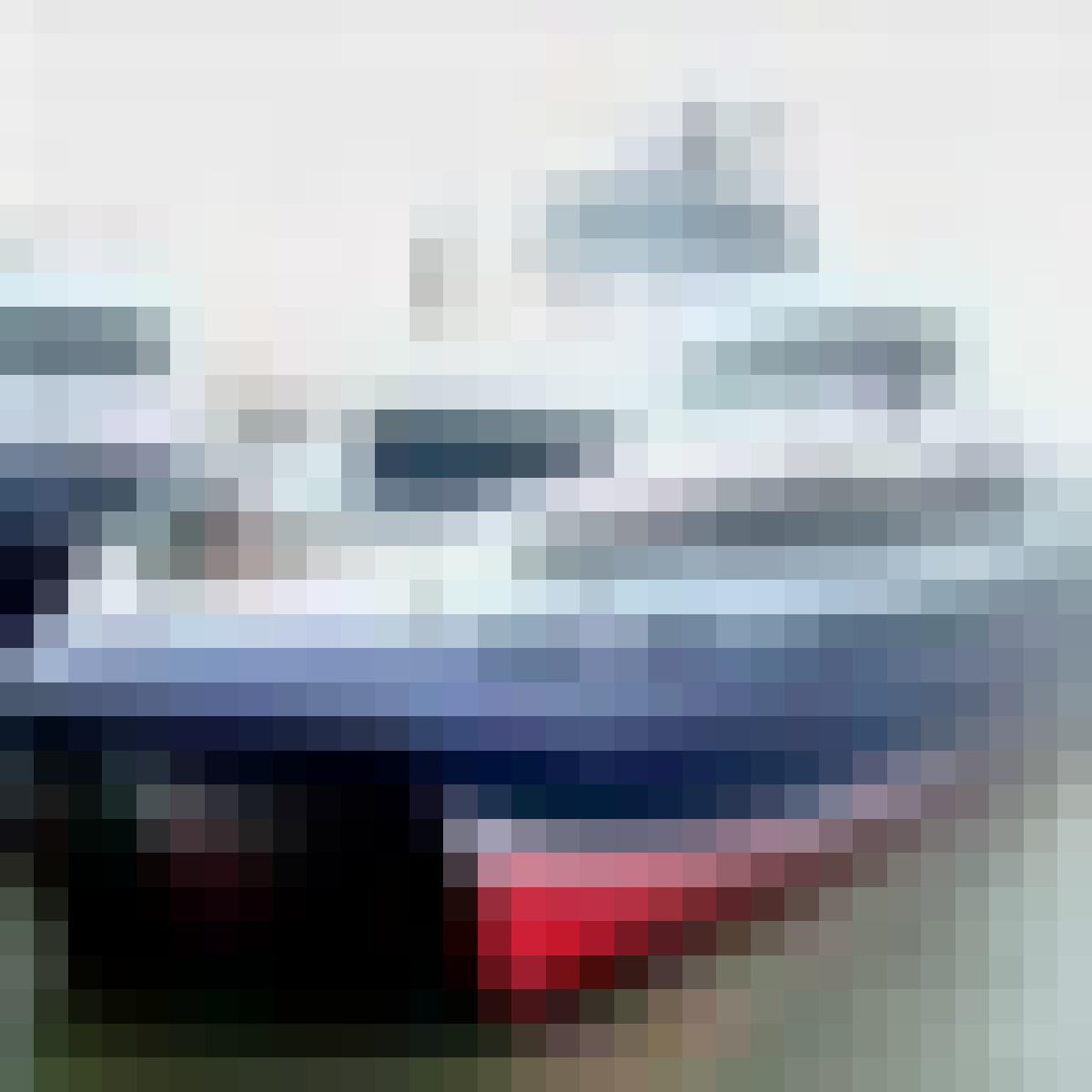} & 
   \includegraphics[width=\linewidth]{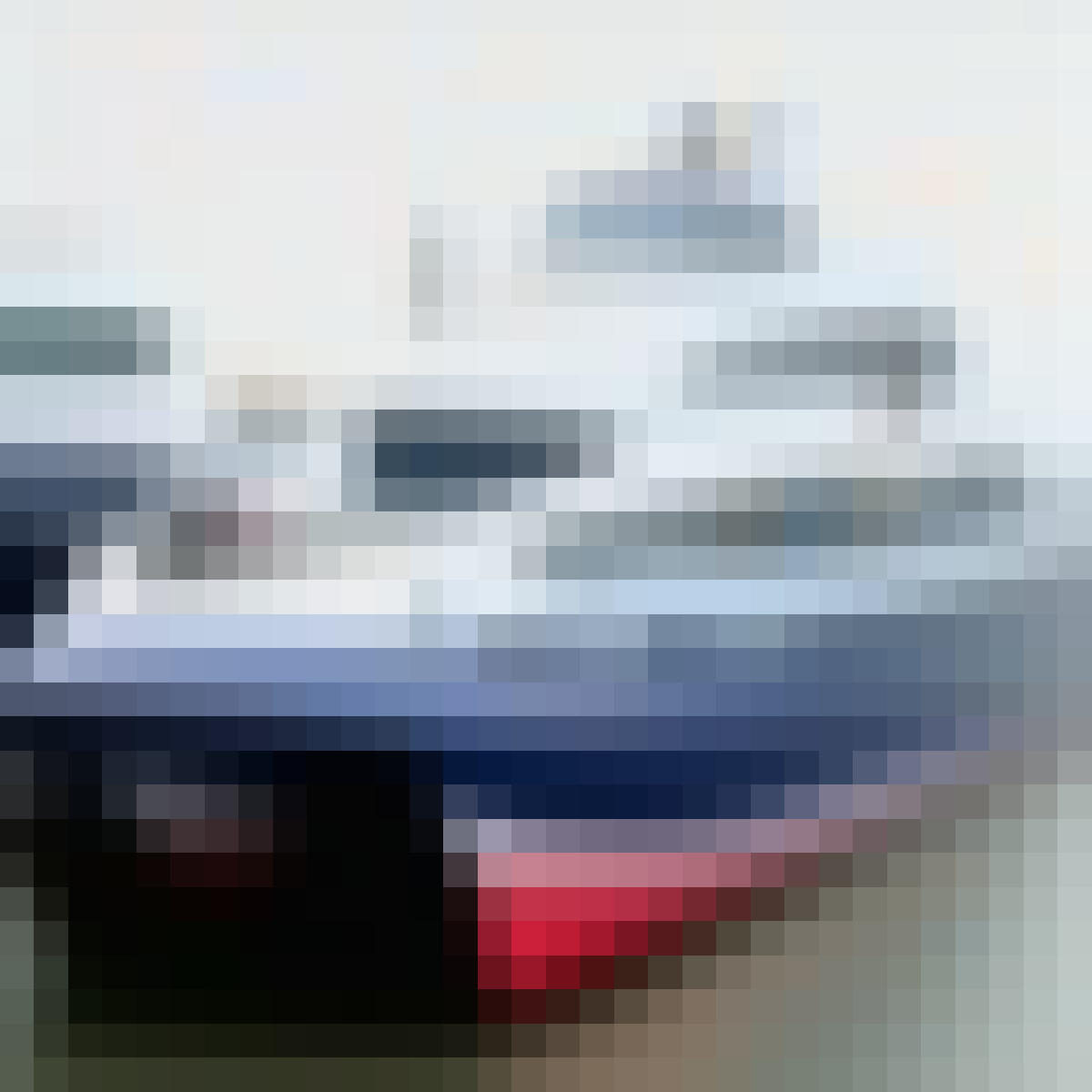} & 
   \includegraphics[width=\linewidth]{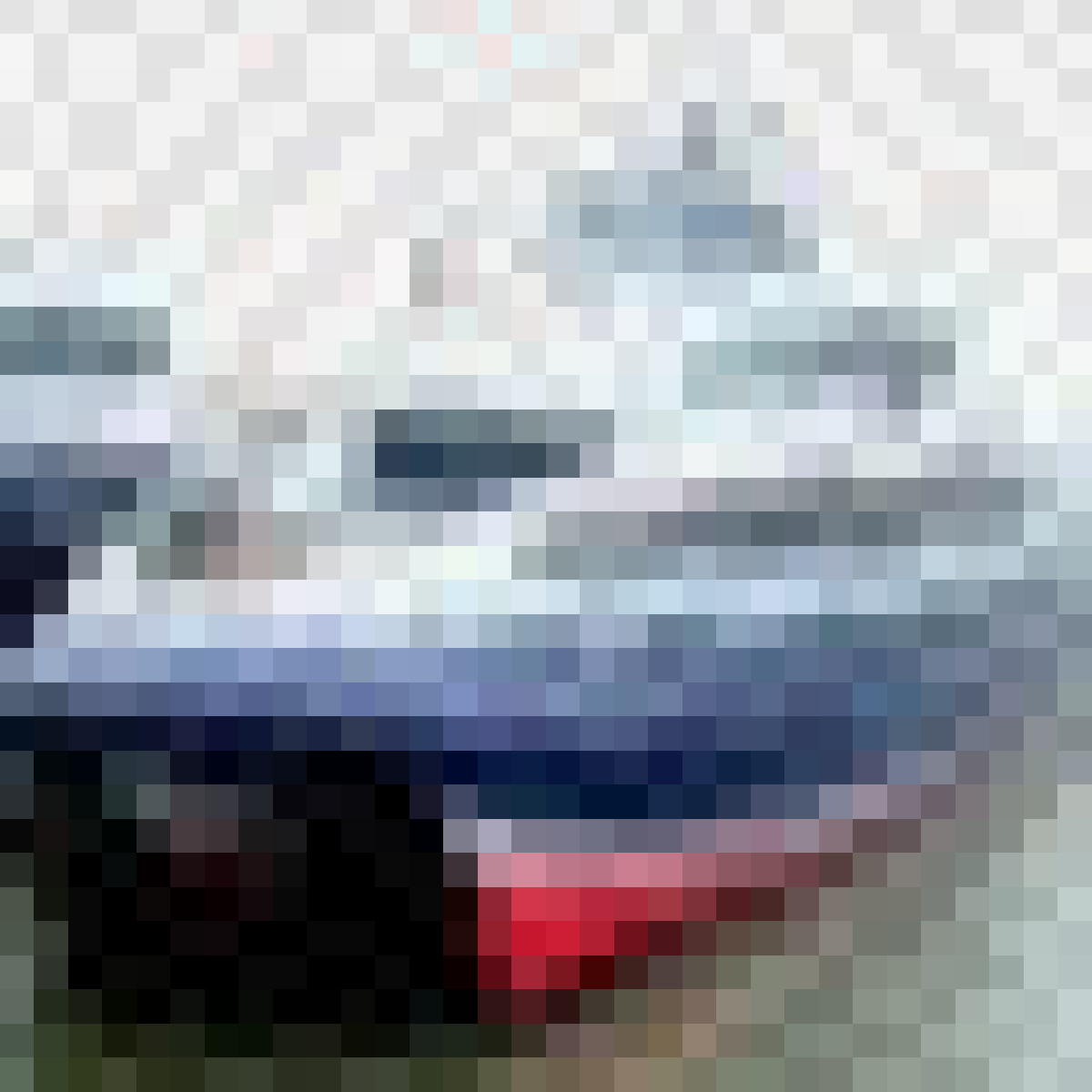} & 
   \includegraphics[width=\linewidth]{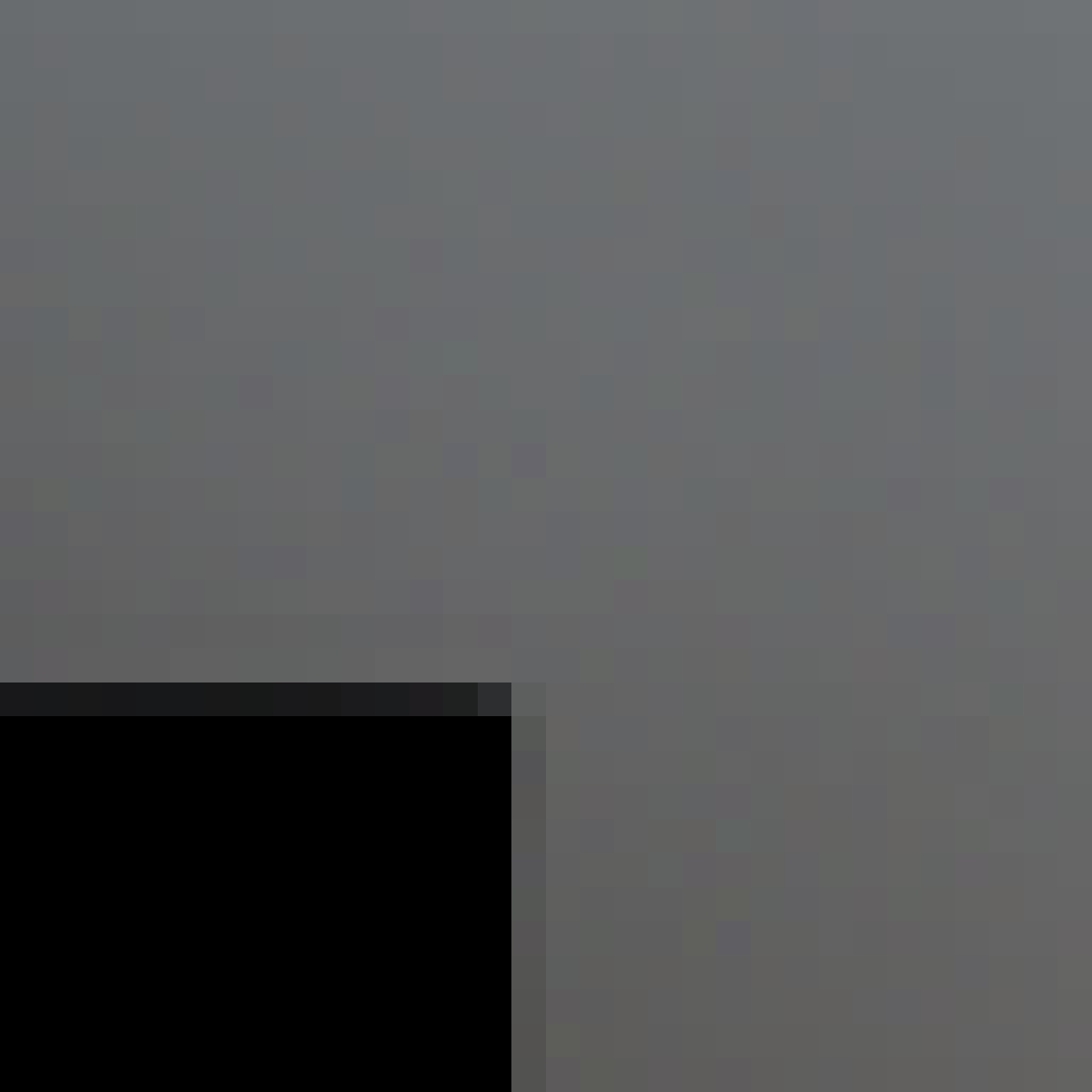} & 
   \includegraphics[width=\linewidth]{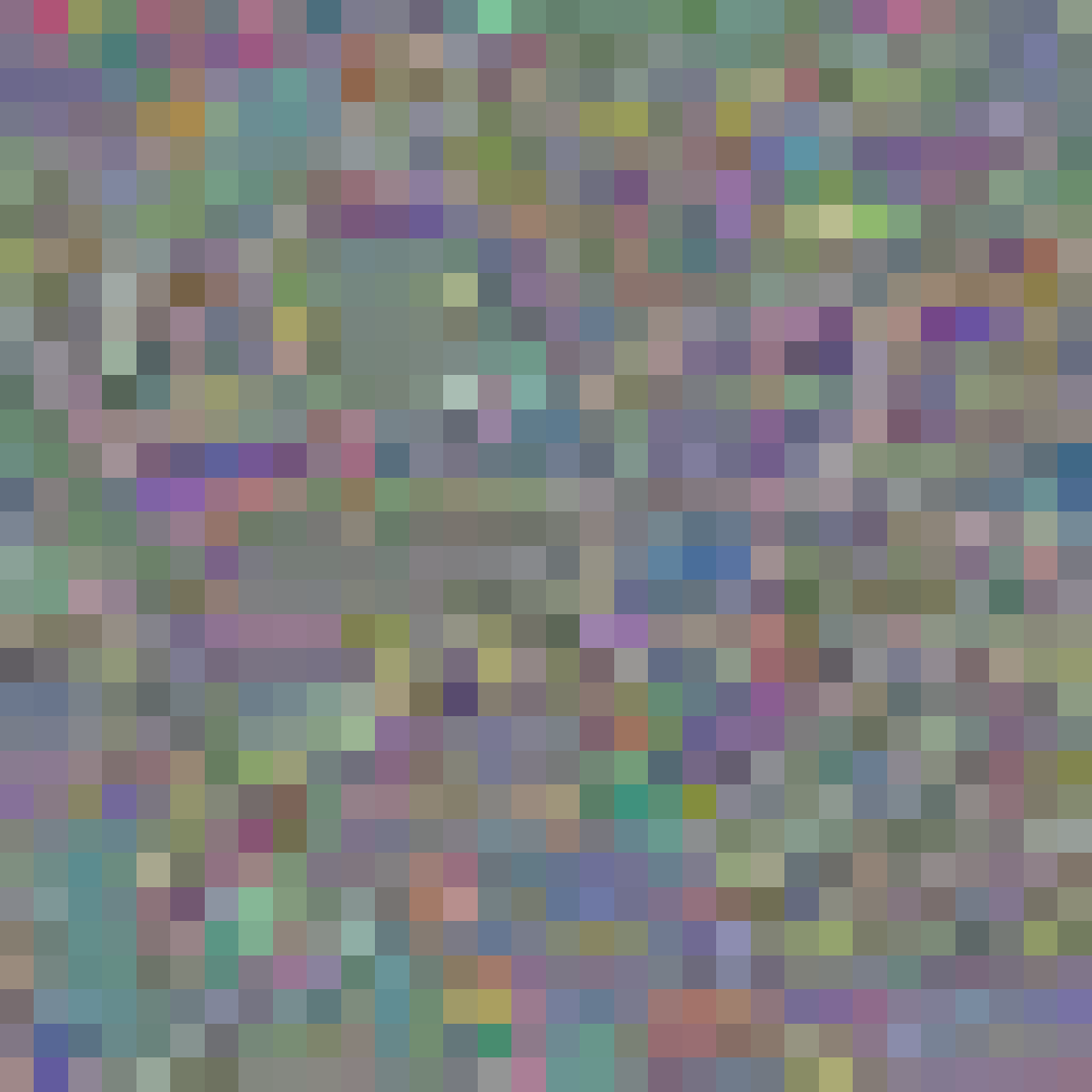} &
   \includegraphics[width=\linewidth]{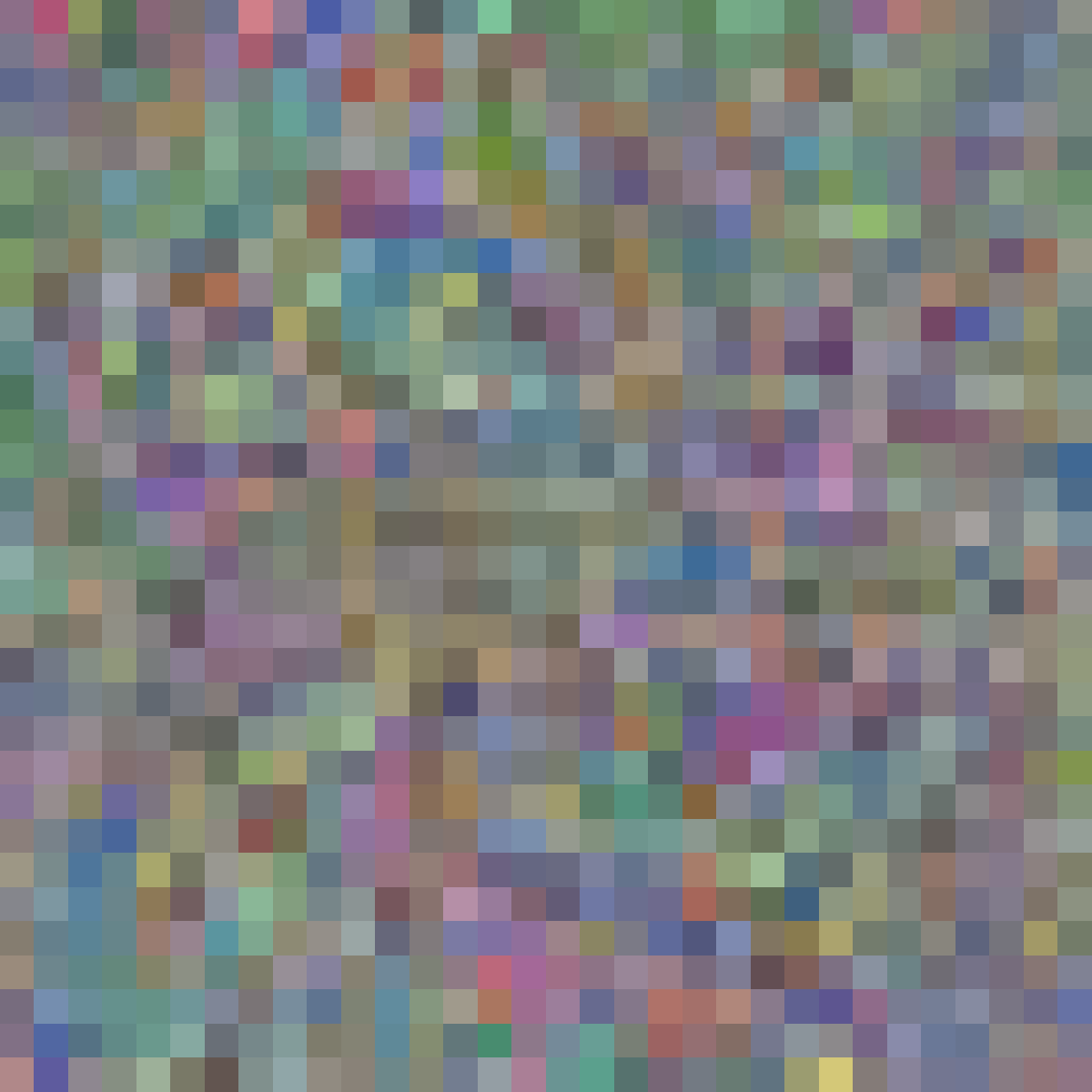} &
   \includegraphics[width=\linewidth]{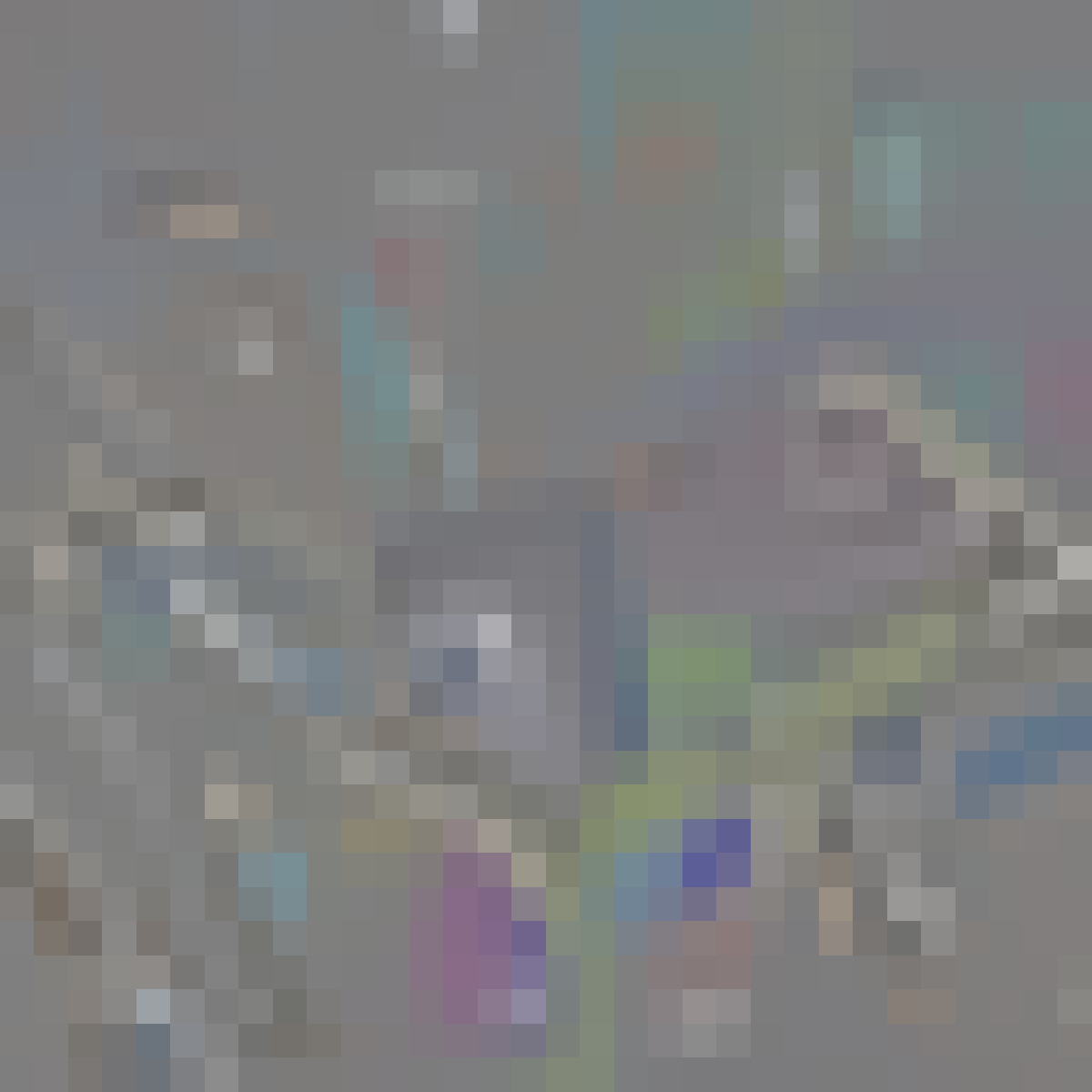} \\
   \includegraphics[width=\linewidth]{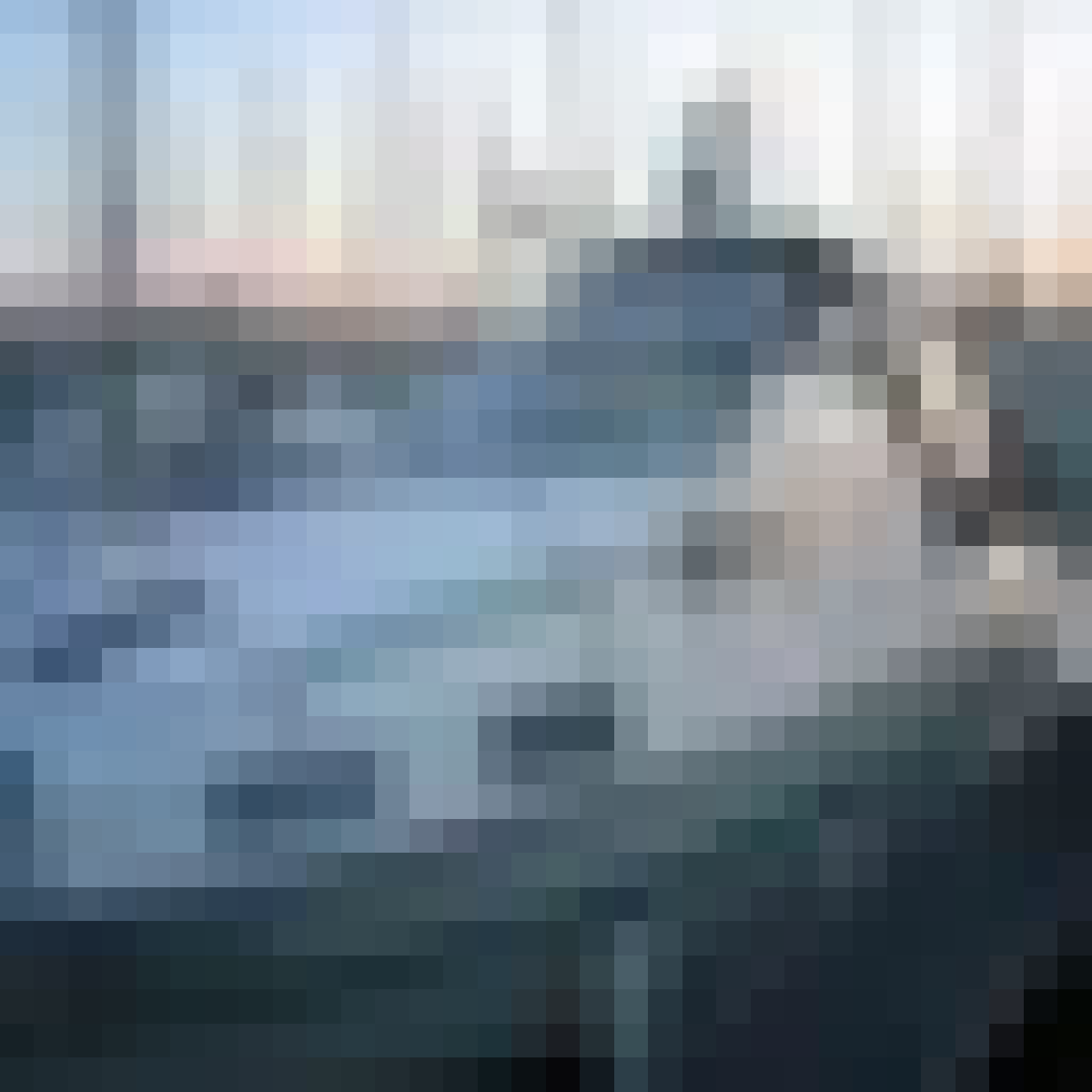} & 
   \includegraphics[width=\linewidth]{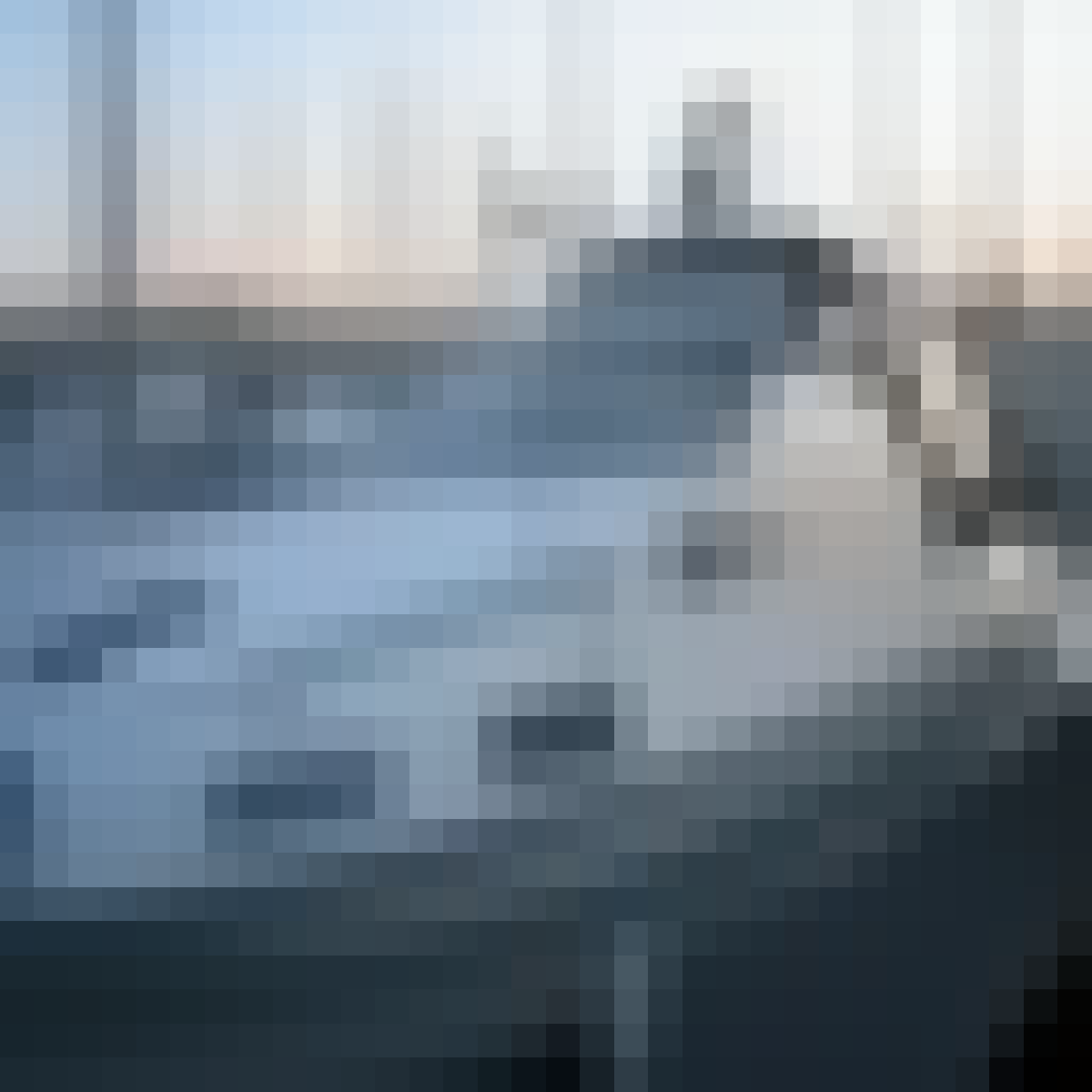} & 
   \includegraphics[width=\linewidth]{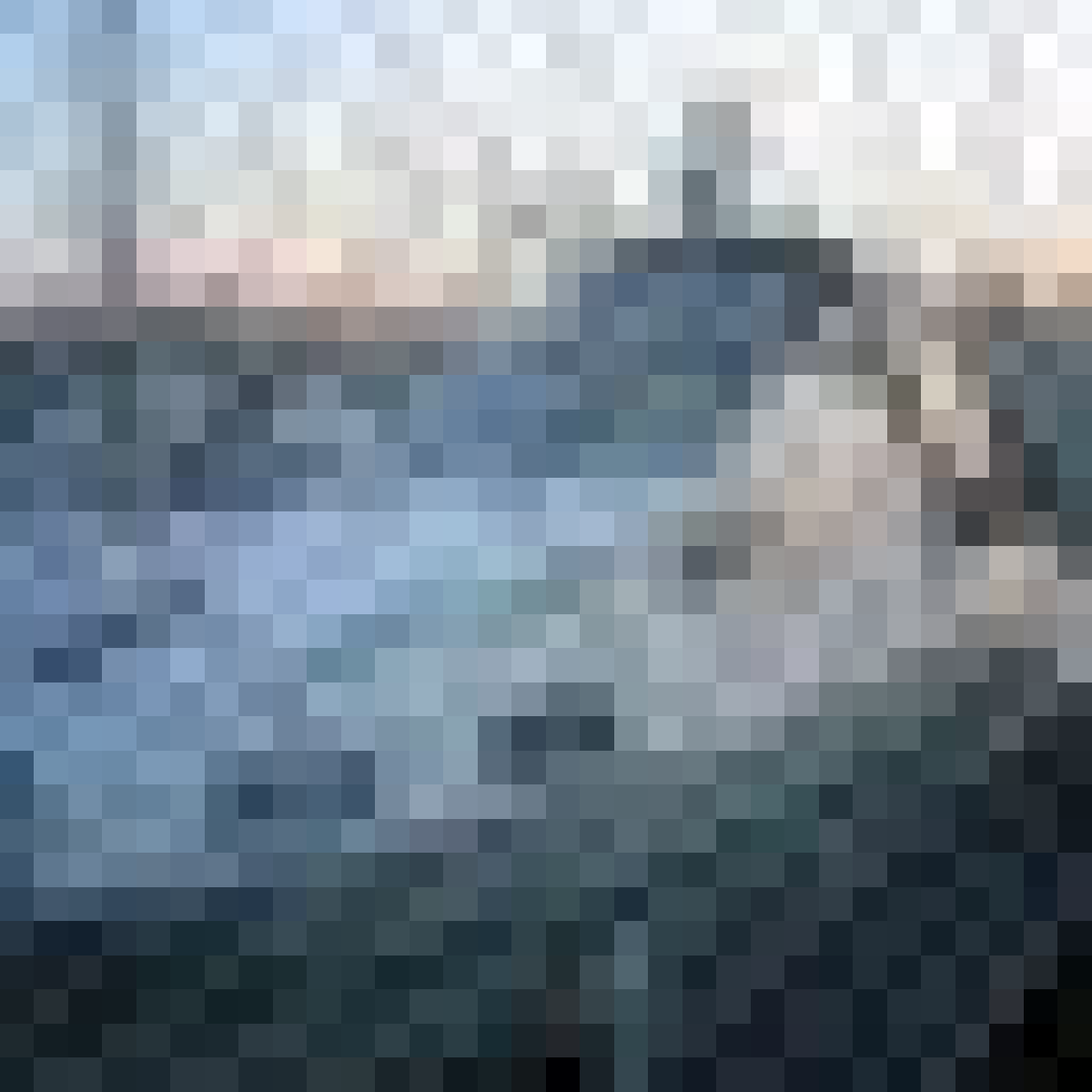} & 
   \includegraphics[width=\linewidth]{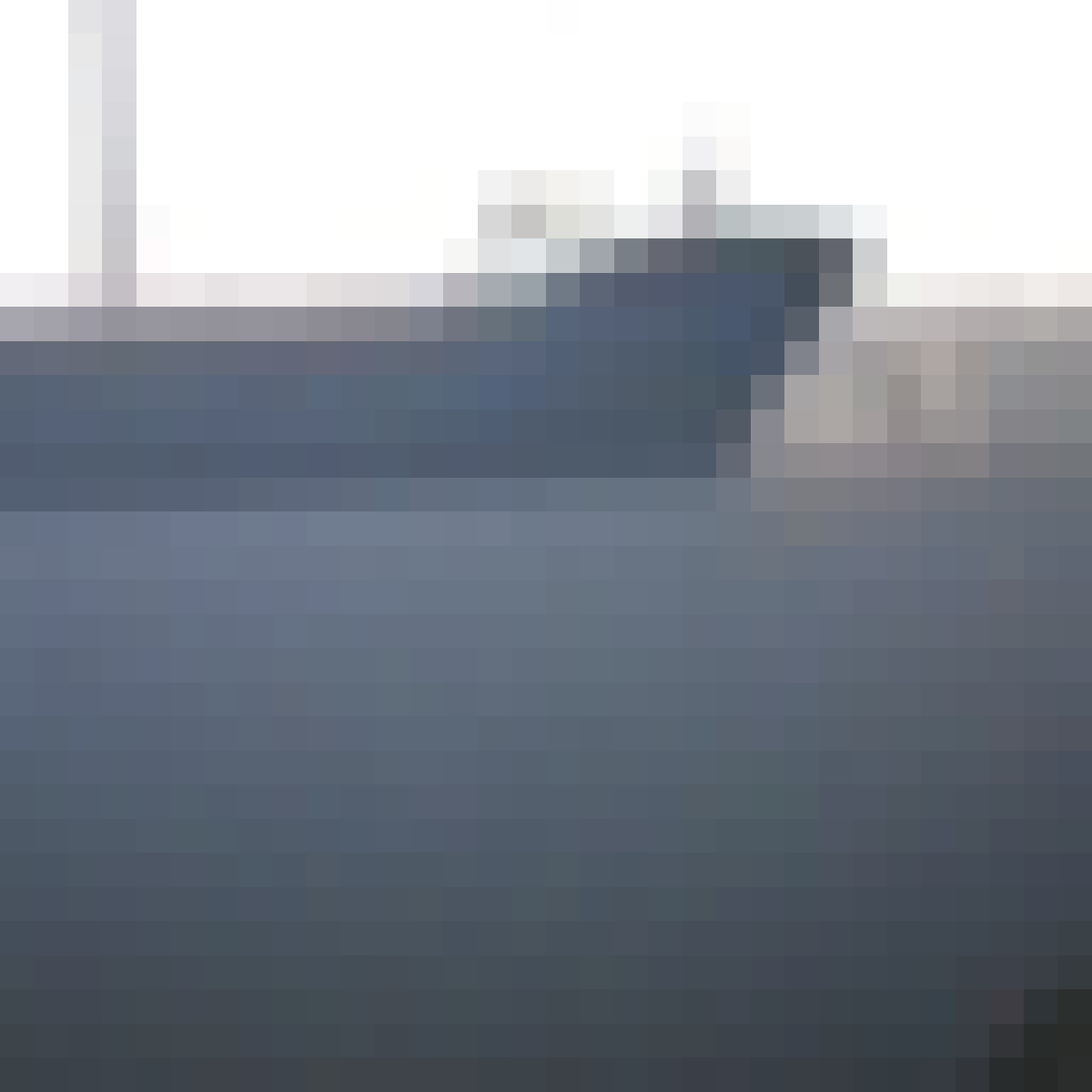} & 
   \includegraphics[width=\linewidth]{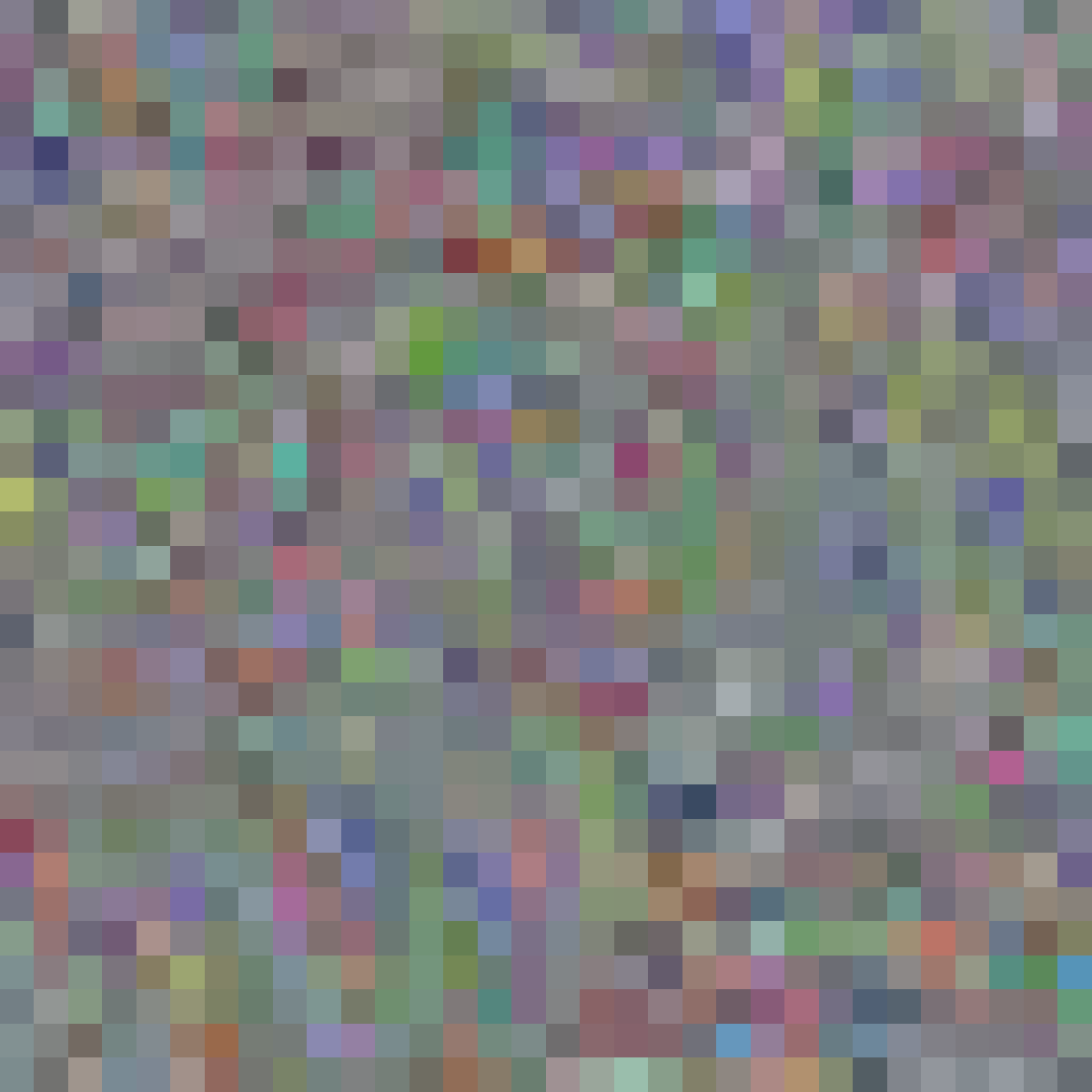} &
   \includegraphics[width=\linewidth]{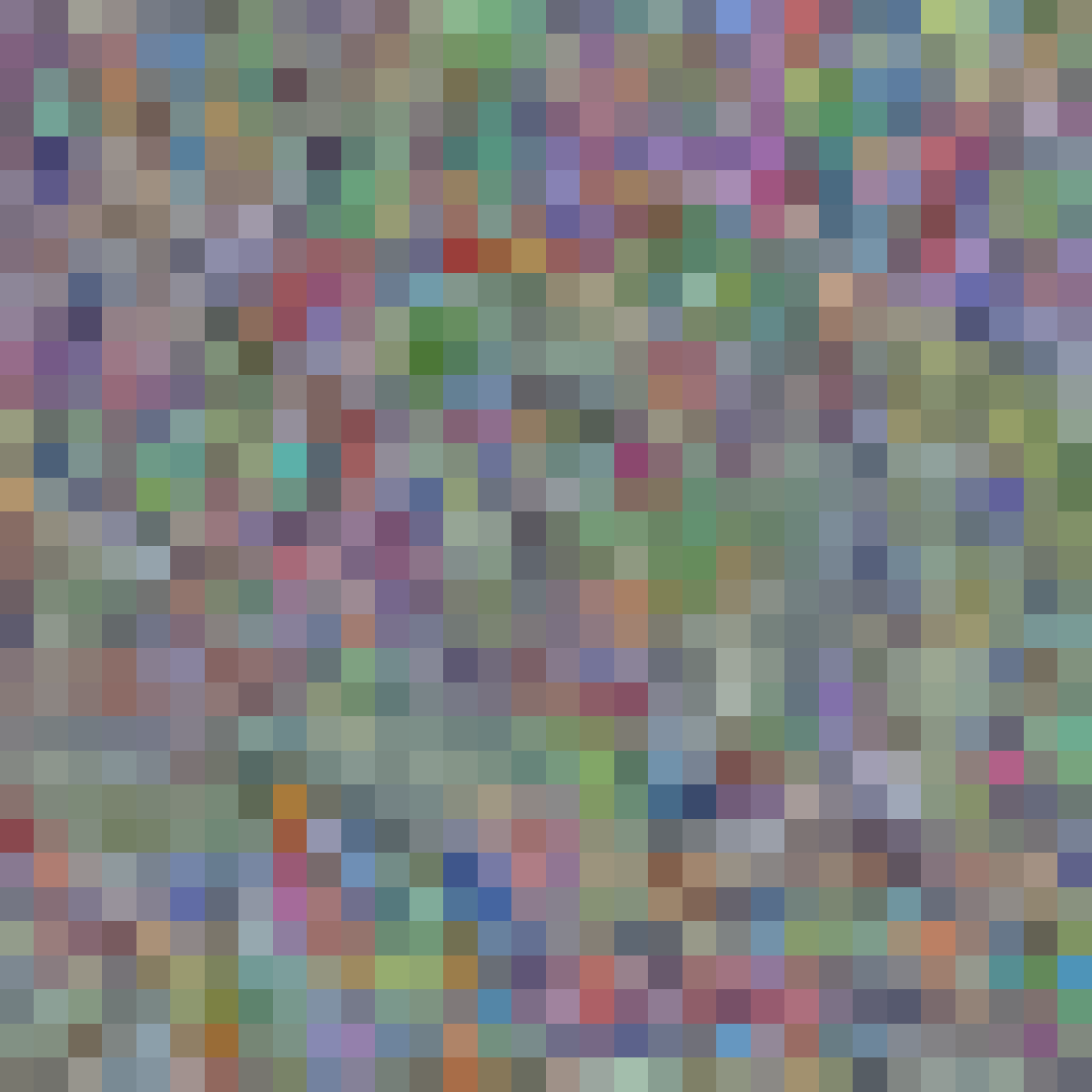} &
   \includegraphics[width=\linewidth]{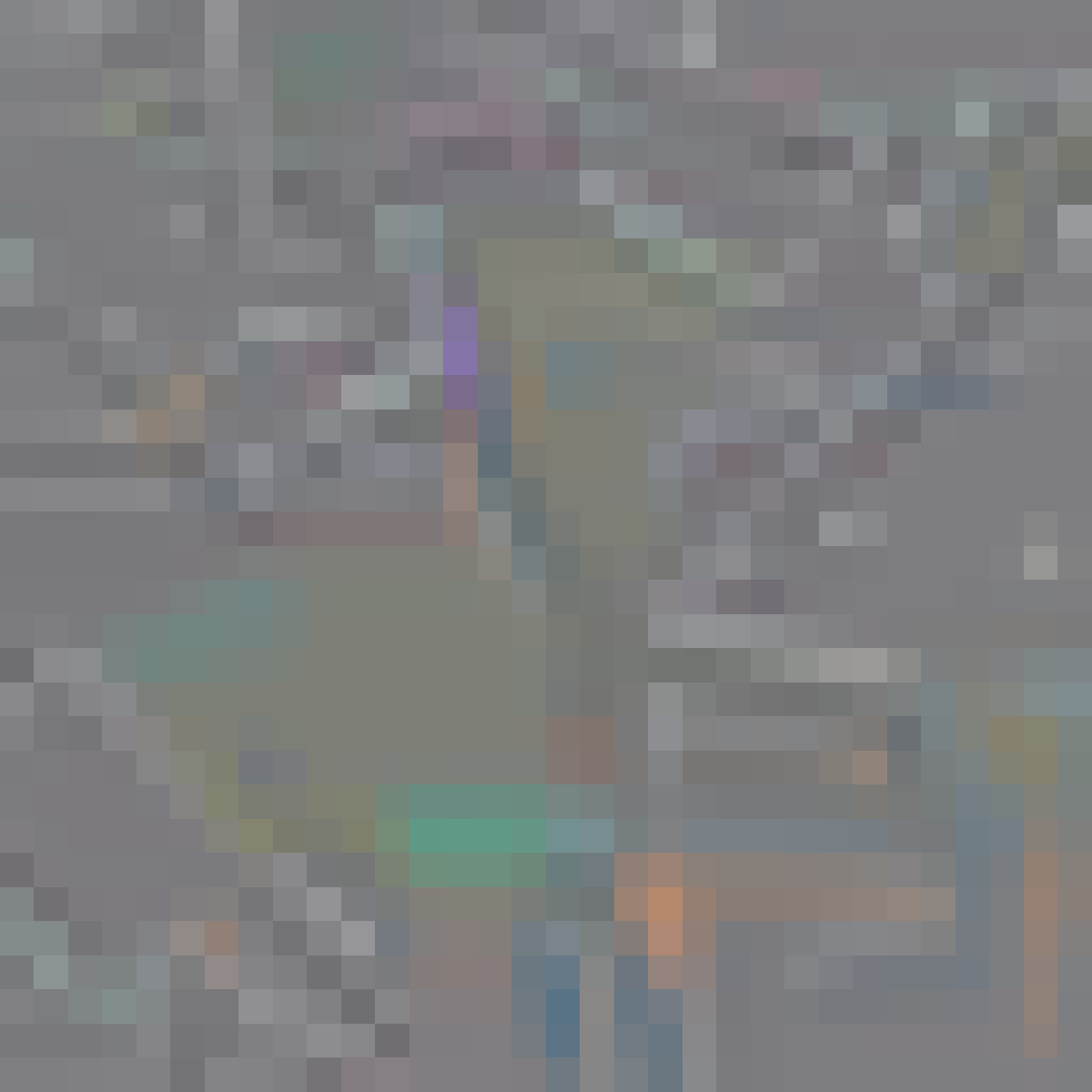} \\
   \includegraphics[width=\linewidth]{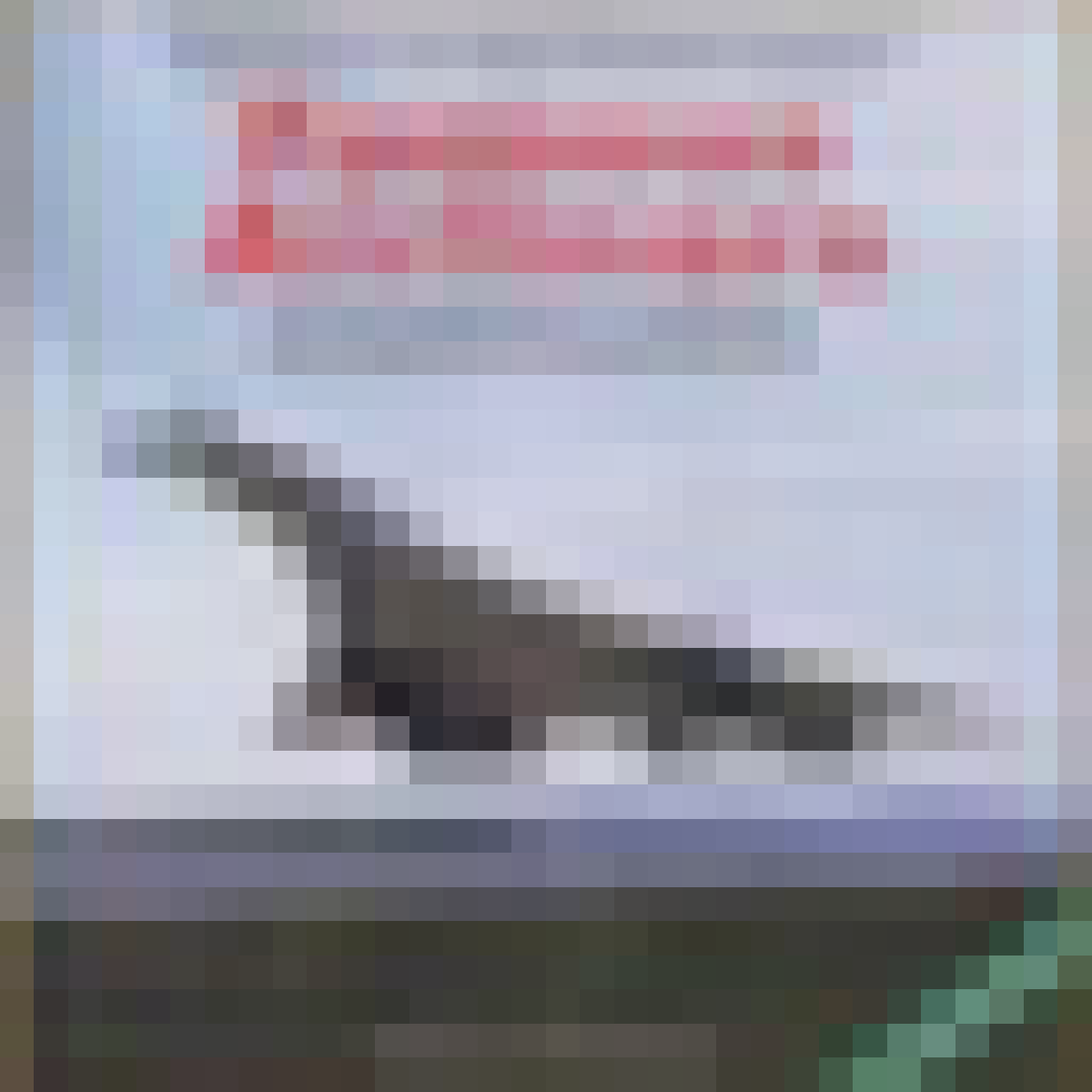} & 
   \includegraphics[width=\linewidth]{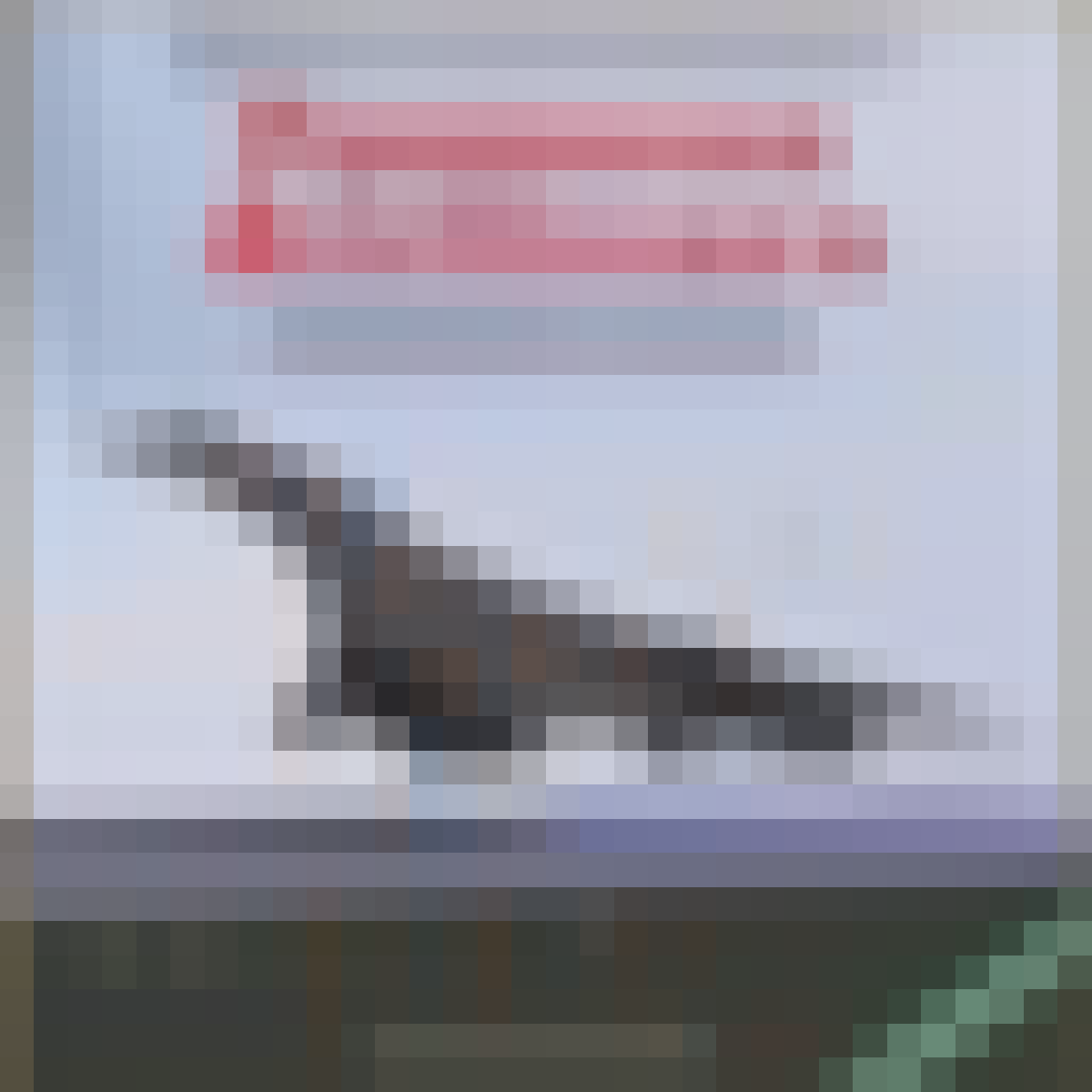} & 
   \includegraphics[width=\linewidth]{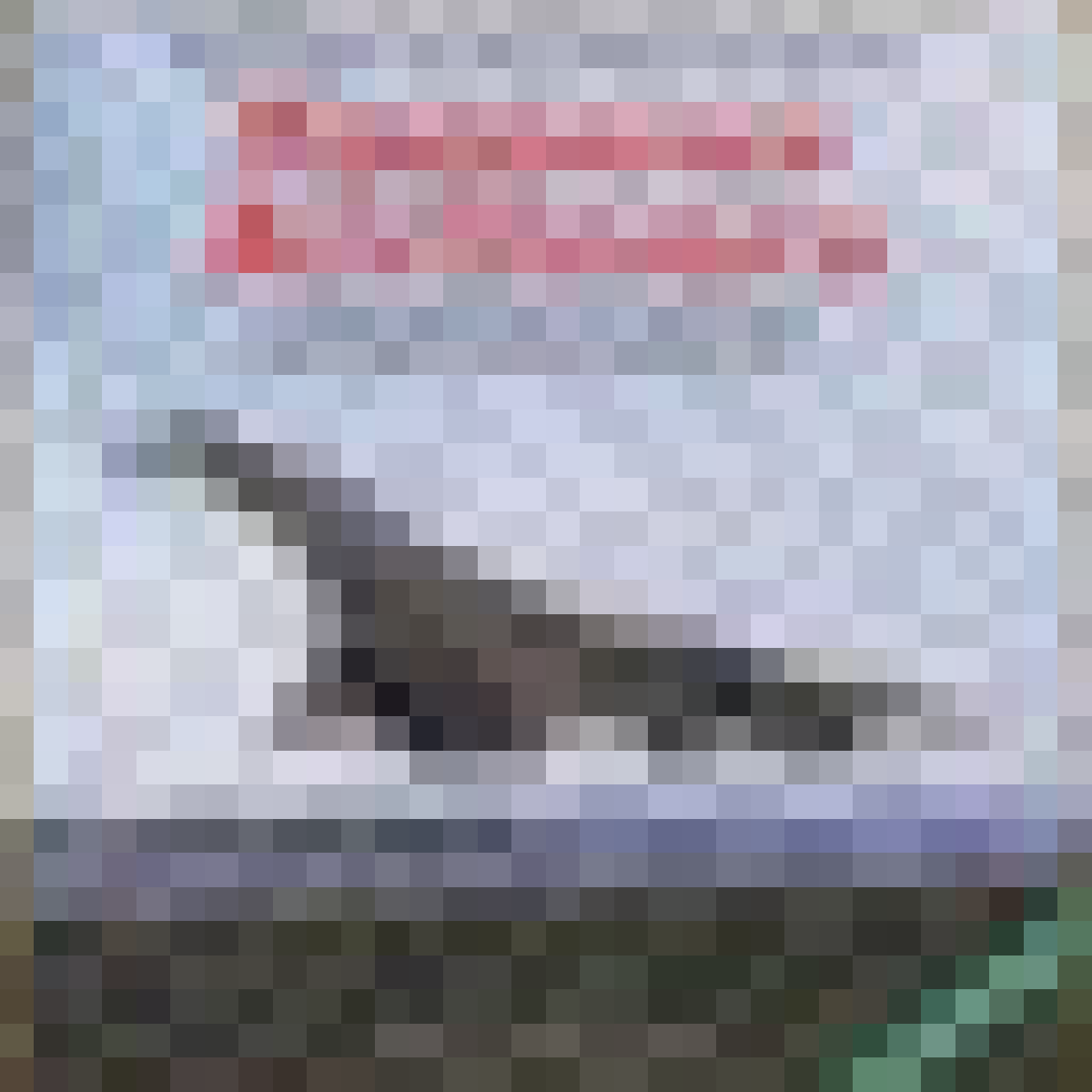} & 
   \includegraphics[width=\linewidth]{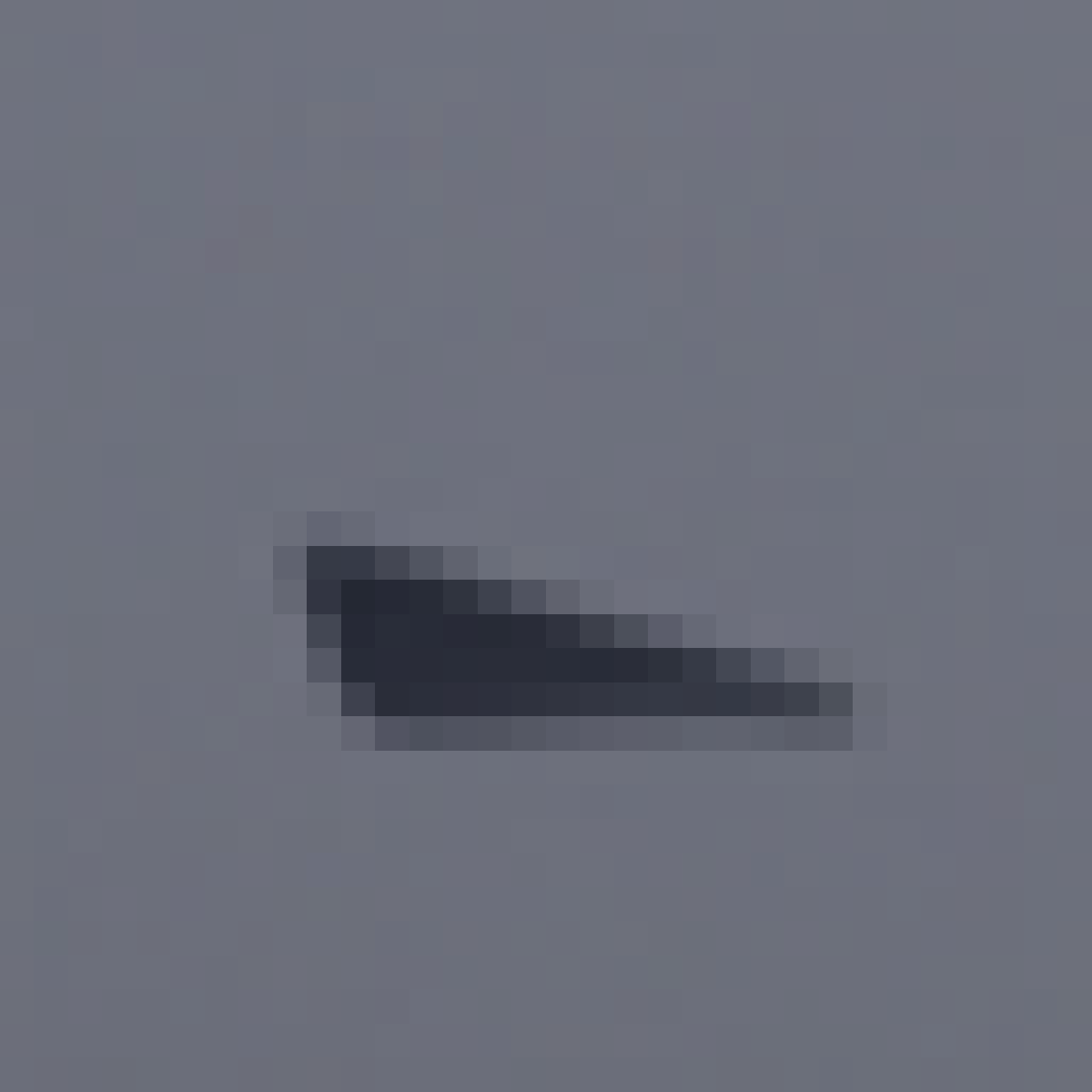} & 
   \includegraphics[width=\linewidth]{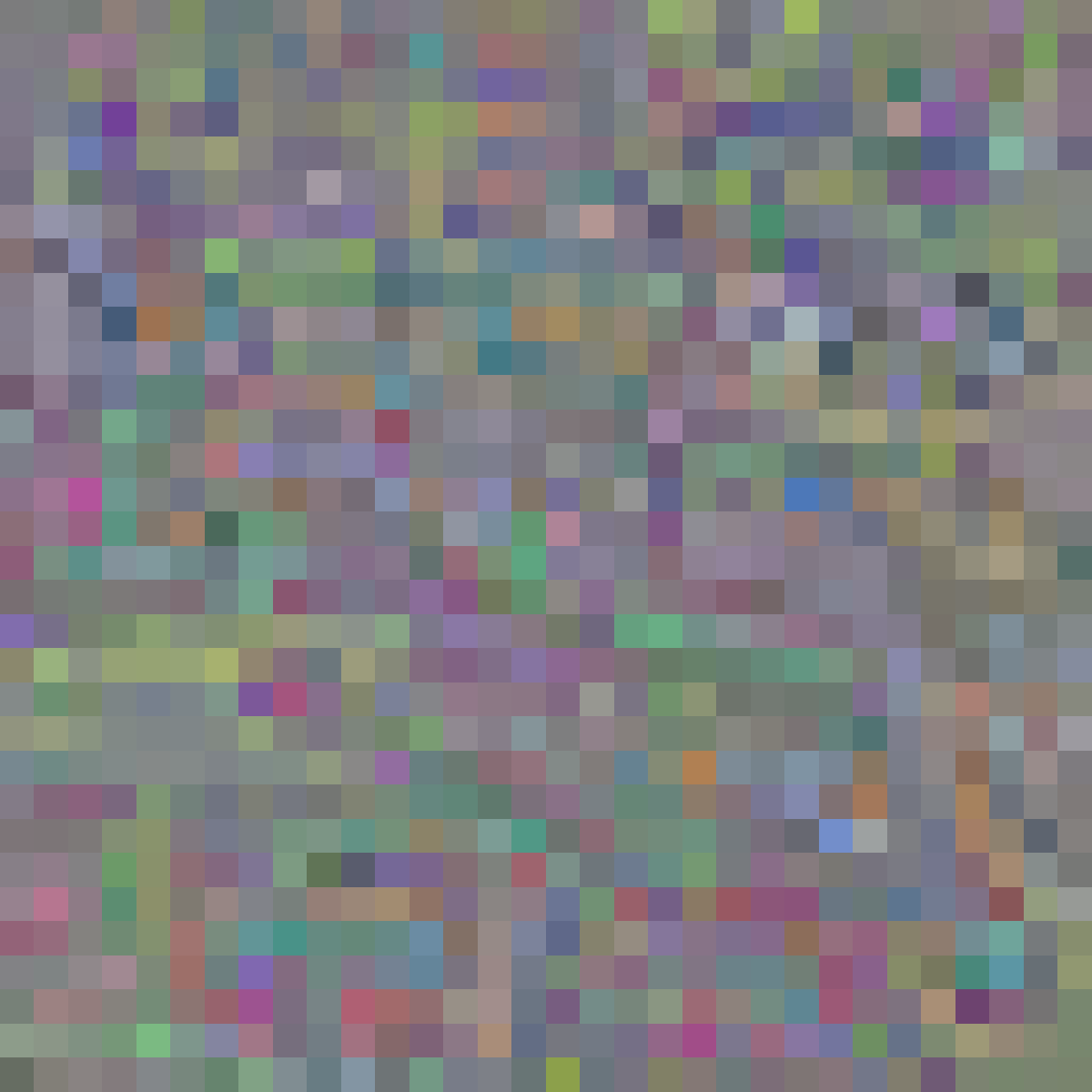} &
   \includegraphics[width=\linewidth]{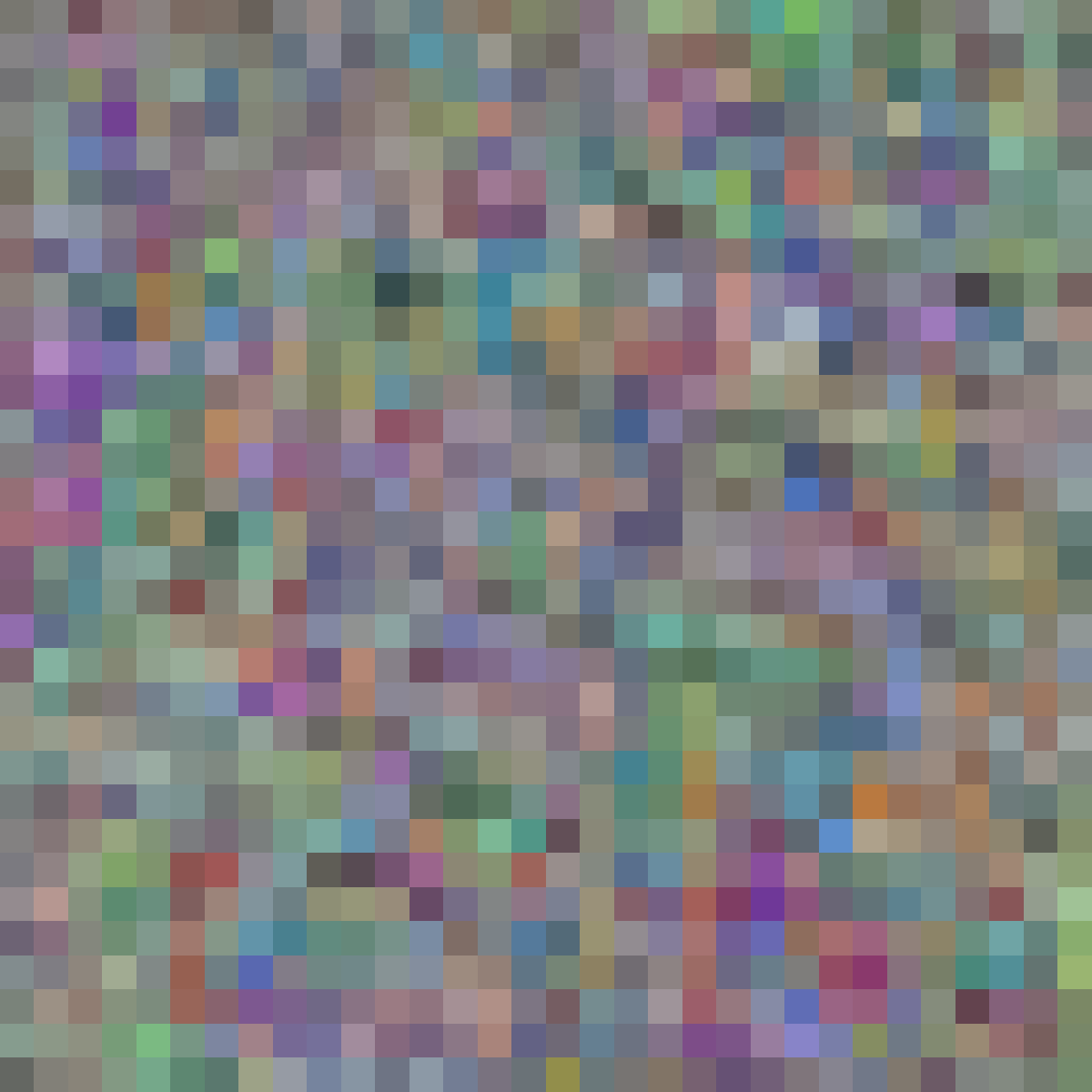} &
   \includegraphics[width=\linewidth]{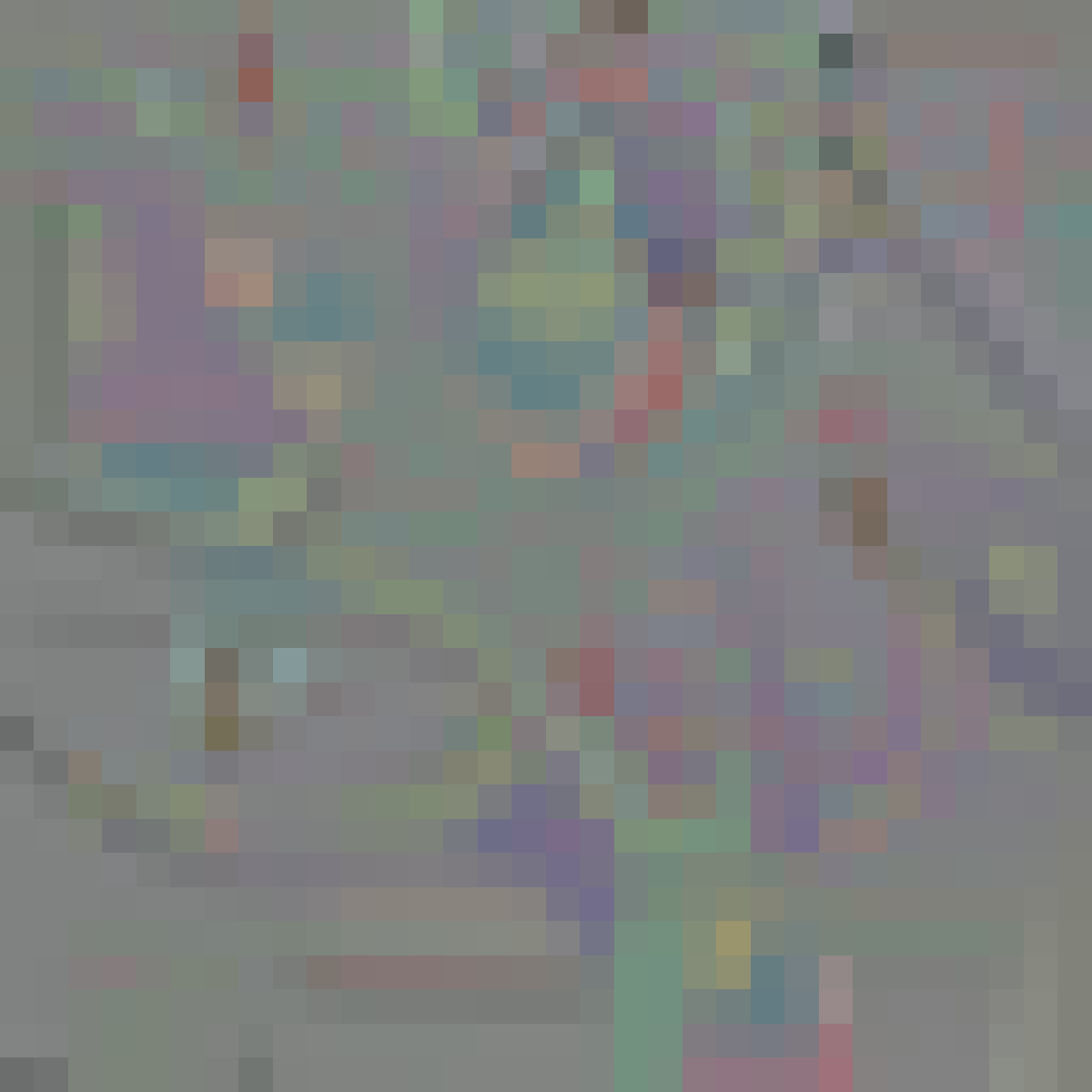} \\
   \includegraphics[width=\linewidth]{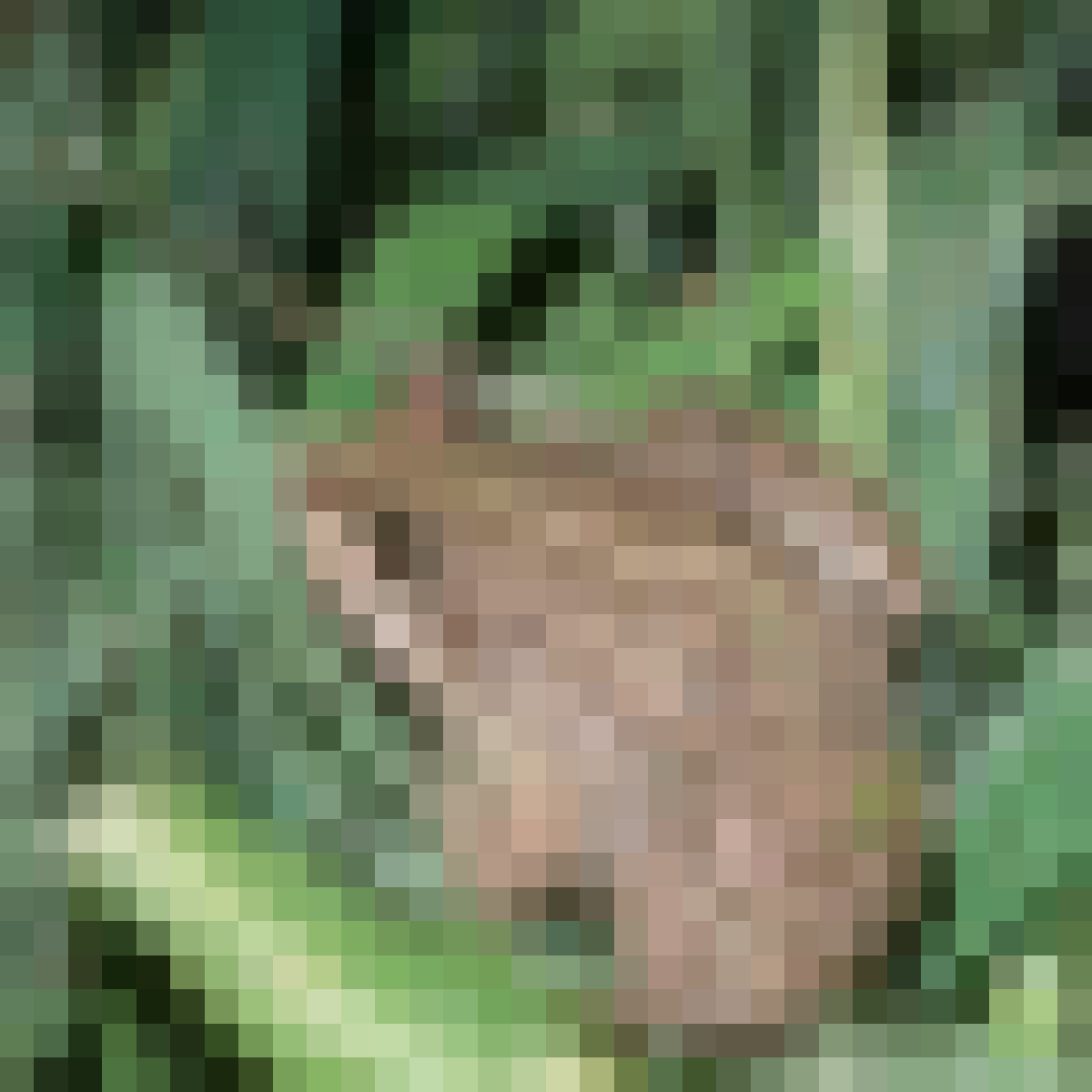} & 
   \includegraphics[width=\linewidth]{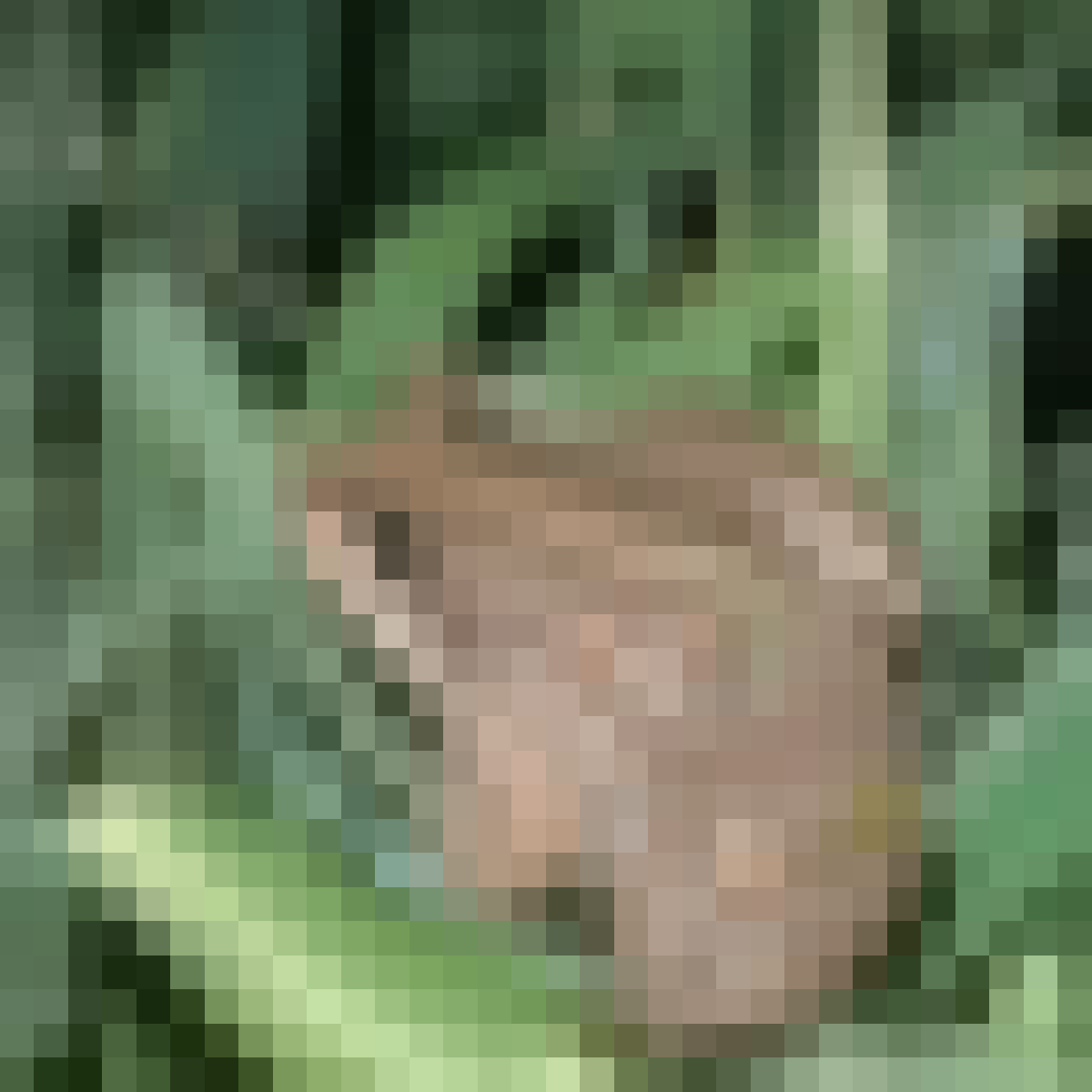} & 
   \includegraphics[width=\linewidth]{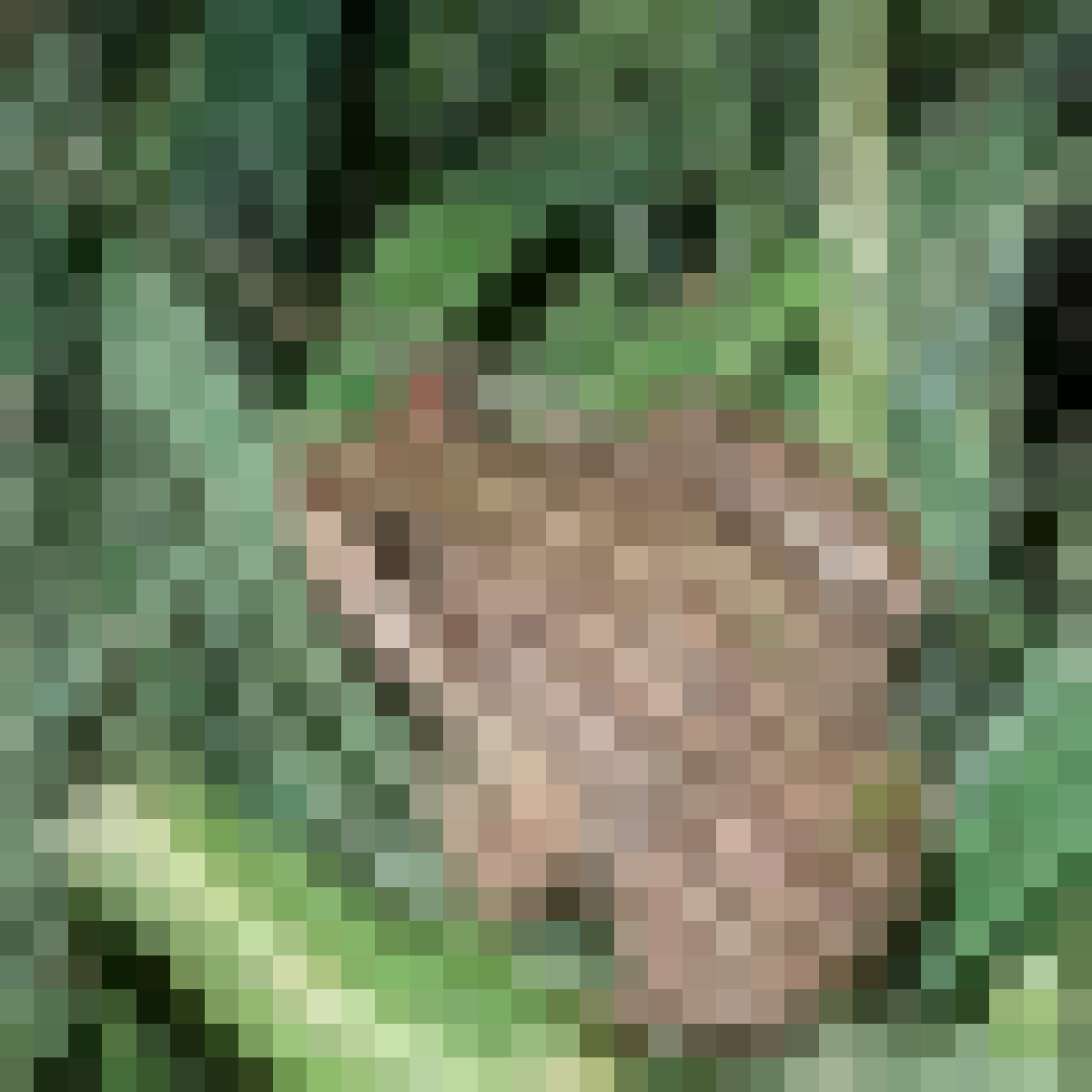} & 
   \includegraphics[width=\linewidth]{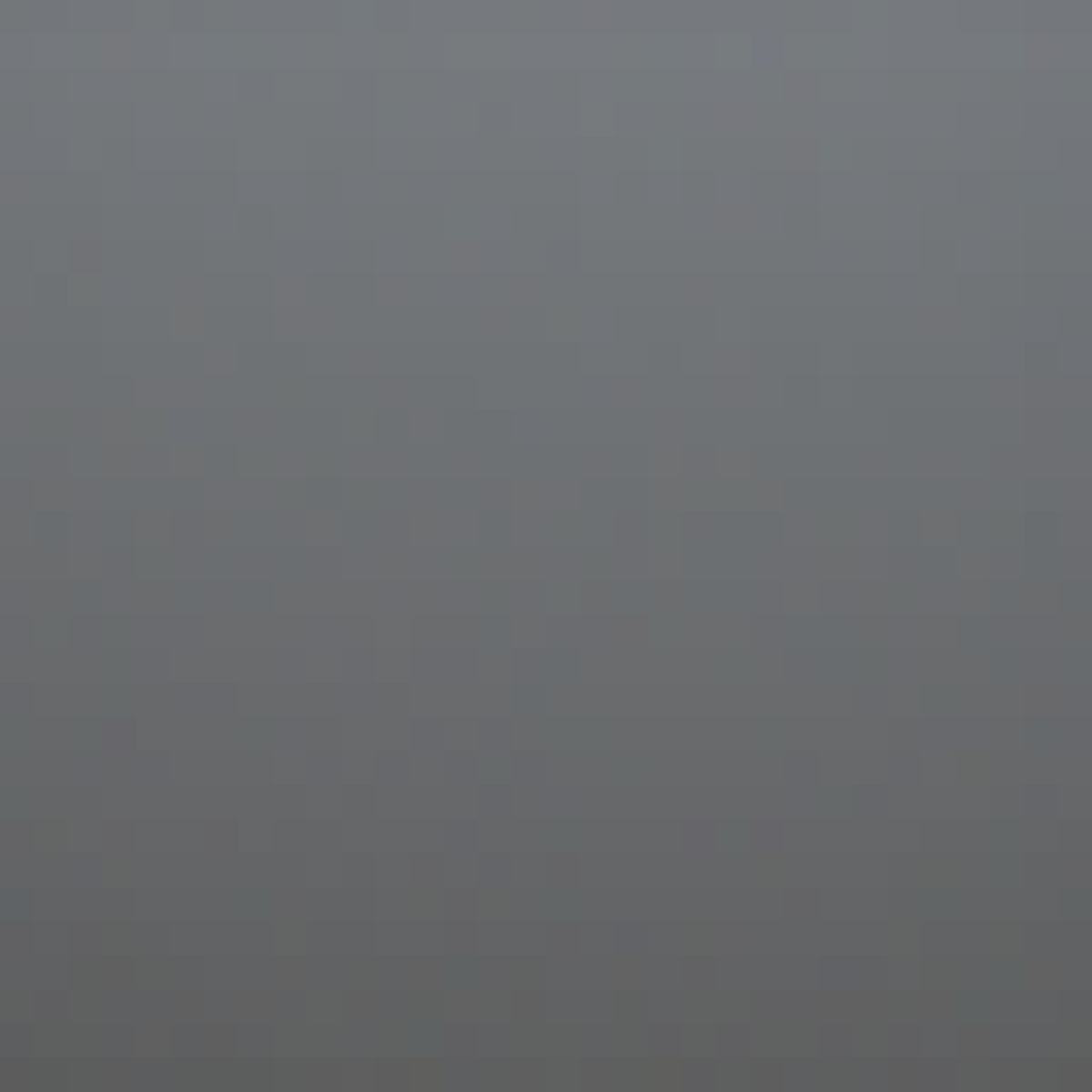} & 
   \includegraphics[width=\linewidth]{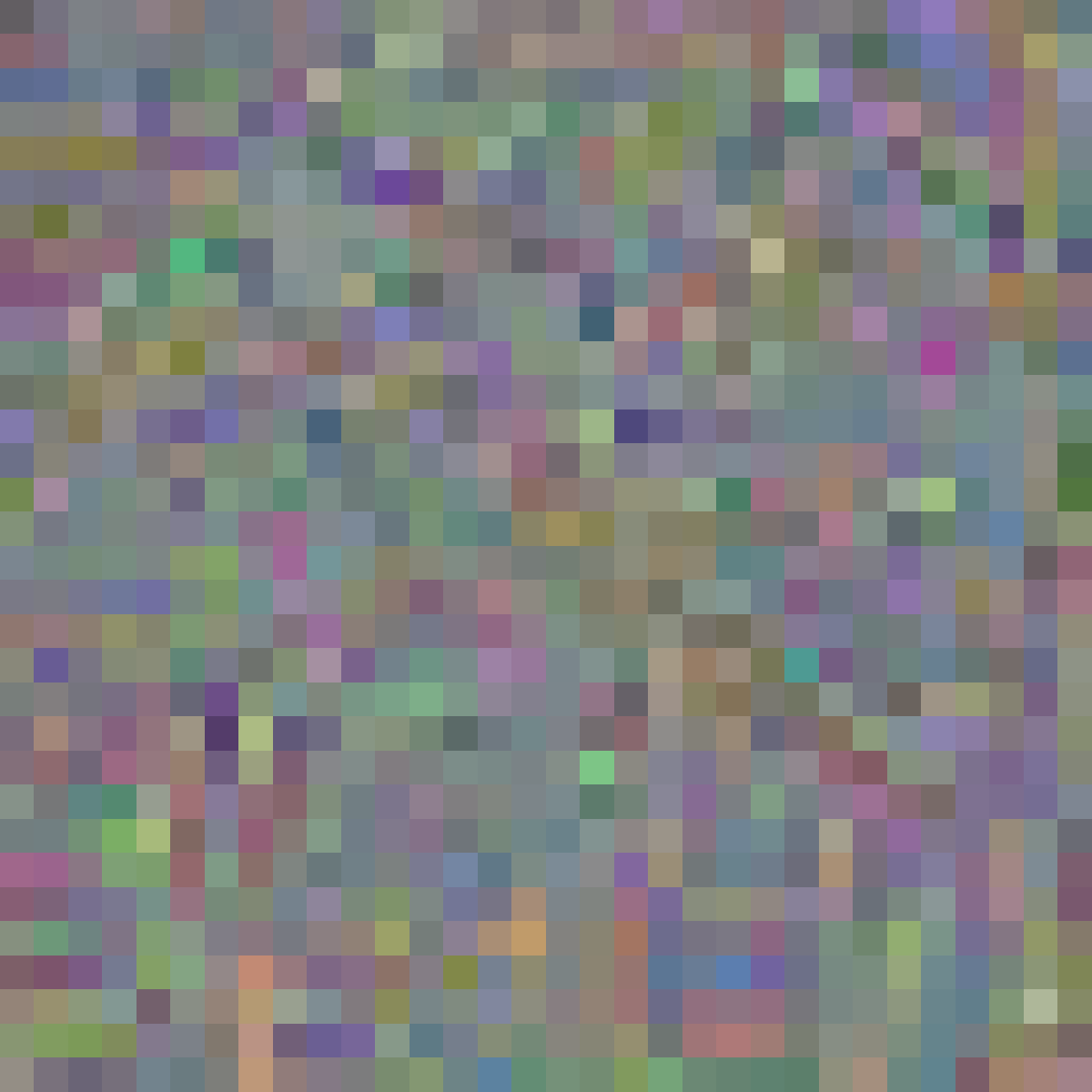} &
   \includegraphics[width=\linewidth]{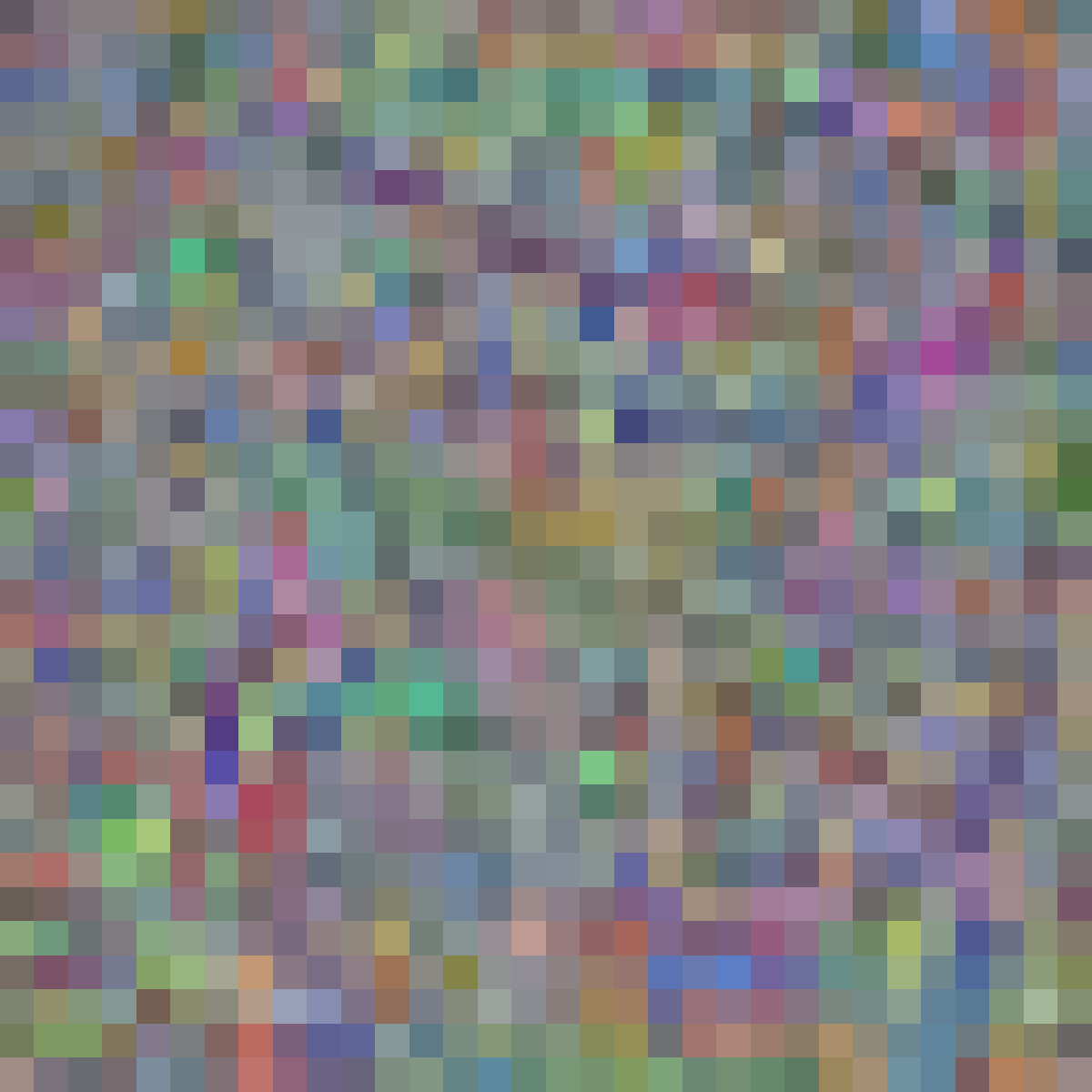} &
   \includegraphics[width=\linewidth]{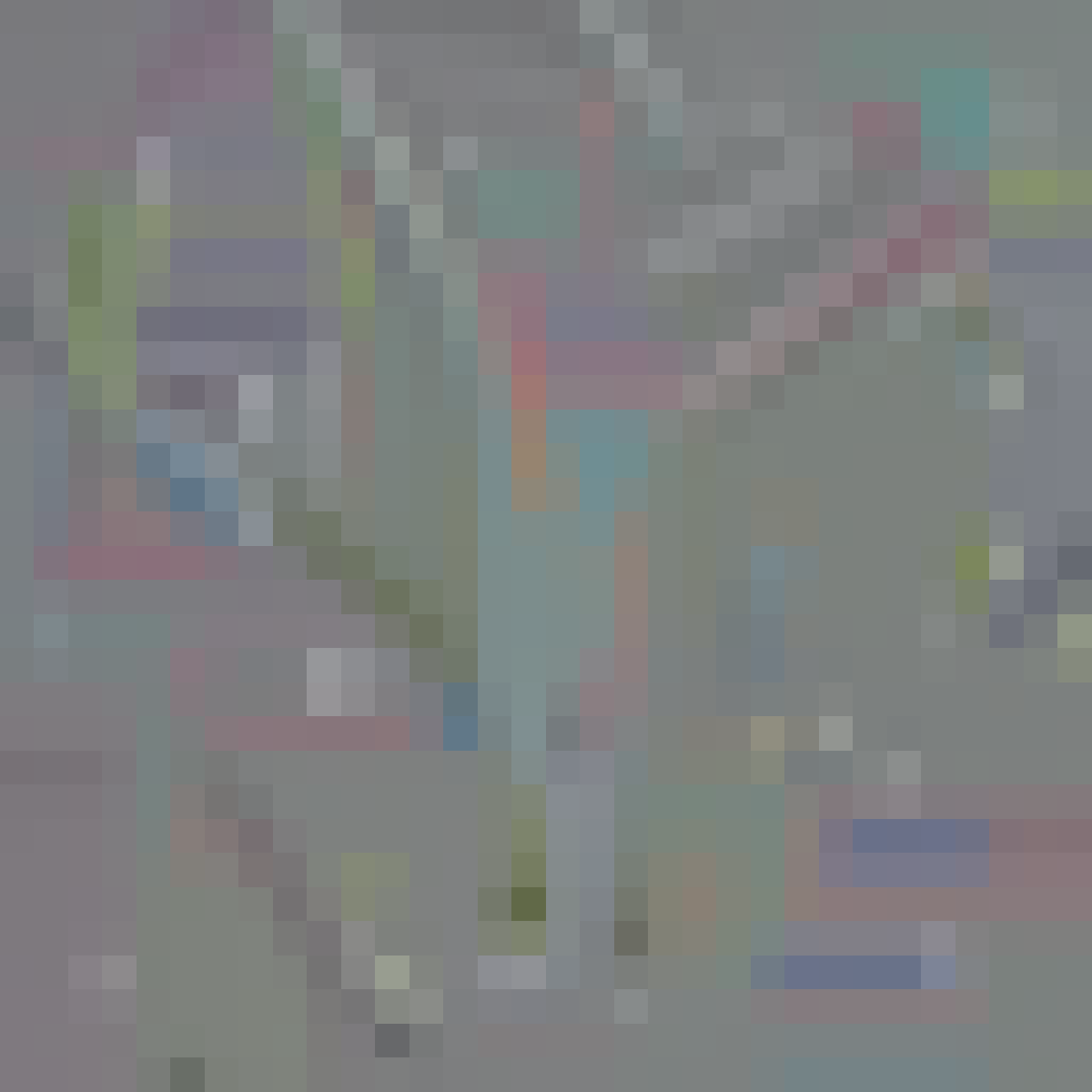} \\
   \includegraphics[width=\linewidth]{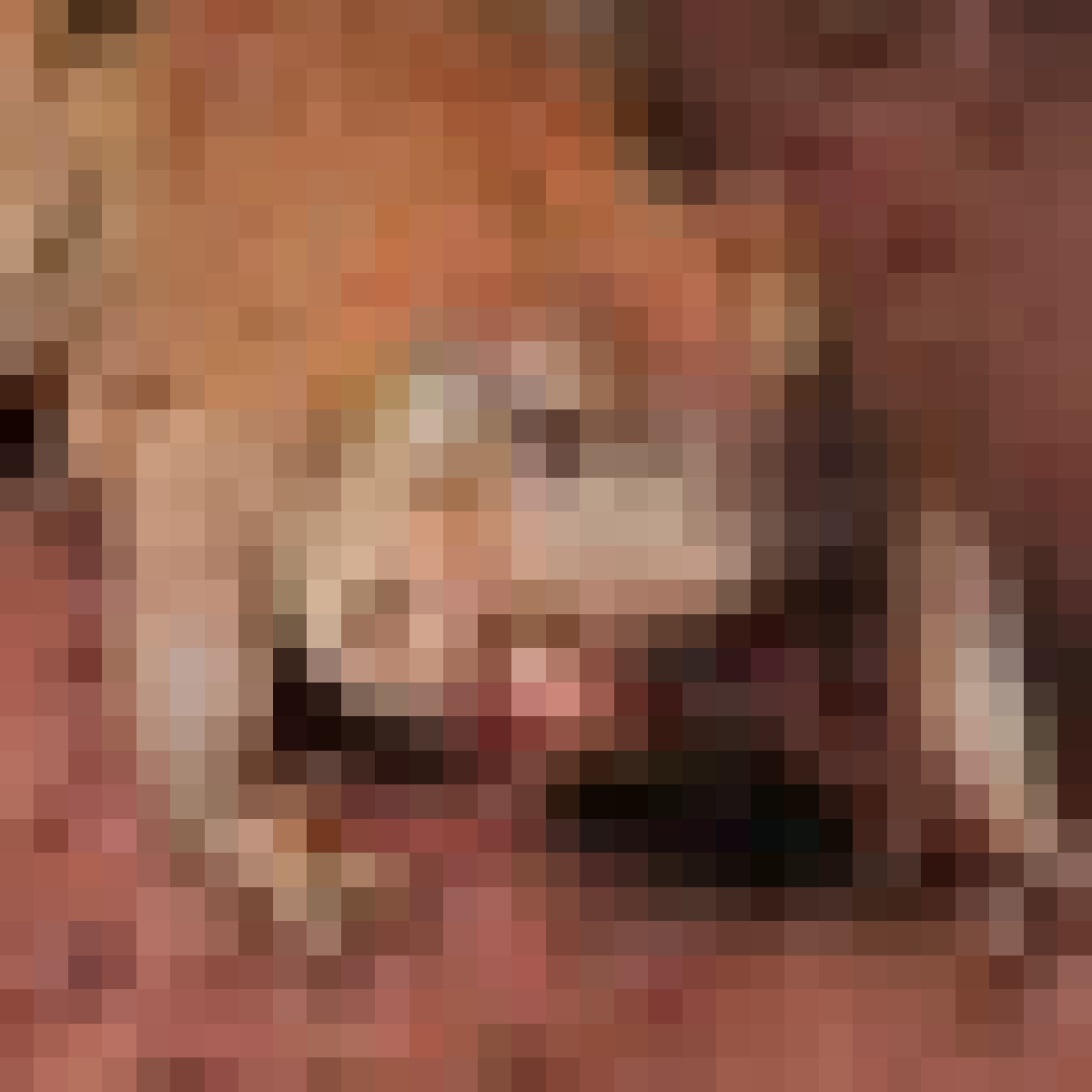} & 
   \includegraphics[width=\linewidth]{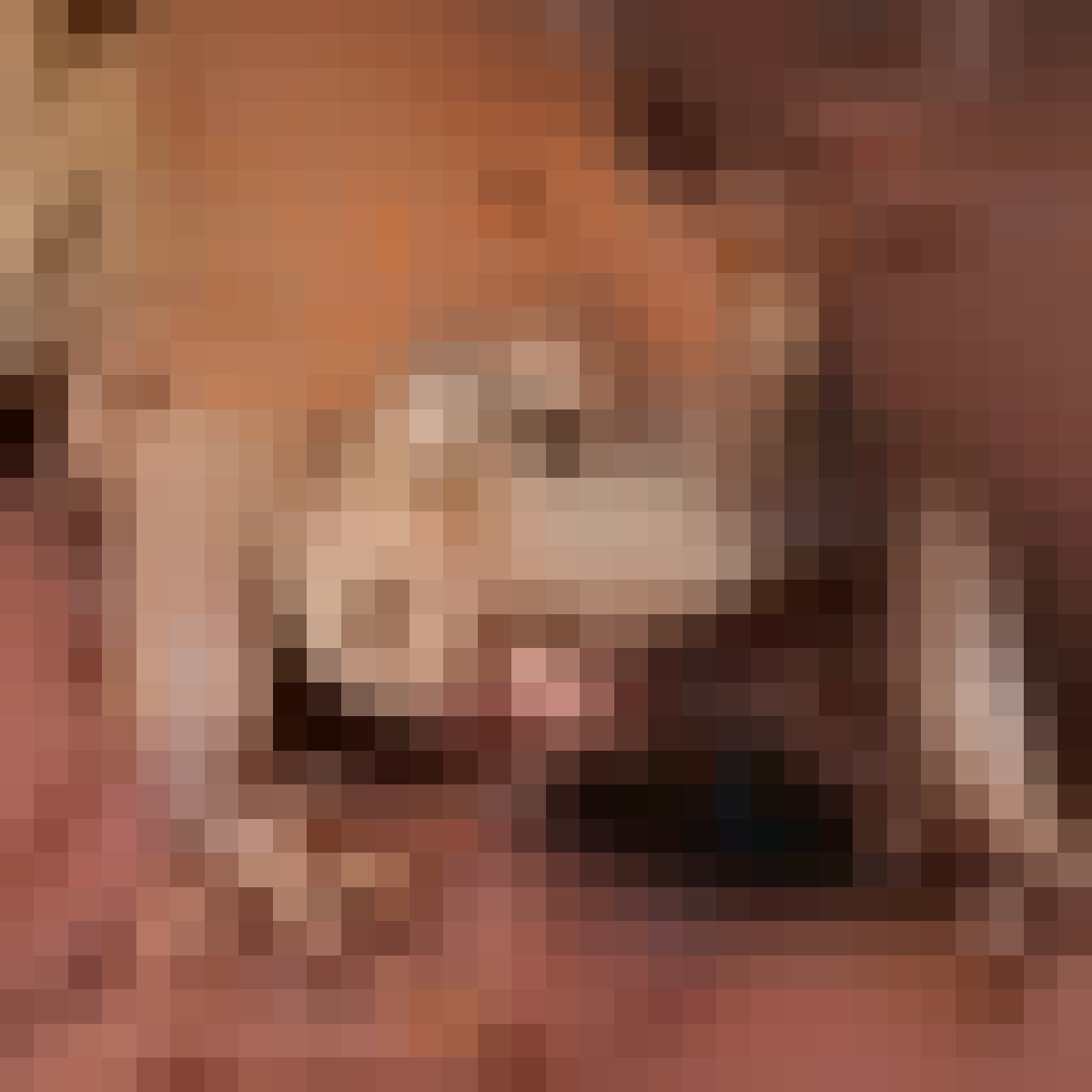} & 
   \includegraphics[width=\linewidth]{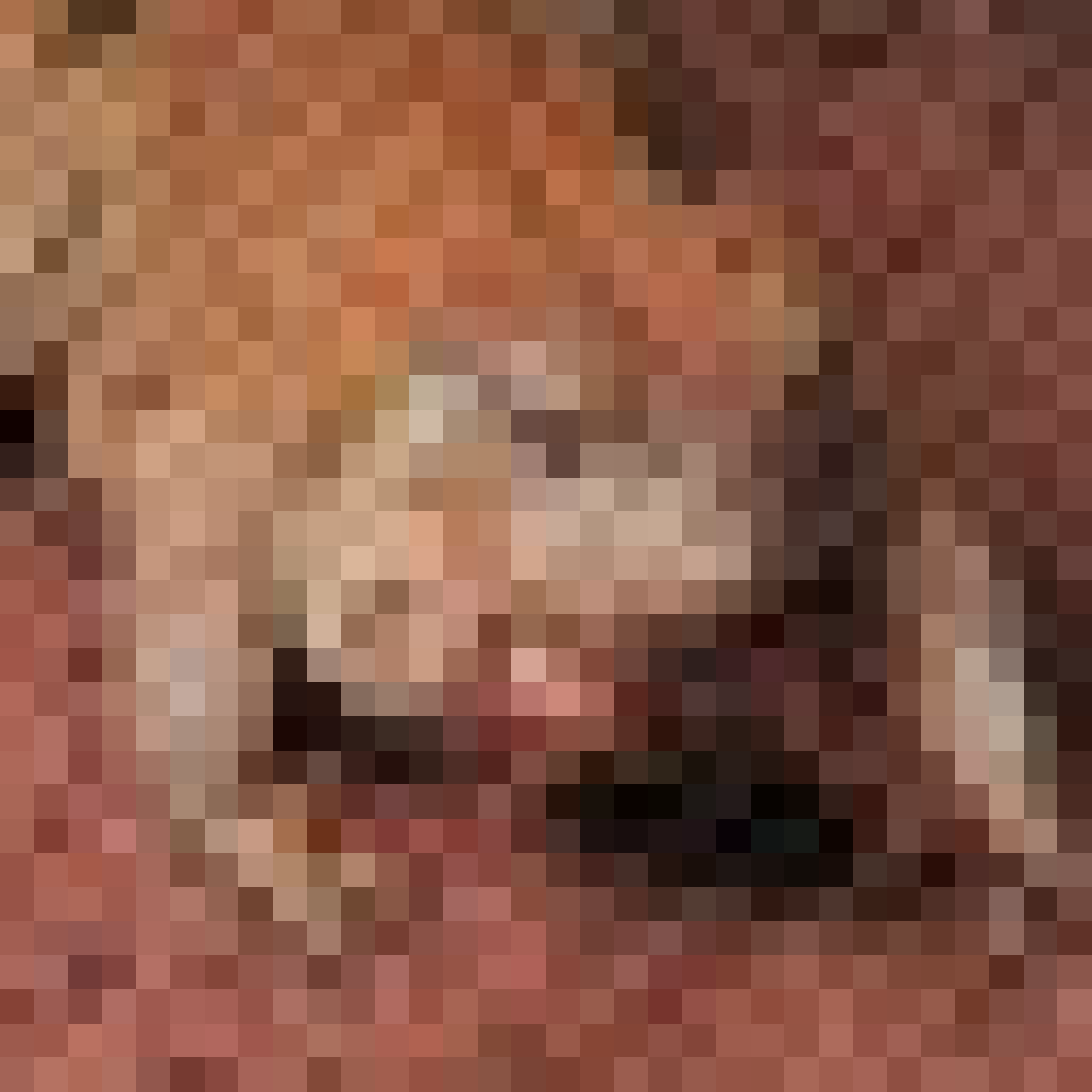} & 
   \includegraphics[width=\linewidth]{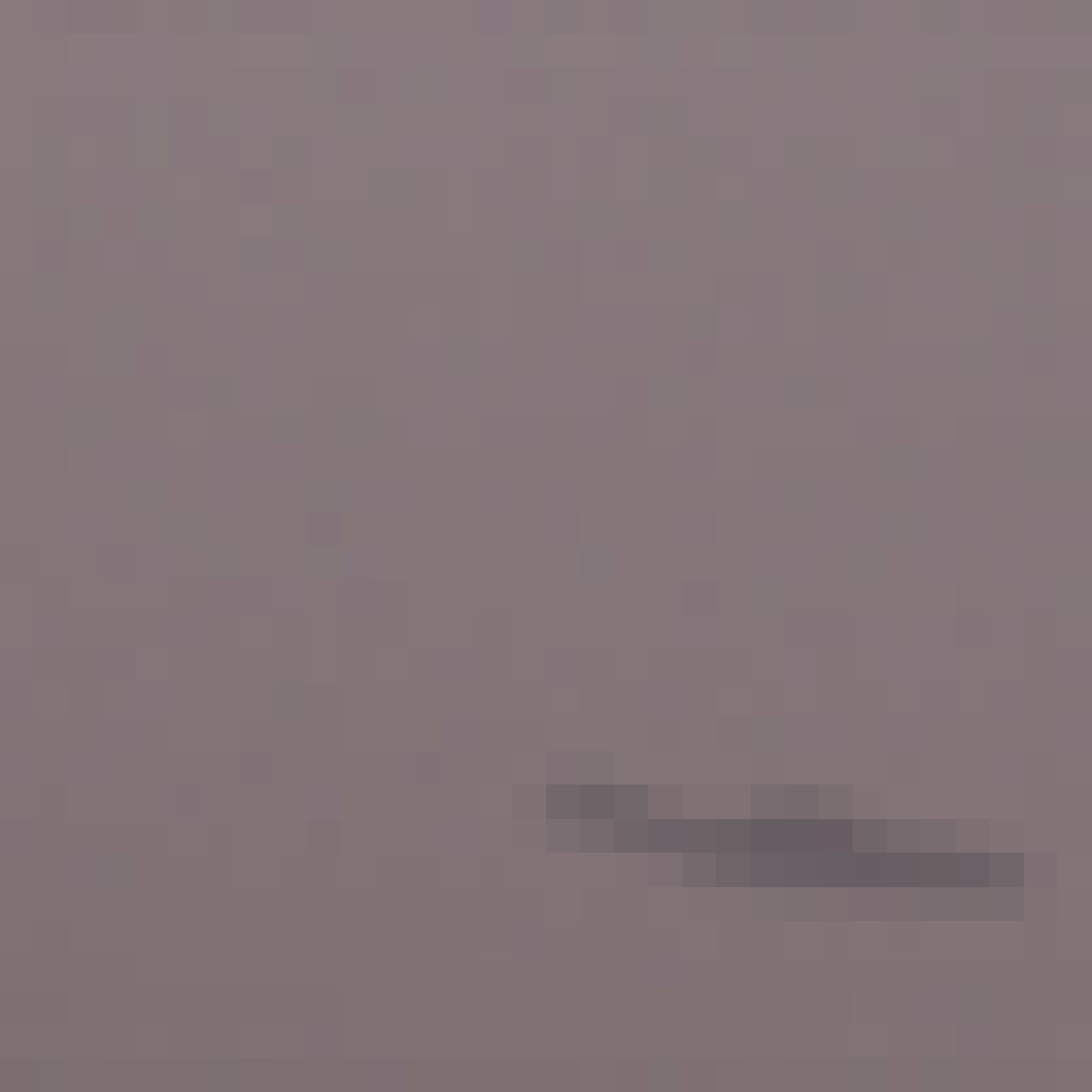} & 
   \includegraphics[width=\linewidth]{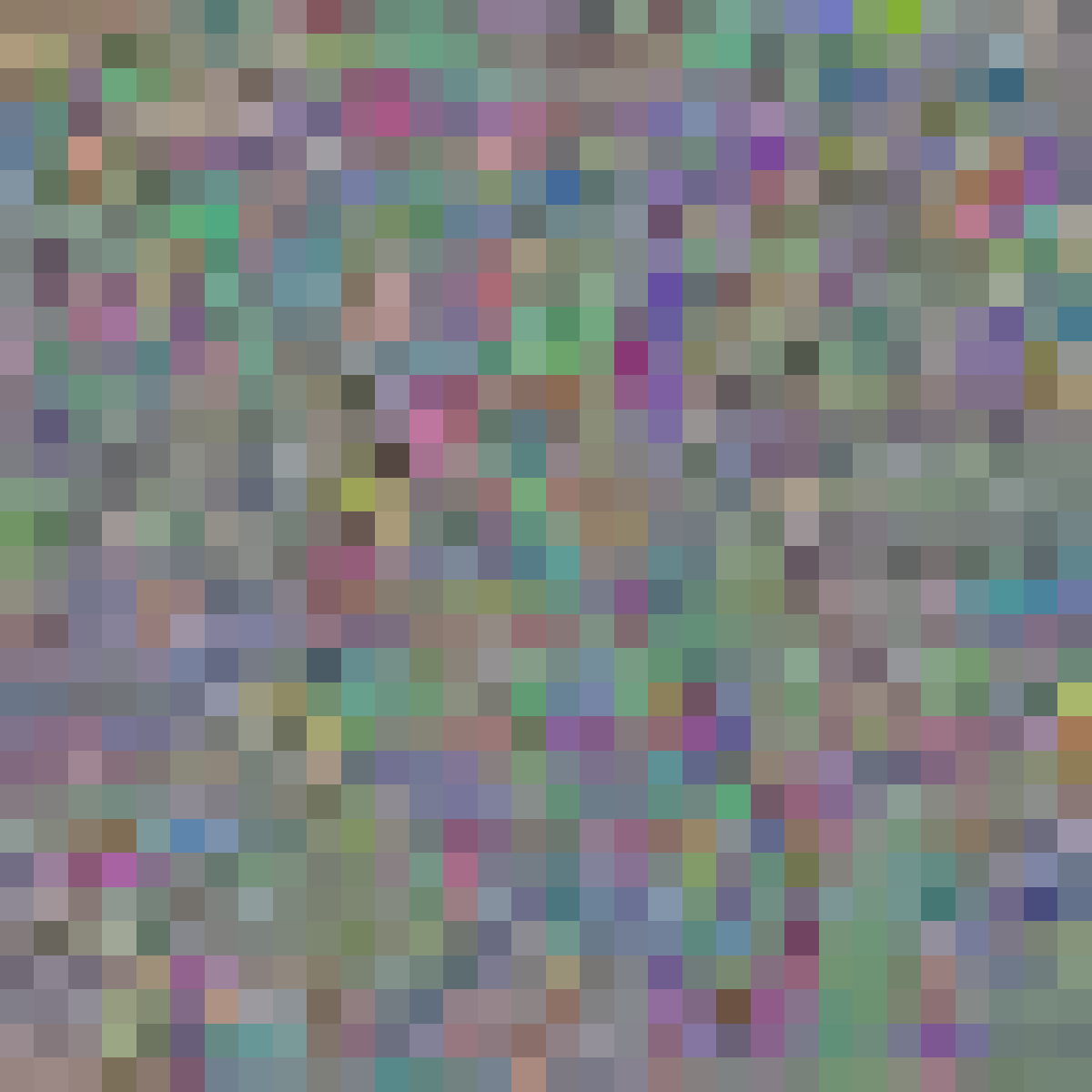} &
   \includegraphics[width=\linewidth]{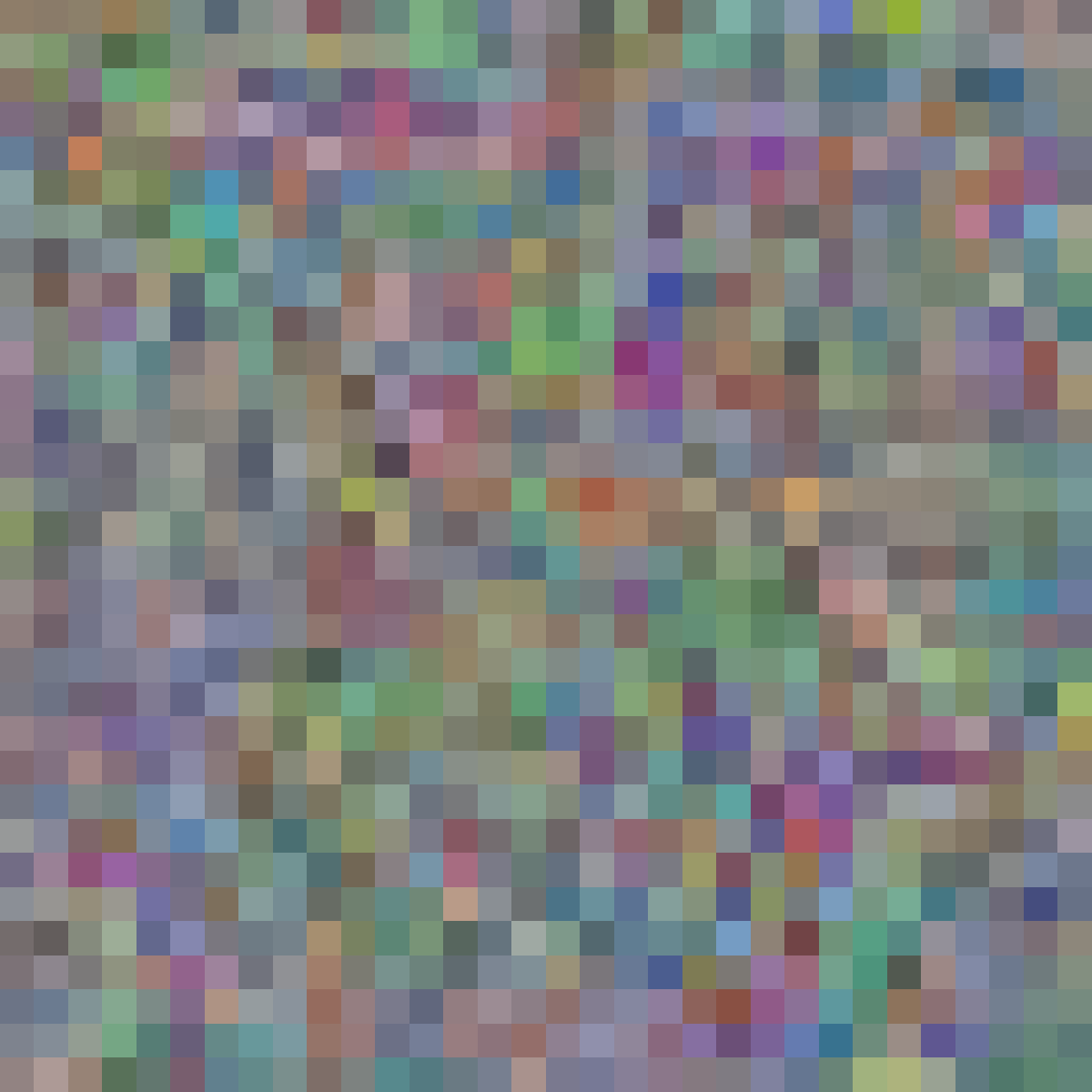} &
   \includegraphics[width=\linewidth]{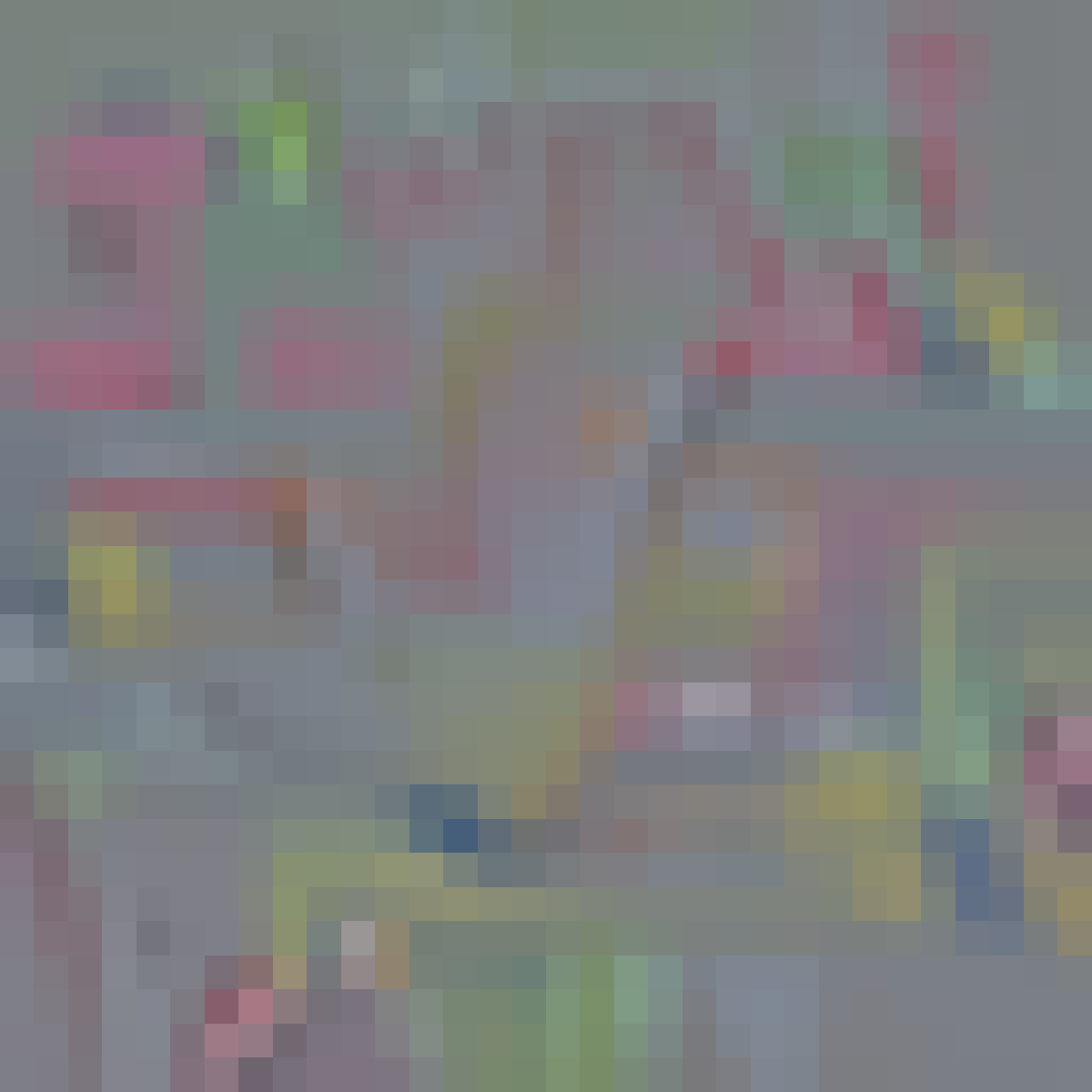} \\
   \includegraphics[width=\linewidth]{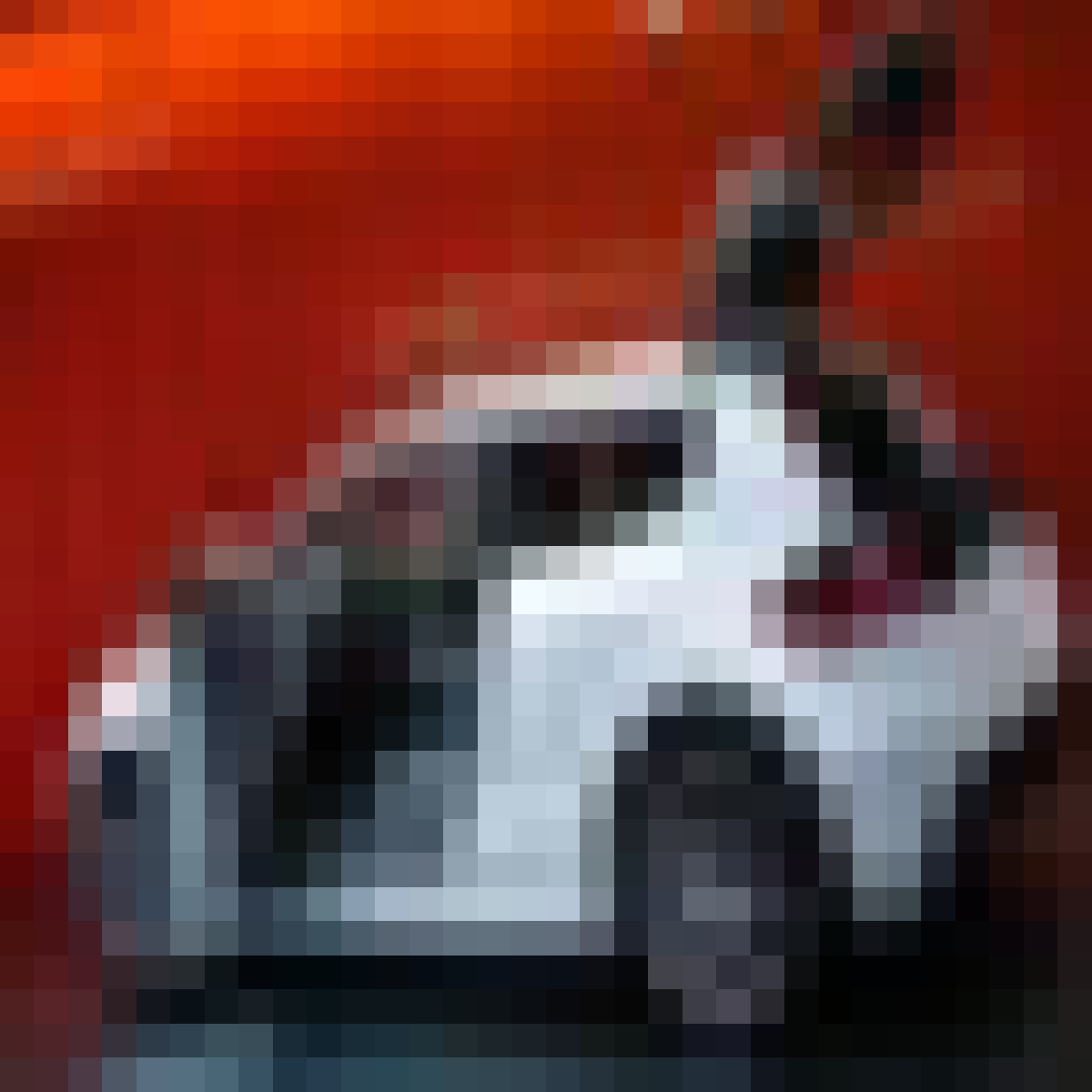} & 
   \includegraphics[width=\linewidth]{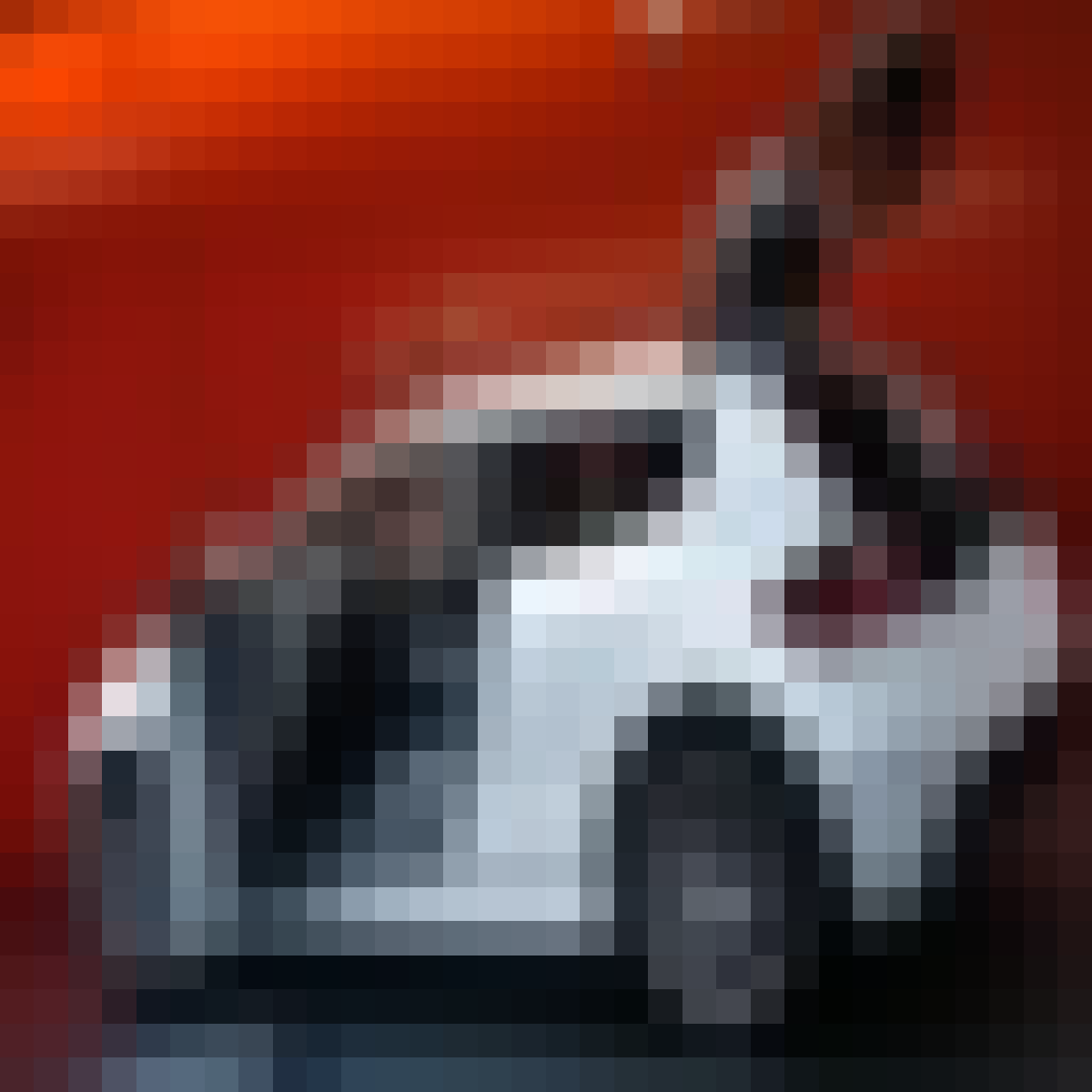} & 
   \includegraphics[width=\linewidth]{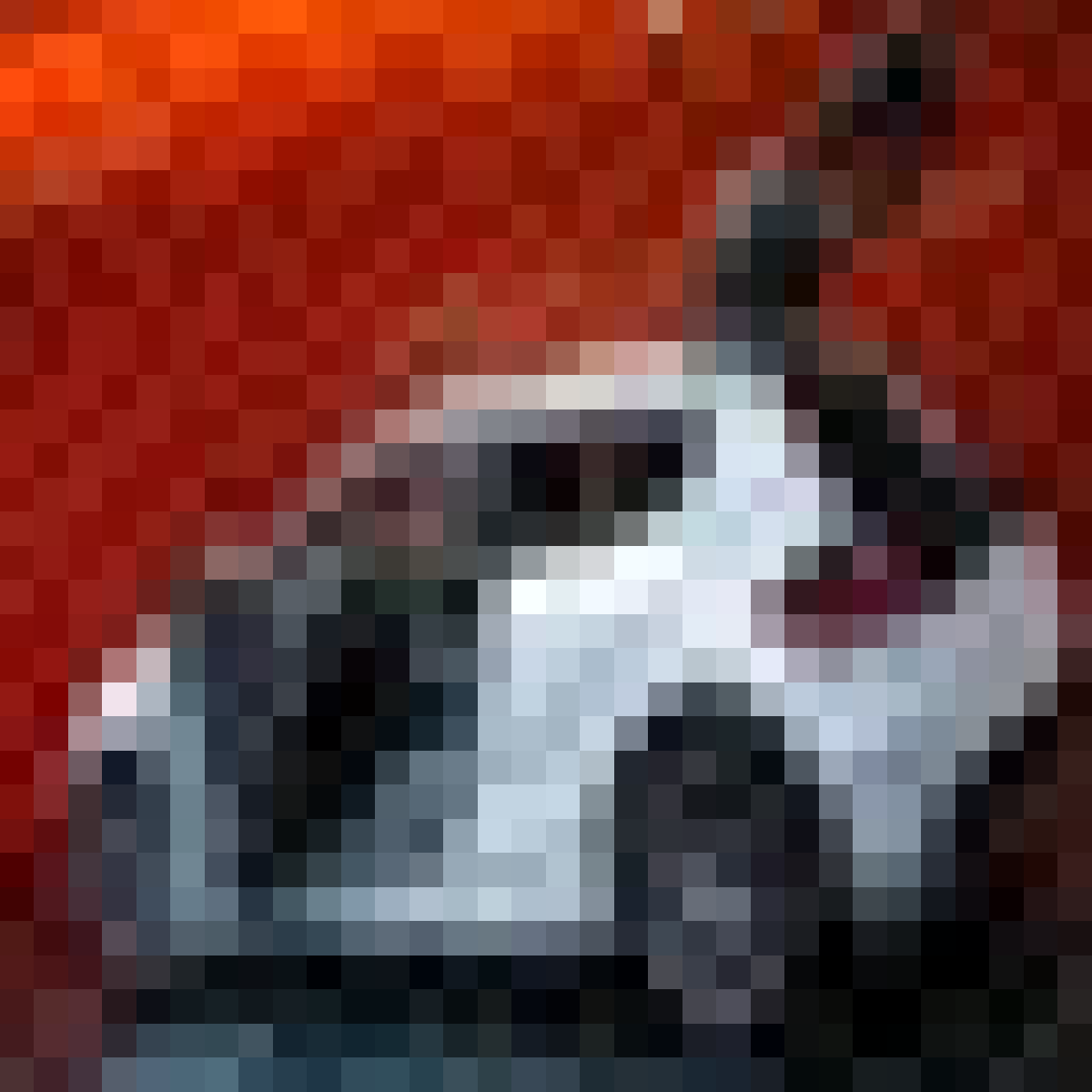} & 
   \includegraphics[width=\linewidth]{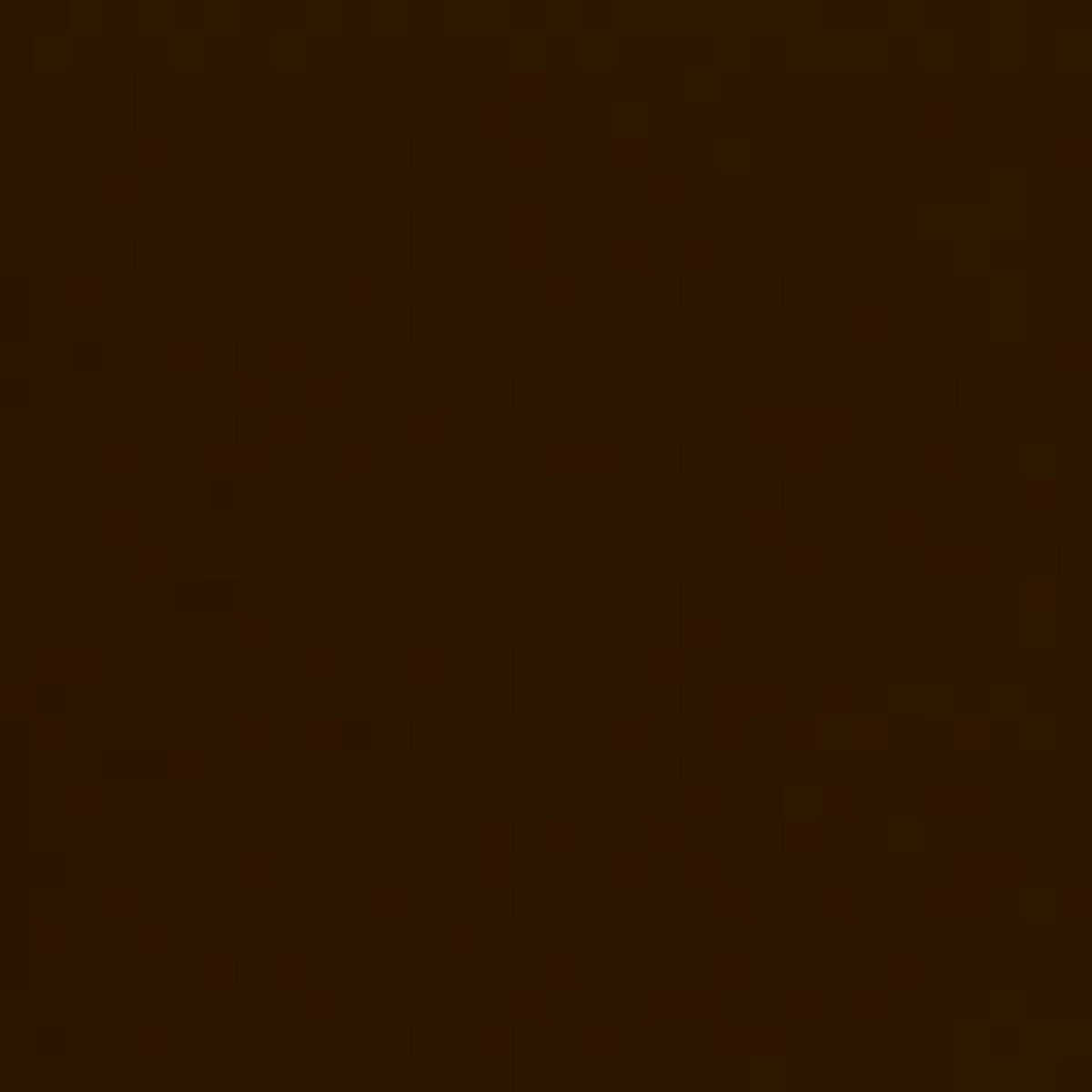} & 
   \includegraphics[width=\linewidth]{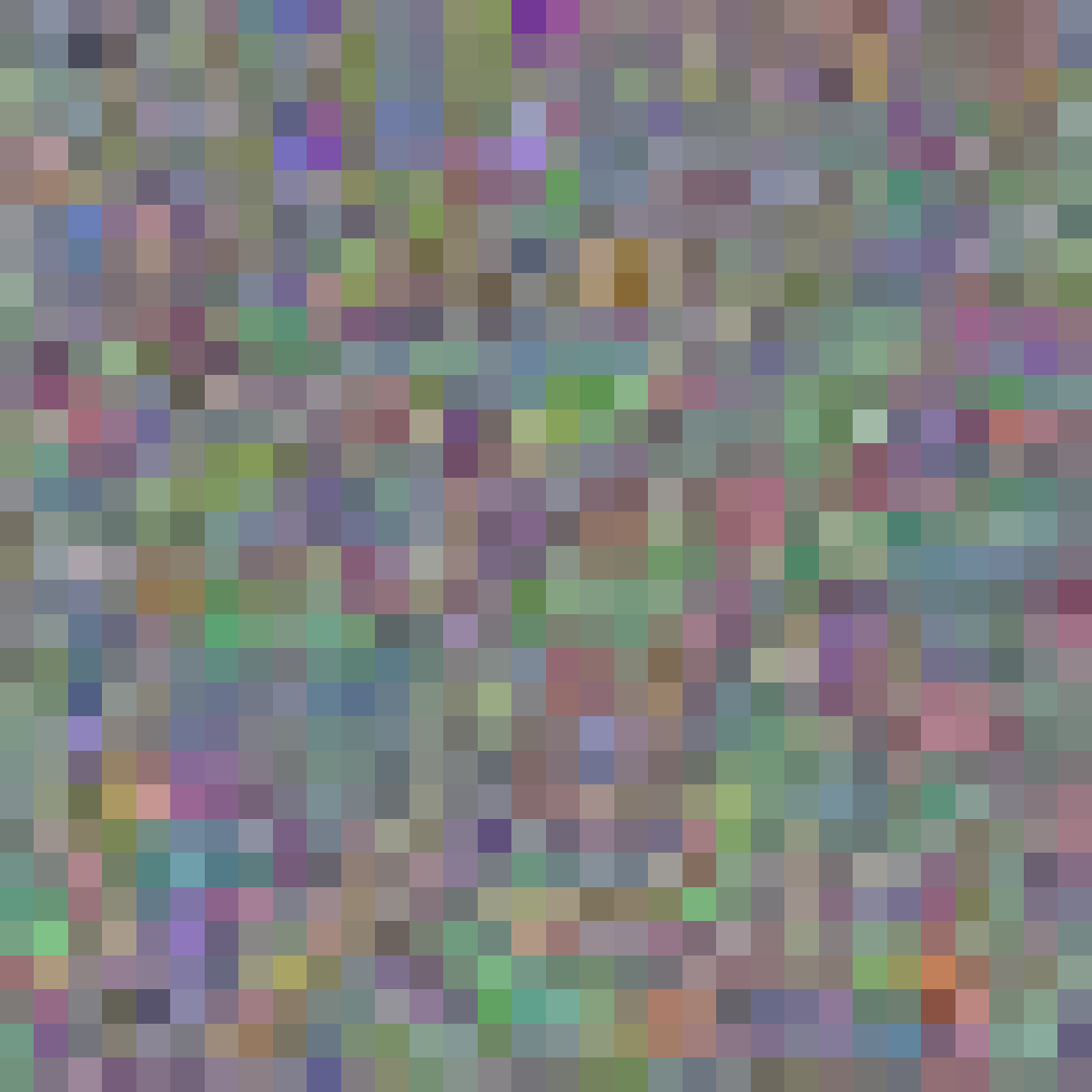} &
   \includegraphics[width=\linewidth]{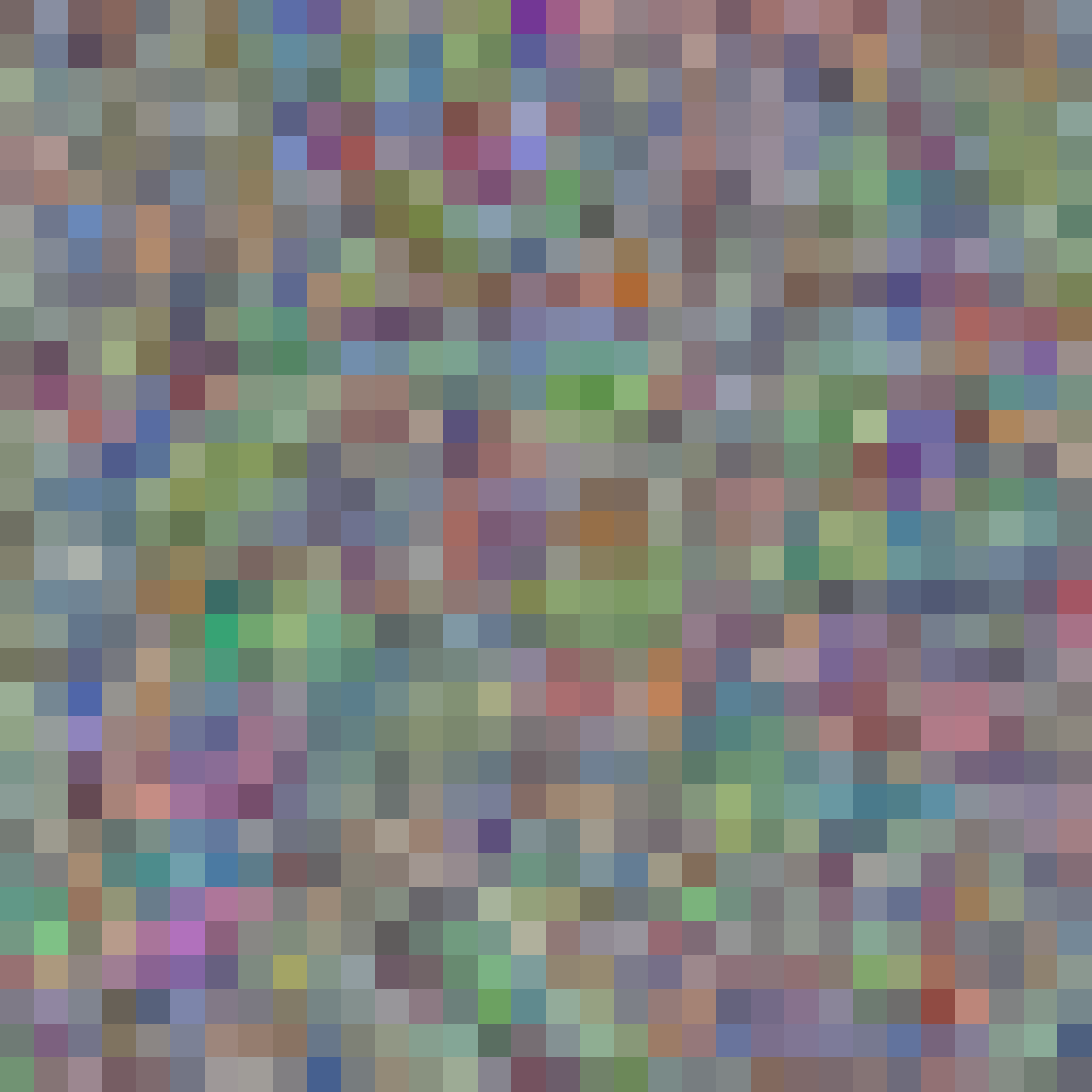} &
   \includegraphics[width=\linewidth]{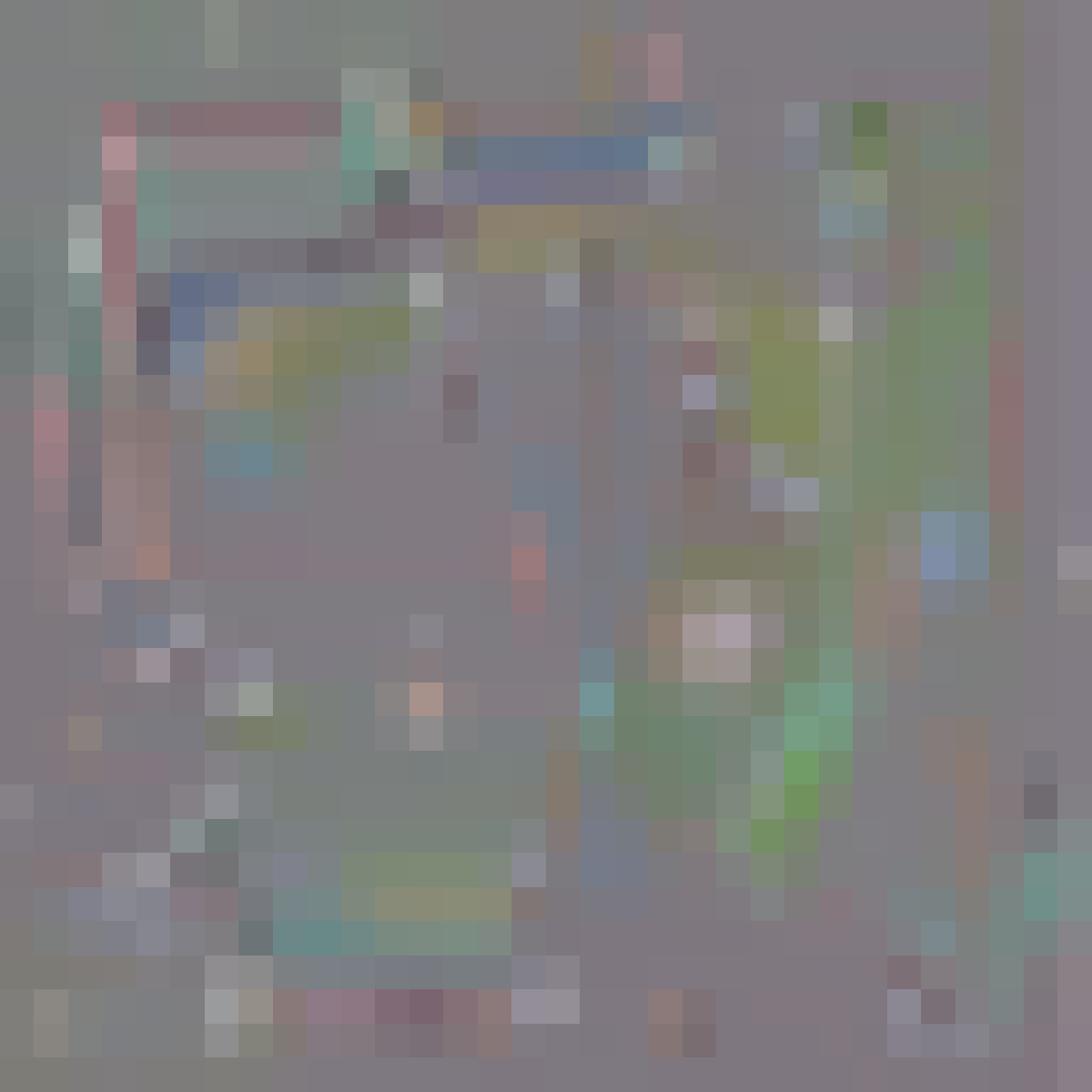} \\
   \includegraphics[width=\linewidth]{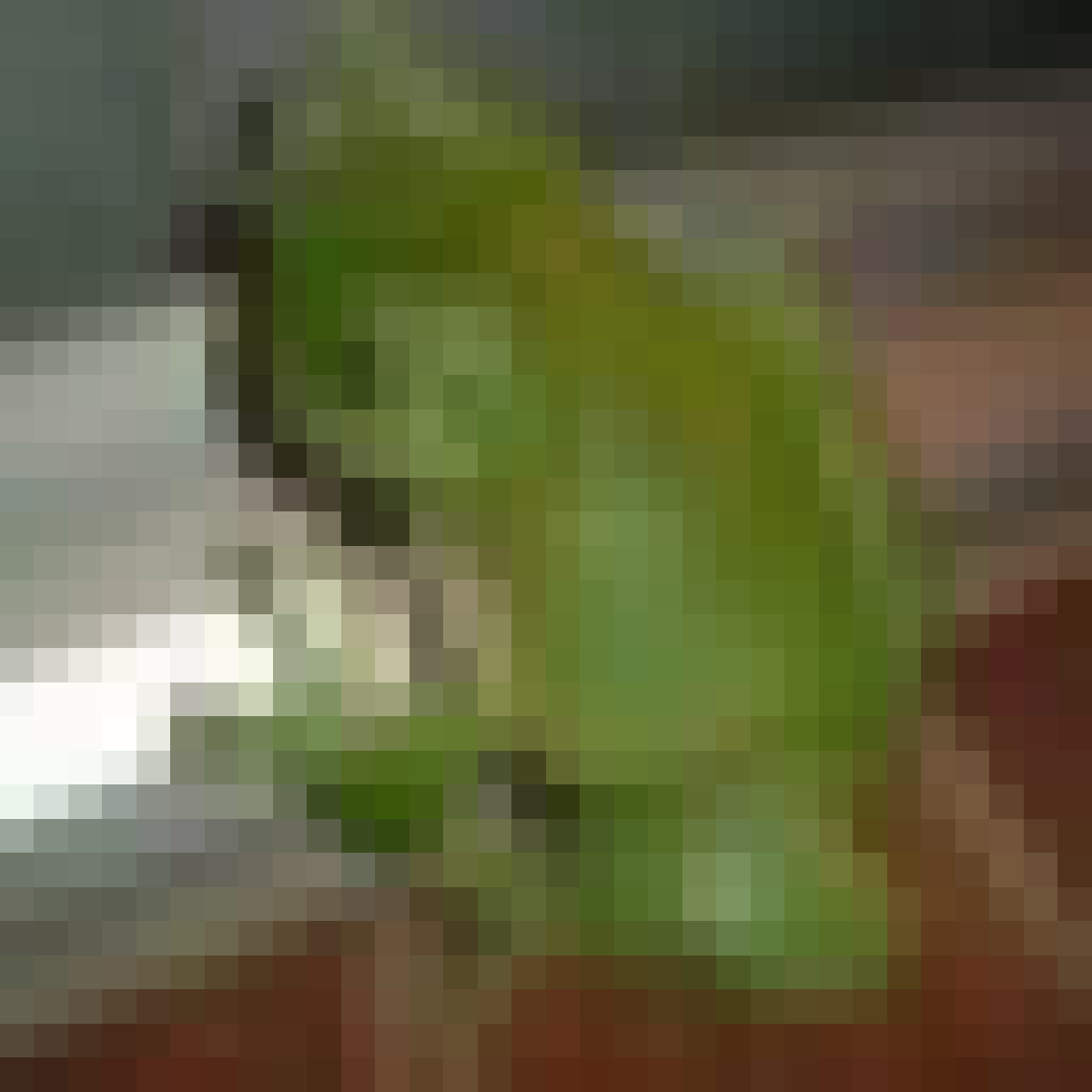} & 
   \includegraphics[width=\linewidth]{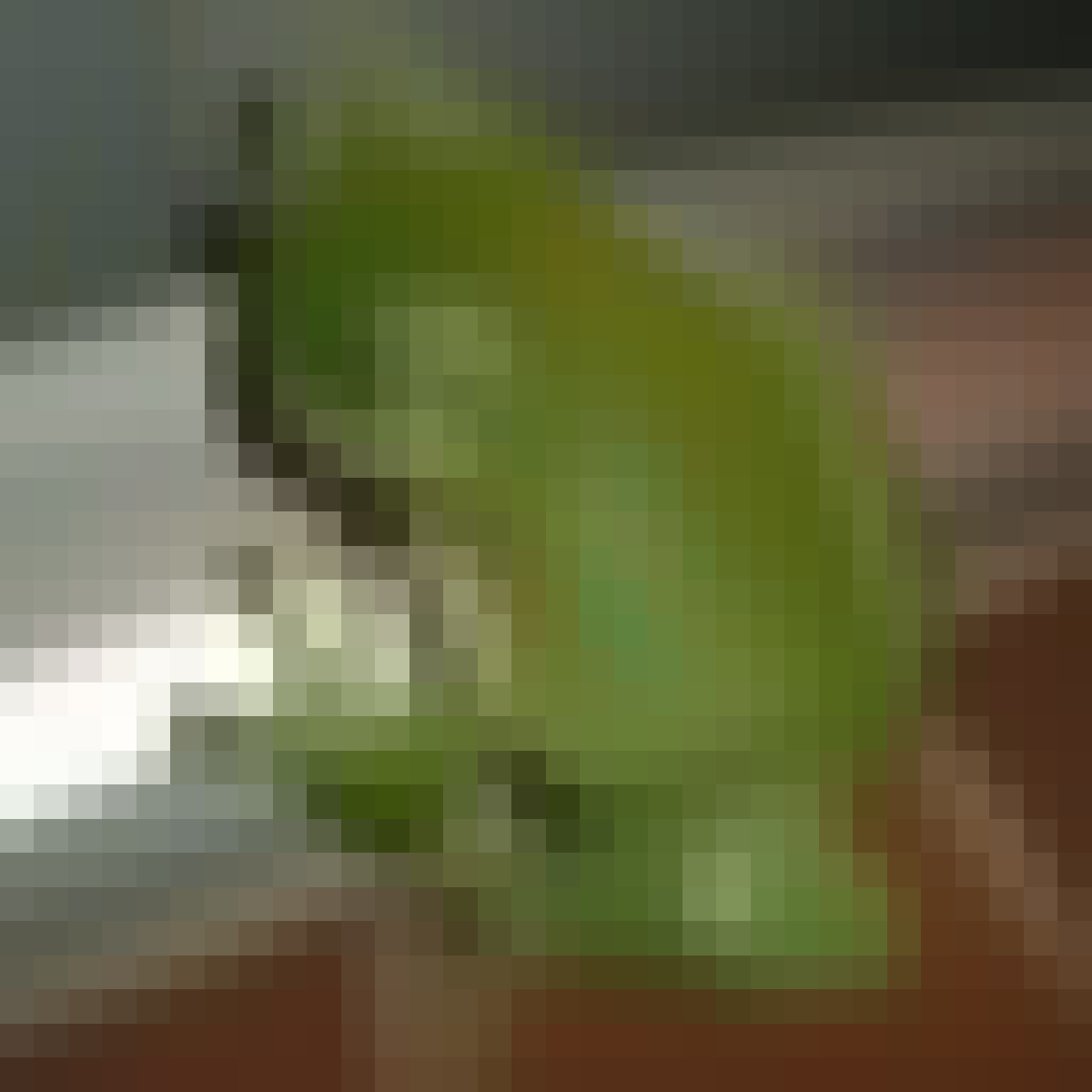} & 
   \includegraphics[width=\linewidth]{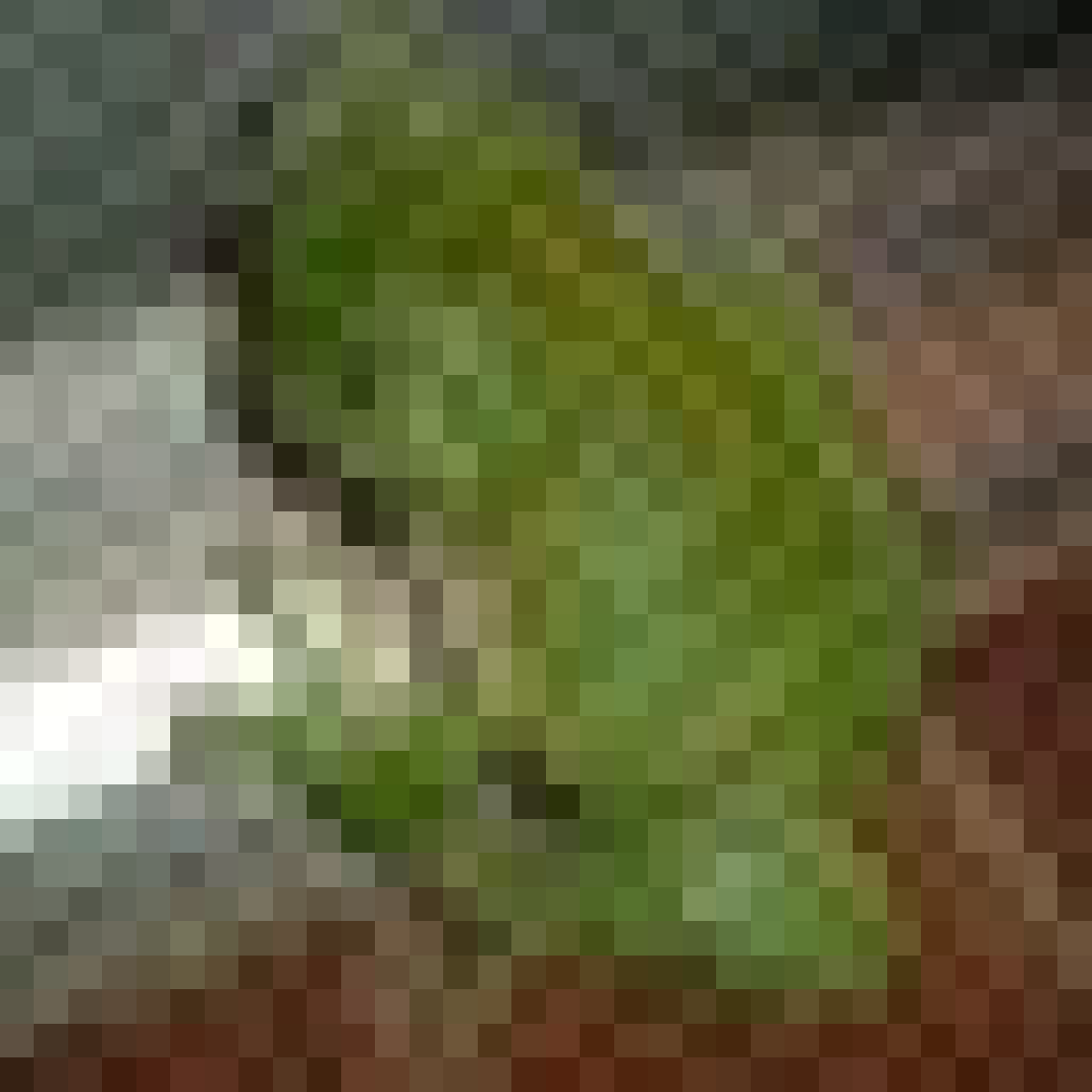} & 
   \includegraphics[width=\linewidth]{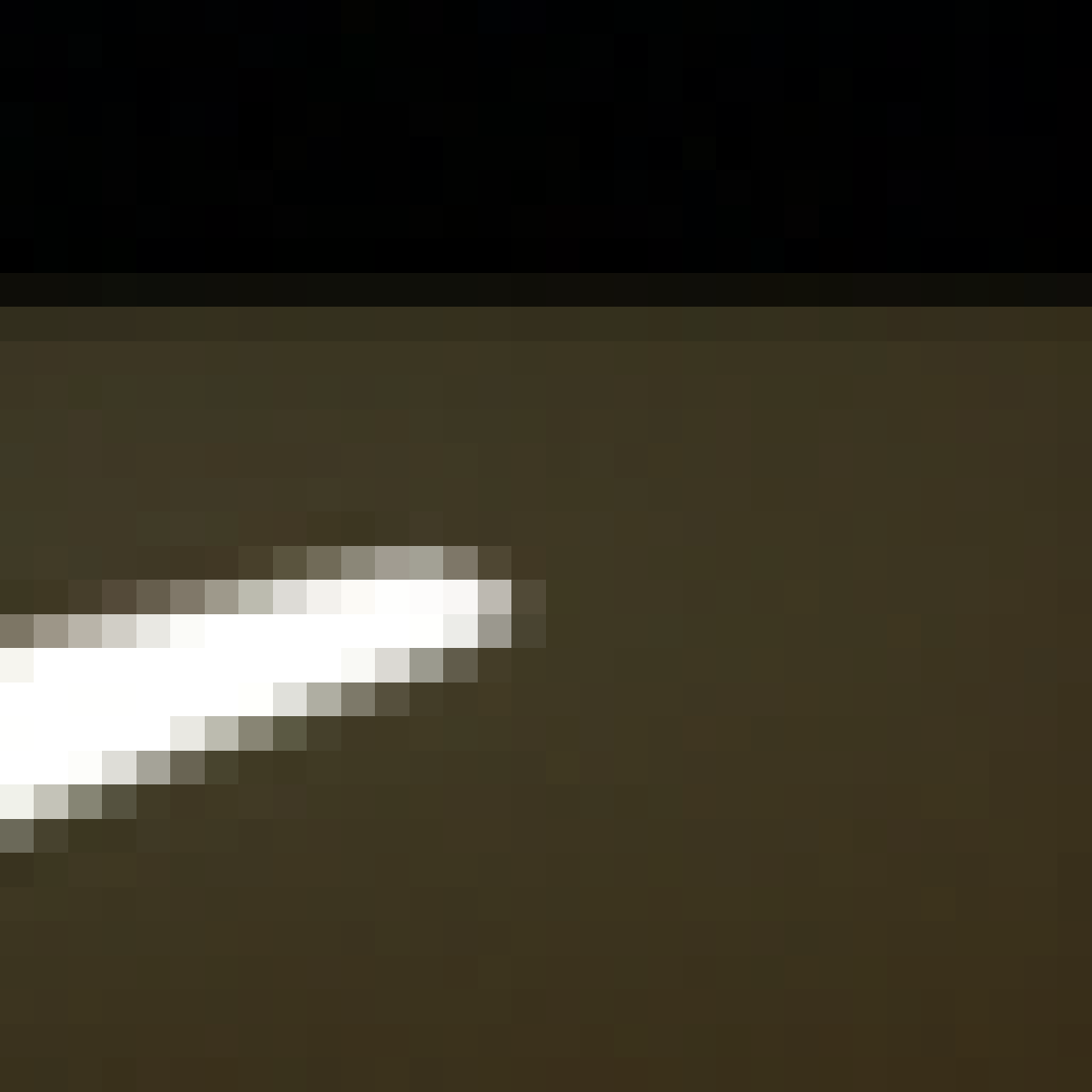} & 
   \includegraphics[width=\linewidth]{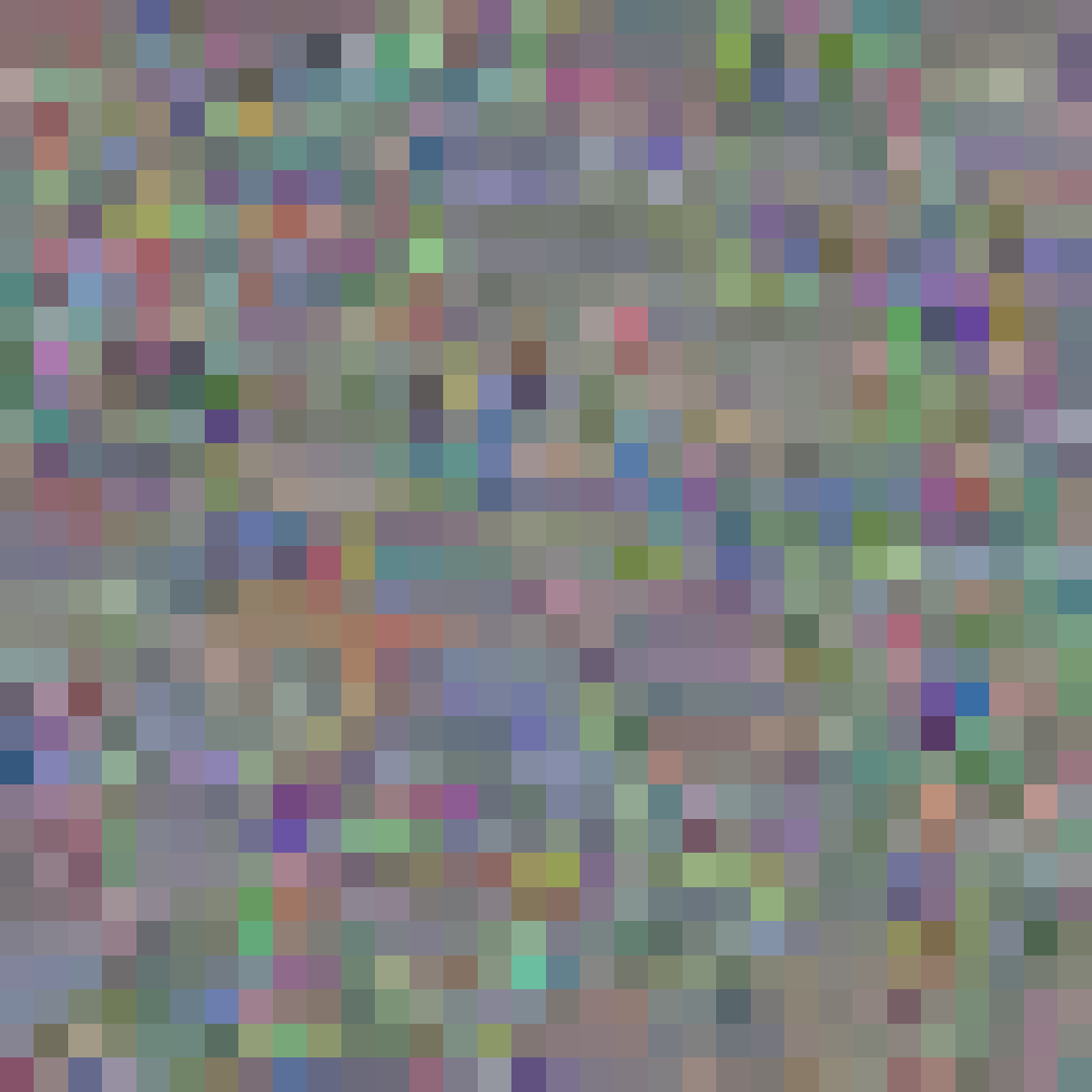} &
   \includegraphics[width=\linewidth]{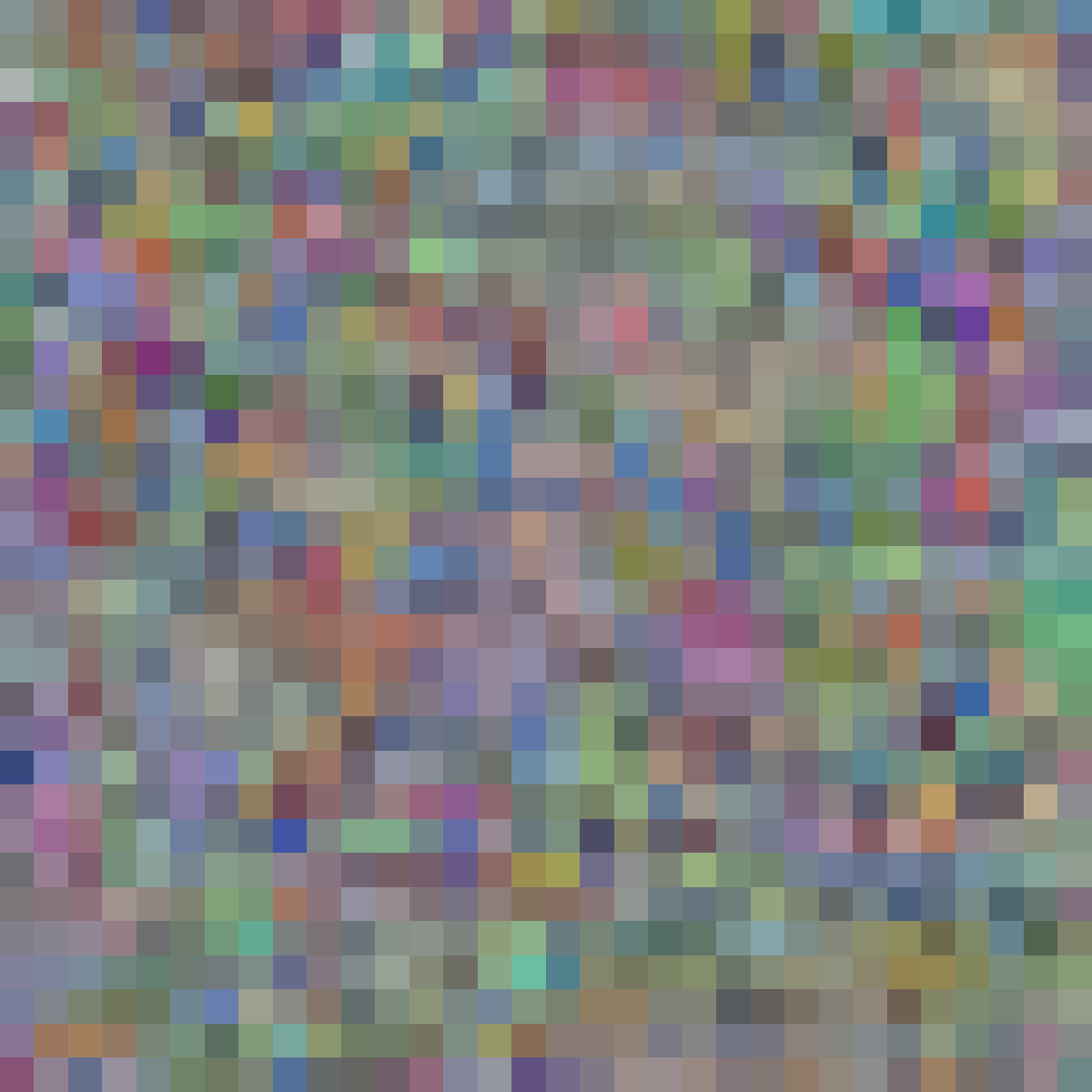} &
   \includegraphics[width=\linewidth]{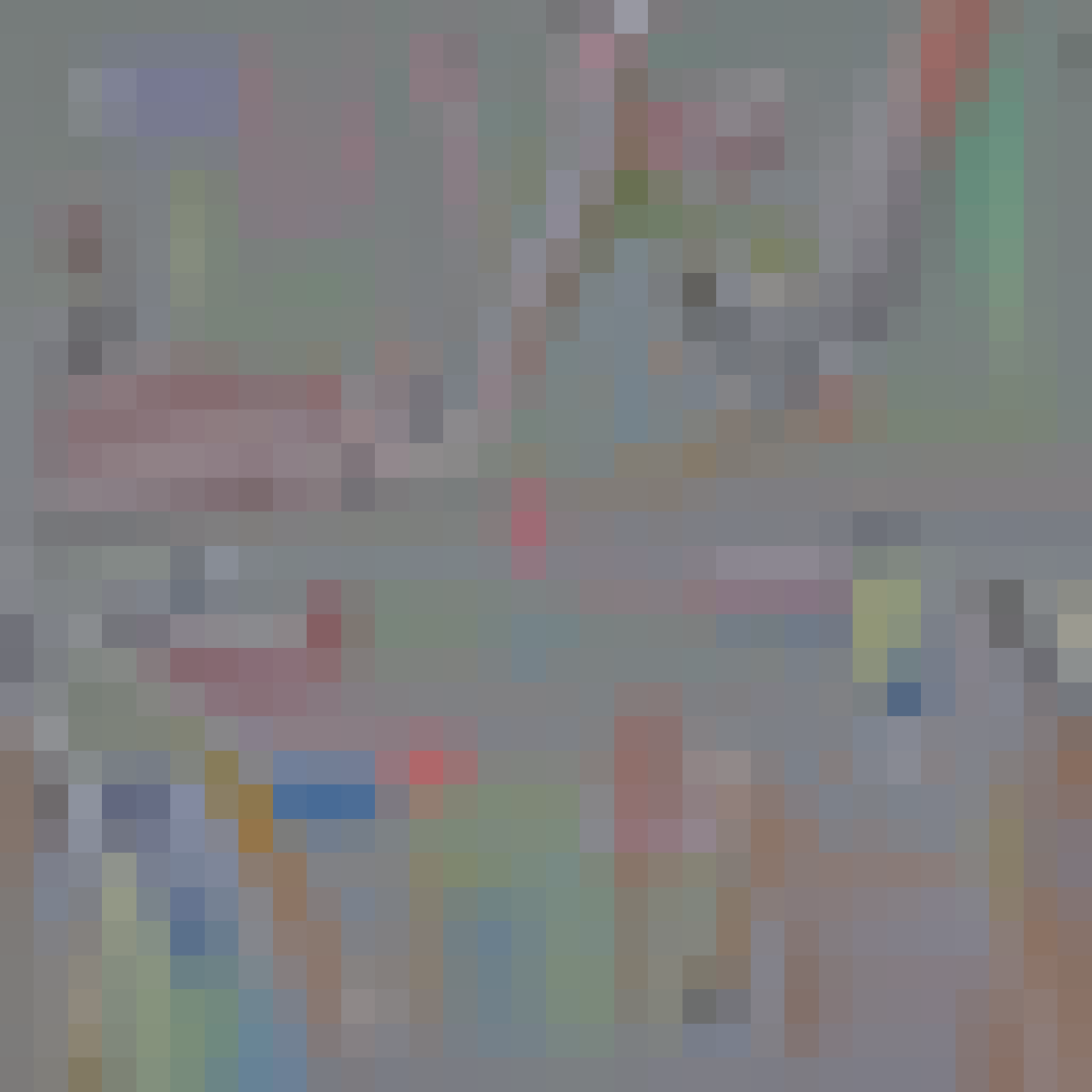} \\
   \includegraphics[width=\linewidth]{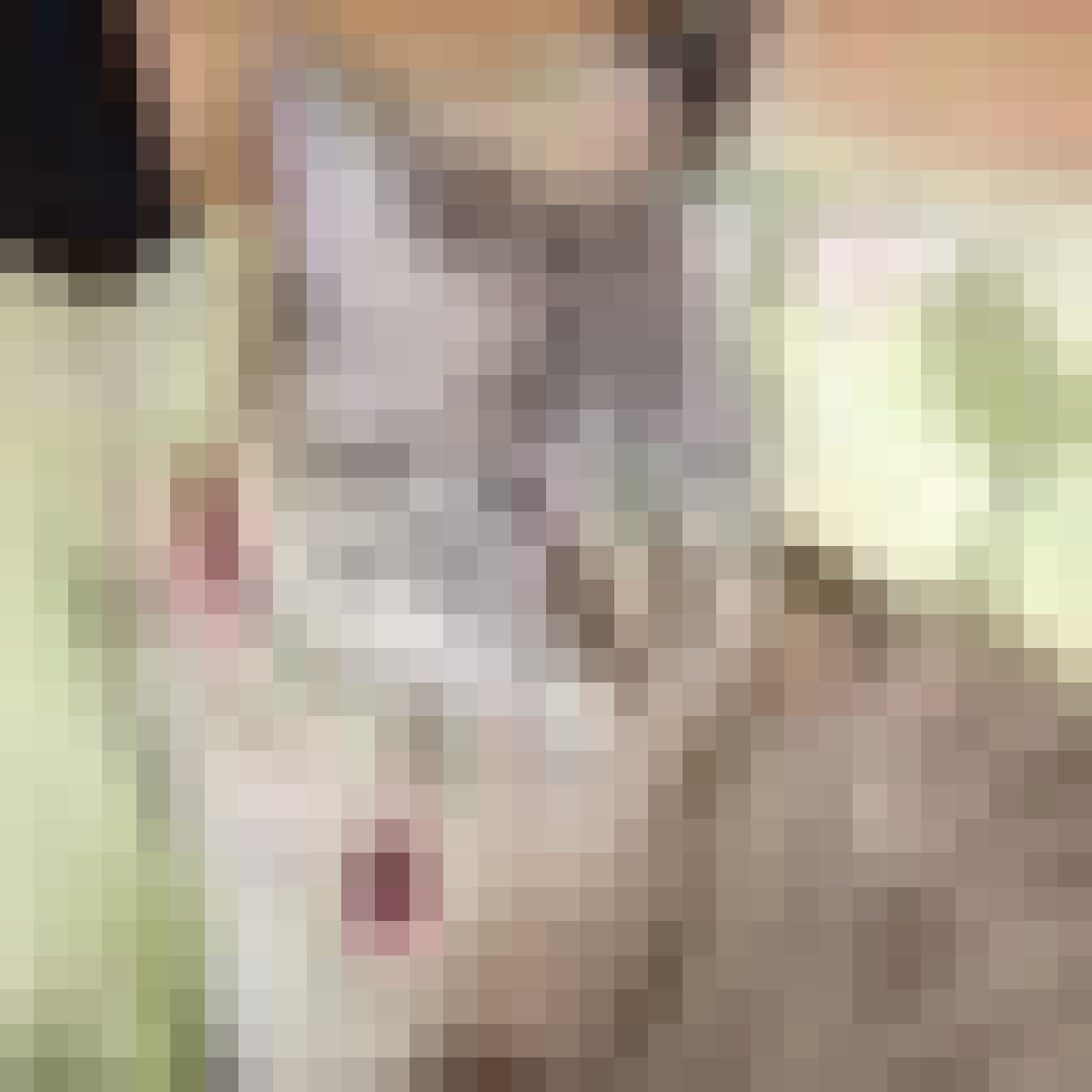} & 
   \includegraphics[width=\linewidth]{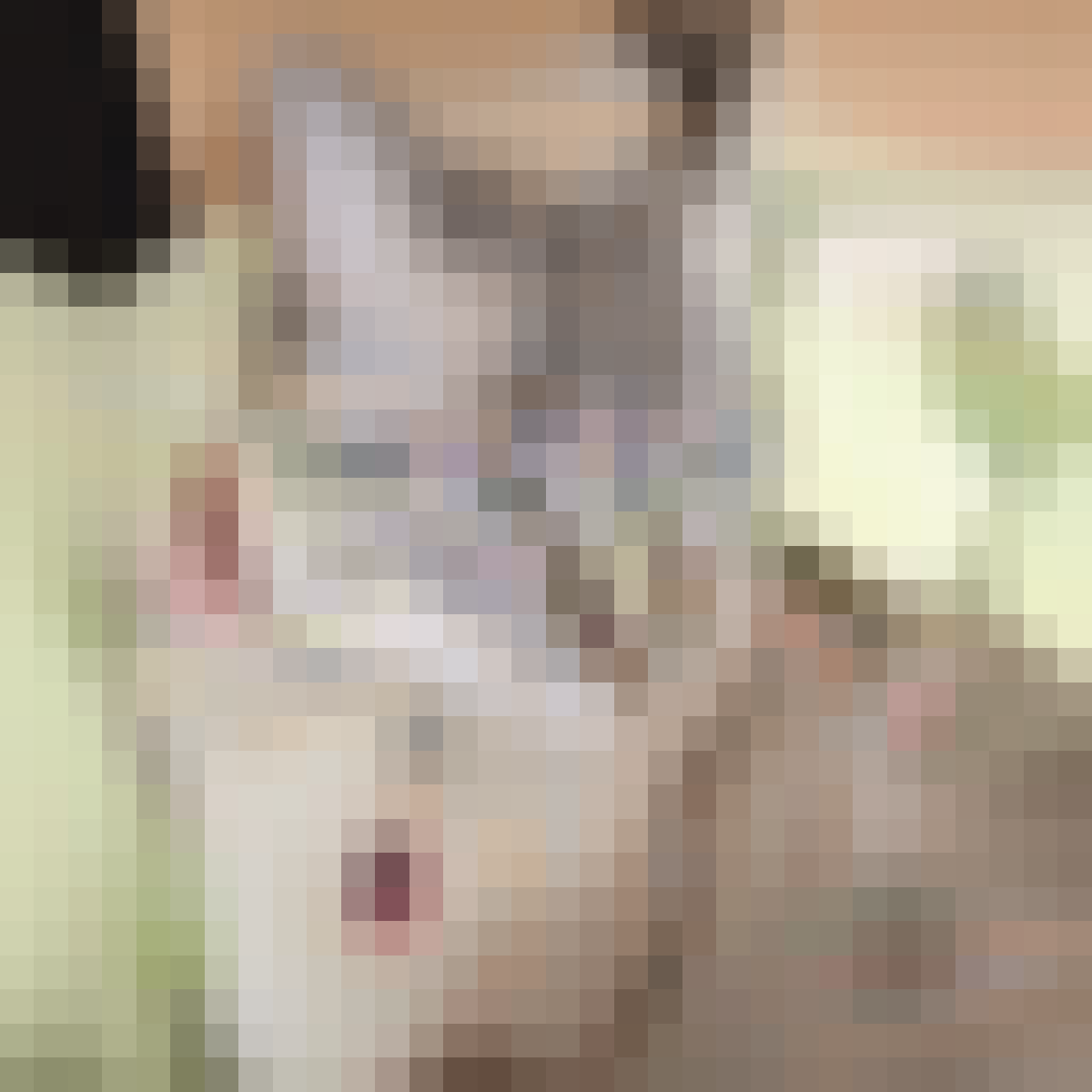} & 
   \includegraphics[width=\linewidth]{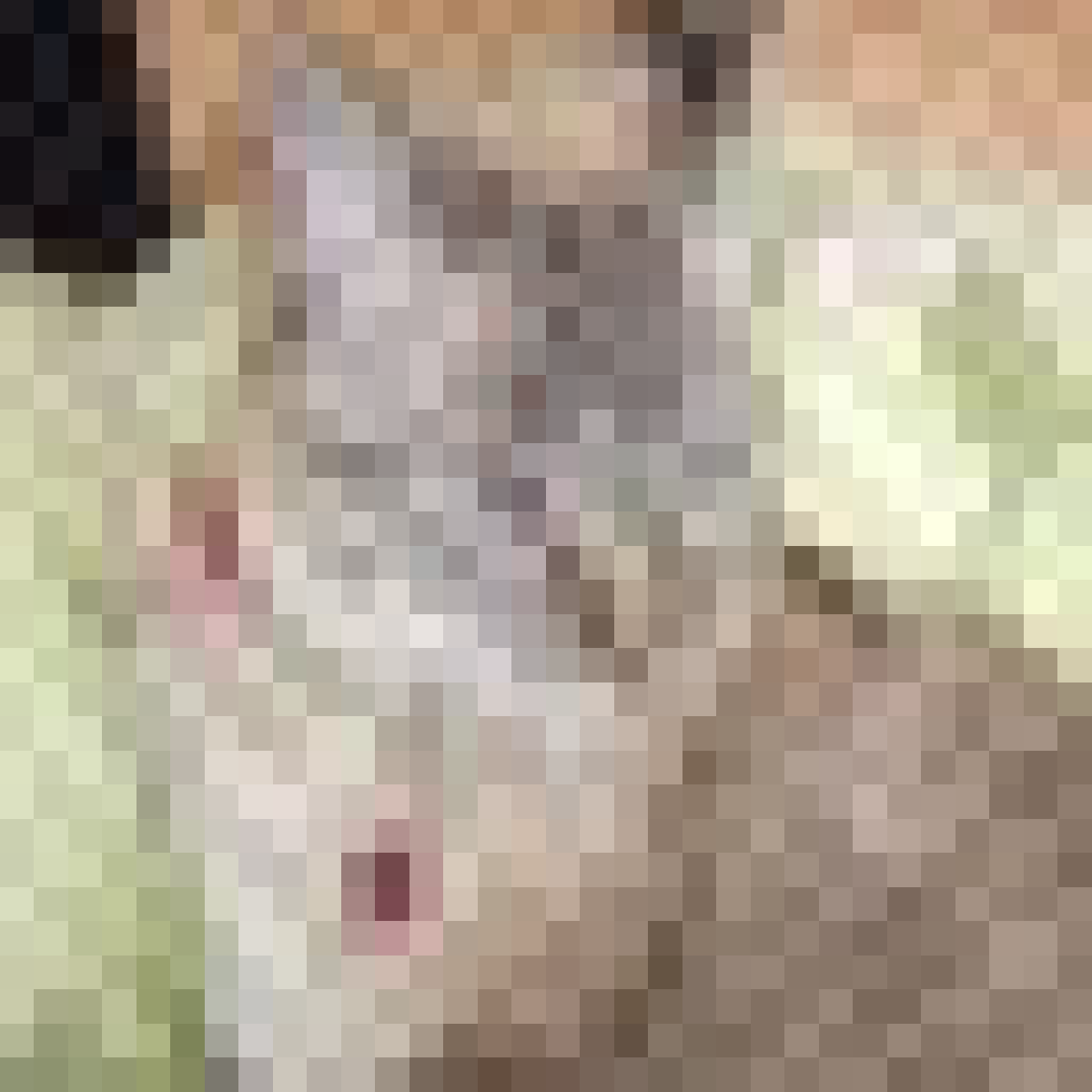} & 
   \includegraphics[width=\linewidth]{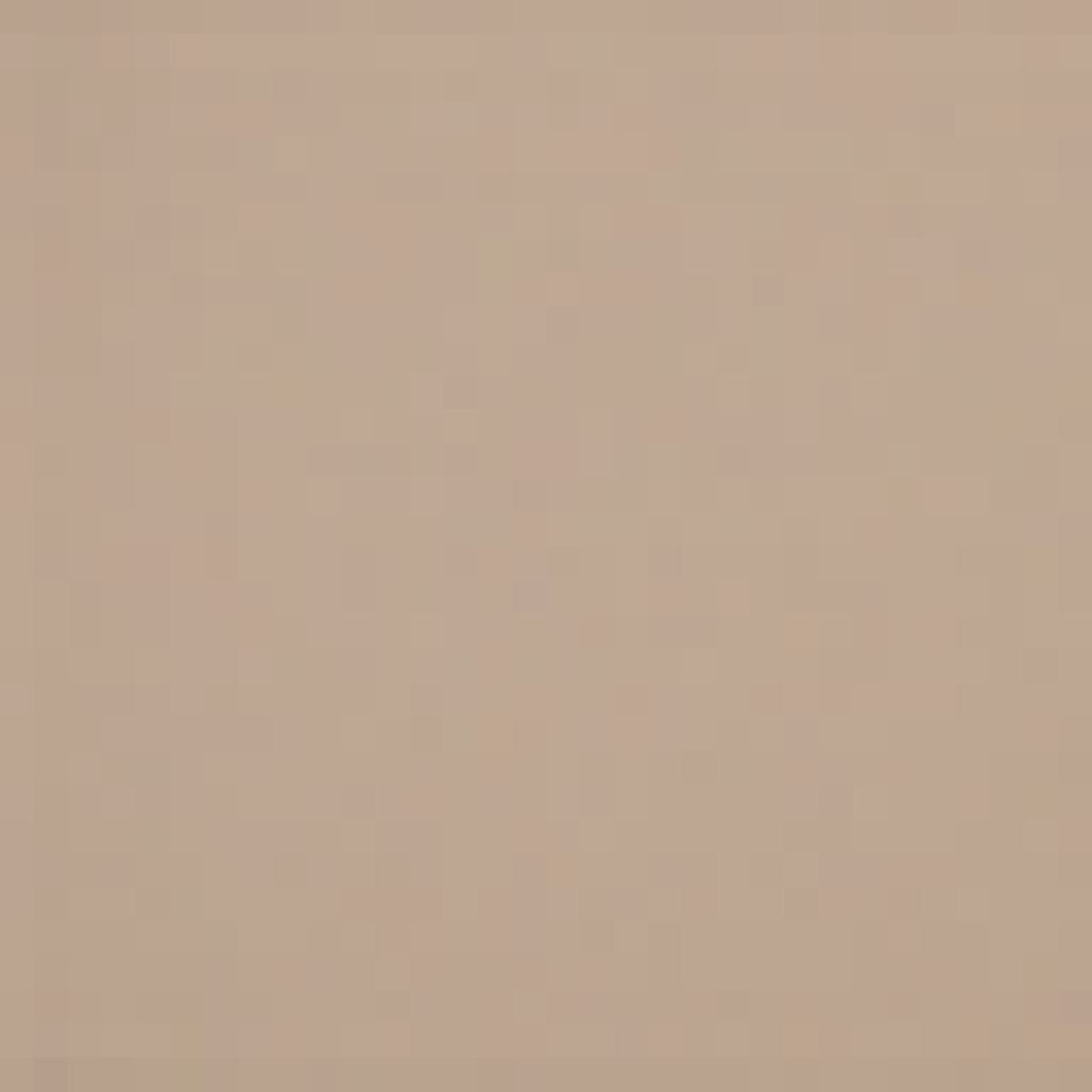} & 
   \includegraphics[width=\linewidth]{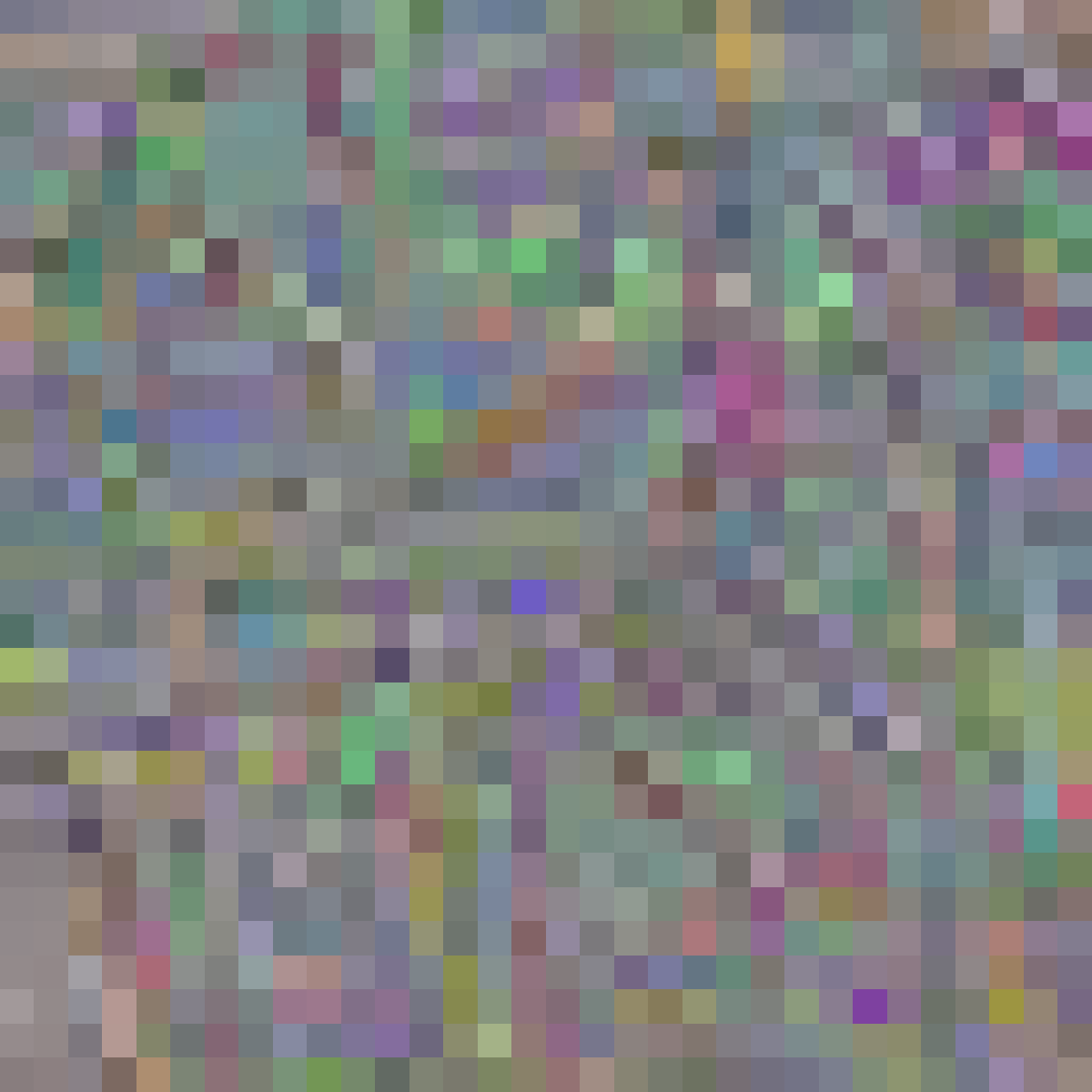} &
   \includegraphics[width=\linewidth]{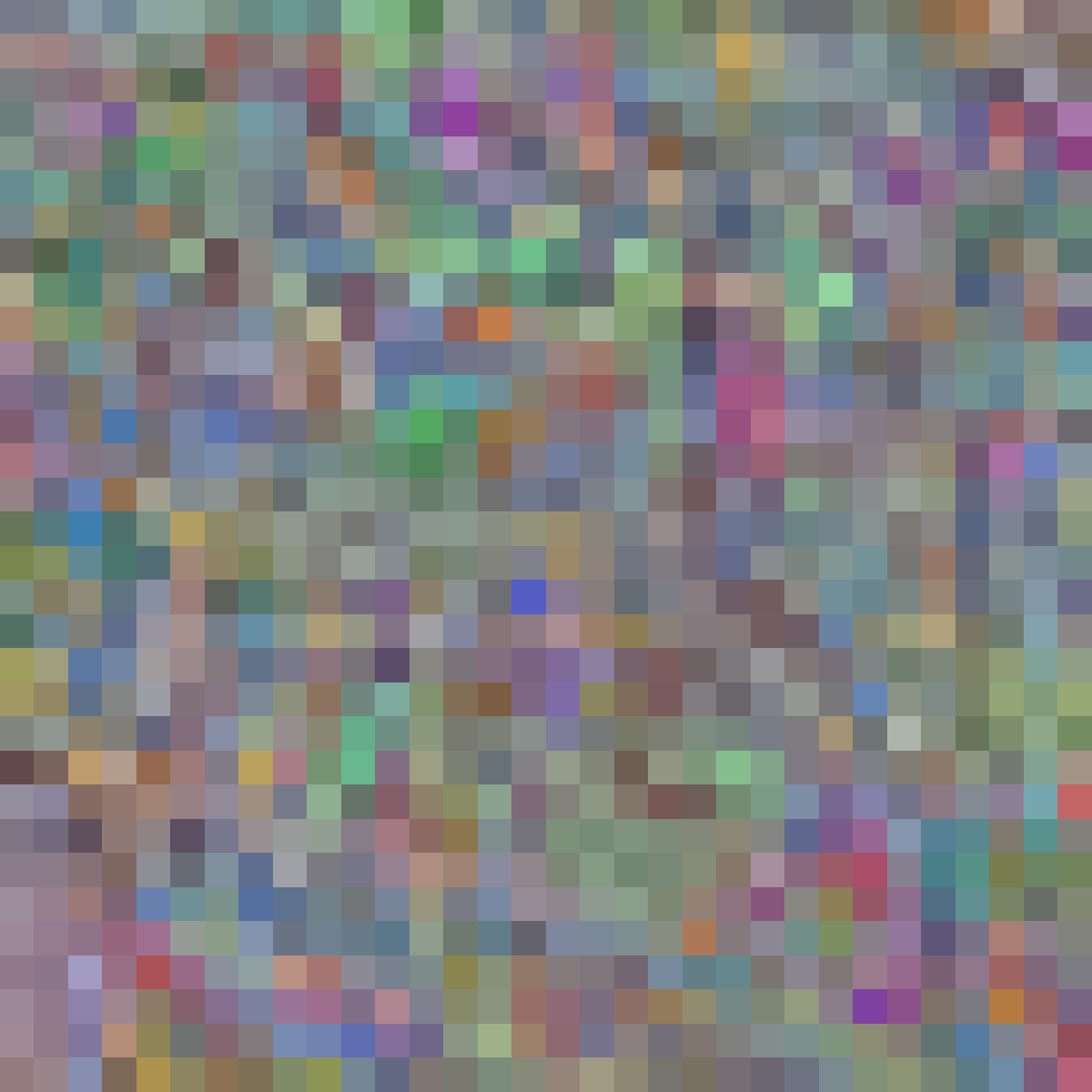} &
   \includegraphics[width=\linewidth]{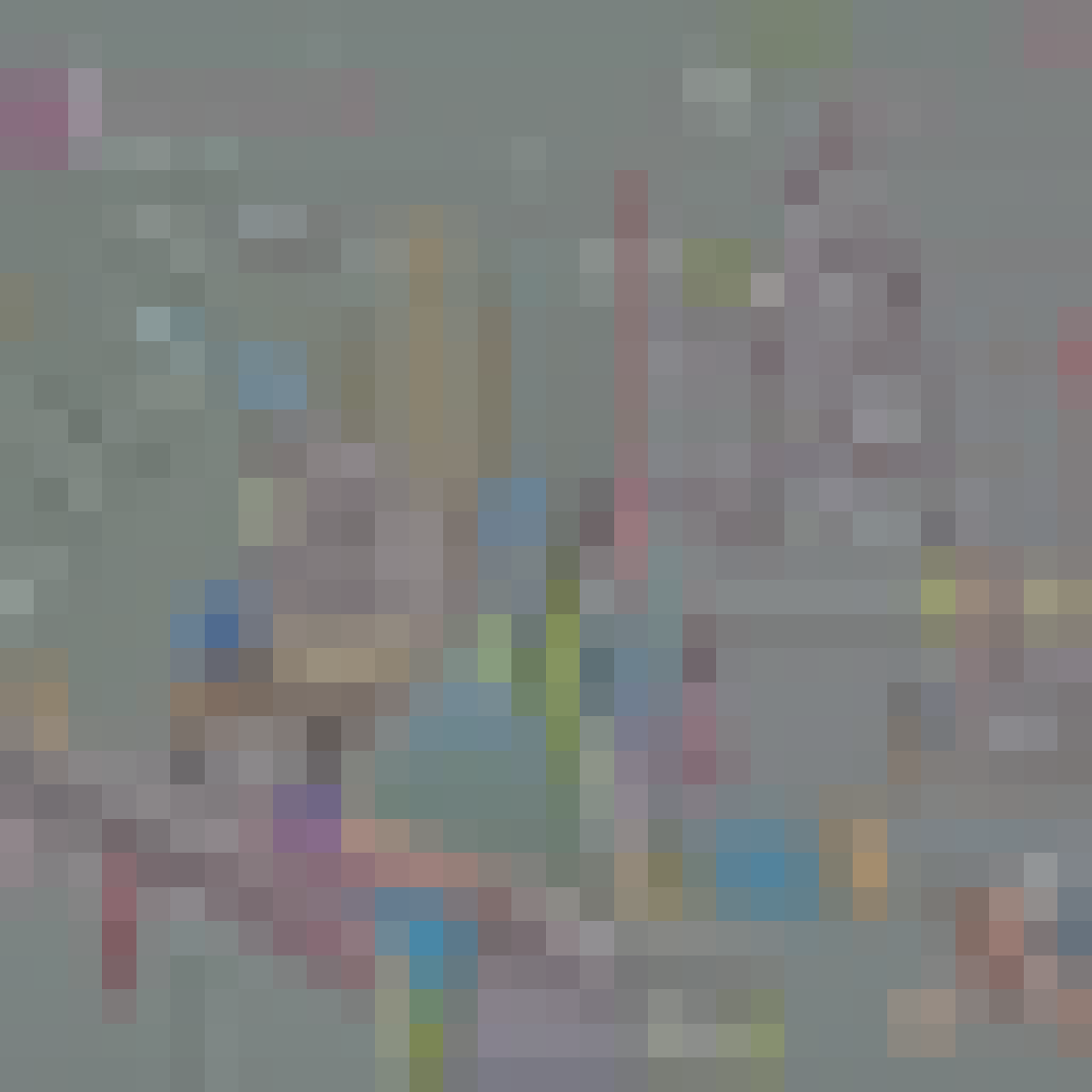} \\
  \end{tabularx}
  \caption{Multiple examples of attack outcomes on CIFAR-10 test samples, in the same setting as Figure~\ref{fig:end2end_attacks}.}%
  \label{fig:more_examples}
\end{figure}

Figure~\ref{fig:more_examples} shows eight additional outcomes in the same setting as Figure~\ref{fig:end2end_attacks}. In all cases, it can be qualitatively observed that:
\begin{itemize}
    \item The near sample attack results in a smoothed version of the benign sample, hence making it more compressible and decreasing complexity.
    \item The reverse integration results in very smooth, monochrome patterns that, in some cases (e.g., second and third row) retain some of the semantic structure of the original sample.
    \item The random region (both with and without the integral term) and the unrestricted attacks result in shape structure present in the sample. This is less visible for the random region attack due to its constraints.
\end{itemize}

\section{Black-box Attacks}
\begin{figure}[tp]
  \centering
  \setlength{\tabcolsep}{2pt}
  \renewcommand\tabularxcolumn[1]{m{#1}}
  \newcolumntype{R}{>{\raggedleft\arraybackslash}X}
  \newcolumntype{C}{>{\centering\arraybackslash}X}
  \begin{tabularx}{1\textwidth}{CCCCCCC}
      \small $\log p_\theta \textnormal{ = } \num{7.55}$  \break\
      $\mathcal{C} \textnormal{ = } \num{0.016}$
      & \small $\log p_\theta \textnormal{ = } \num{5.59}$  \break\
      $\mathcal{C} \textnormal{ = } \num{0.019}$
      & \small $\log p_\theta \textnormal{ = } \num{5.57}$  \break\
      $\mathcal{C} \textnormal{ = } \num{0.019}$
      & \small $\log p_\theta \textnormal{ = } \num{5.72}$  \break\
      $\mathcal{C} \textnormal{ = } \num{0.019}$
      &  \small $\log p_\theta \textnormal{ = } \num{5.96}$  \break\
      $\mathcal{C} \textnormal{ = } \num{0.019}$
      & \small $\log p_\theta \textnormal{ = } \num{6.09}$  \break\
      $\mathcal{C} \textnormal{ = } \num{0.019}$
      &  \small $\log p_\theta \textnormal{ = } \num{4.63}$  \break\
      $\mathcal{C} \textnormal{ = } \num{0.019}$ \\
      
   \includegraphics[width=\linewidth]{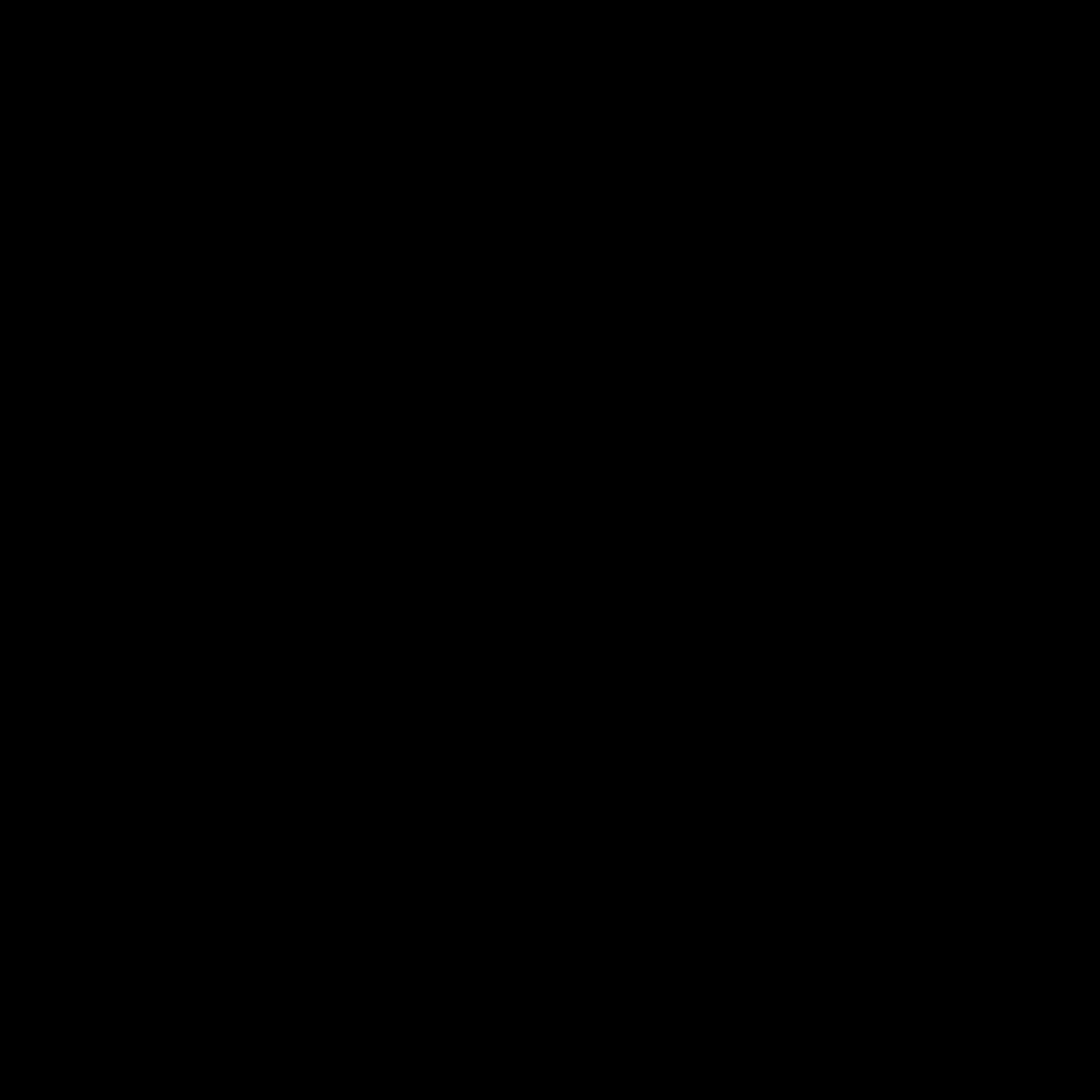} & 
   \includegraphics[width=\linewidth]{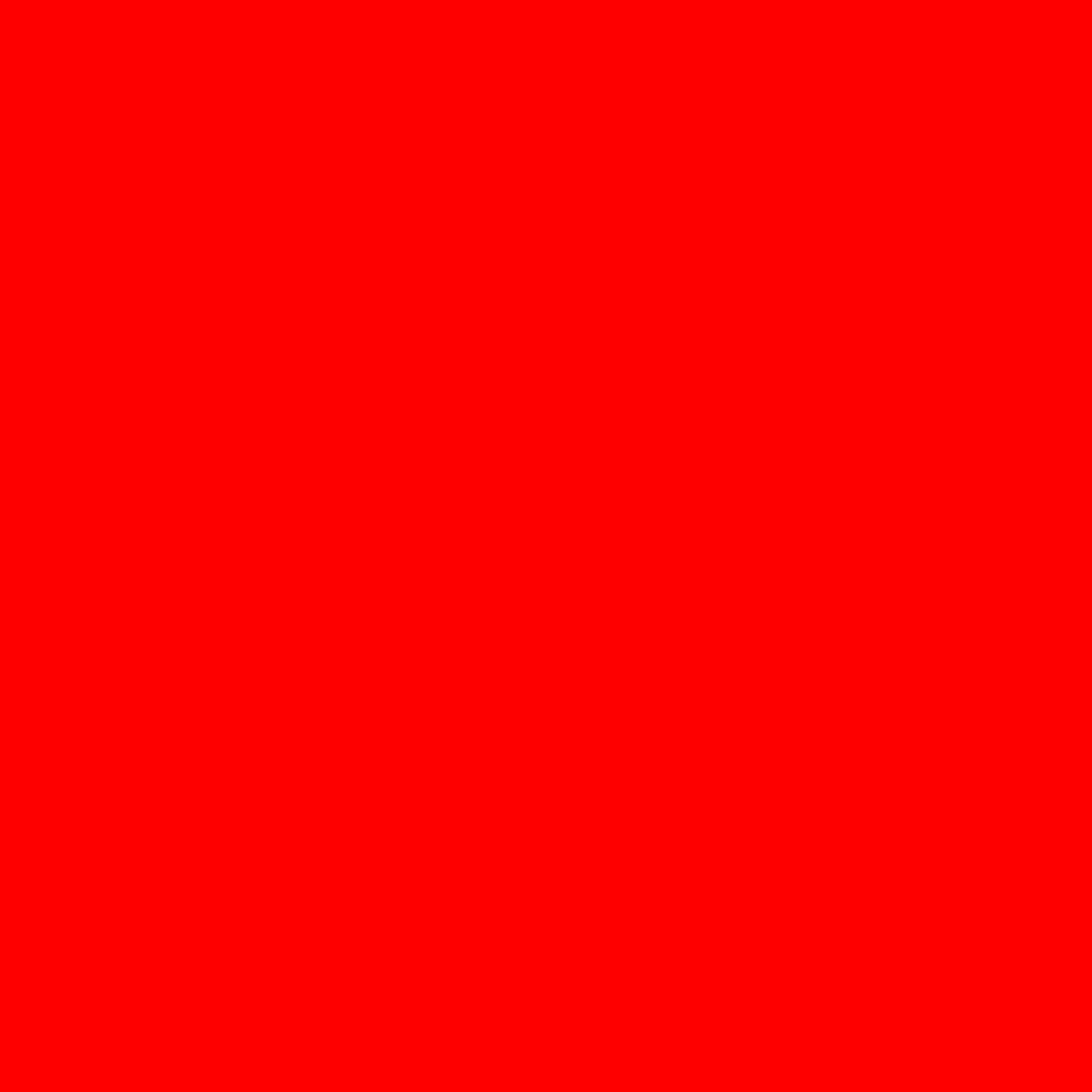} & 
   \includegraphics[width=\linewidth]{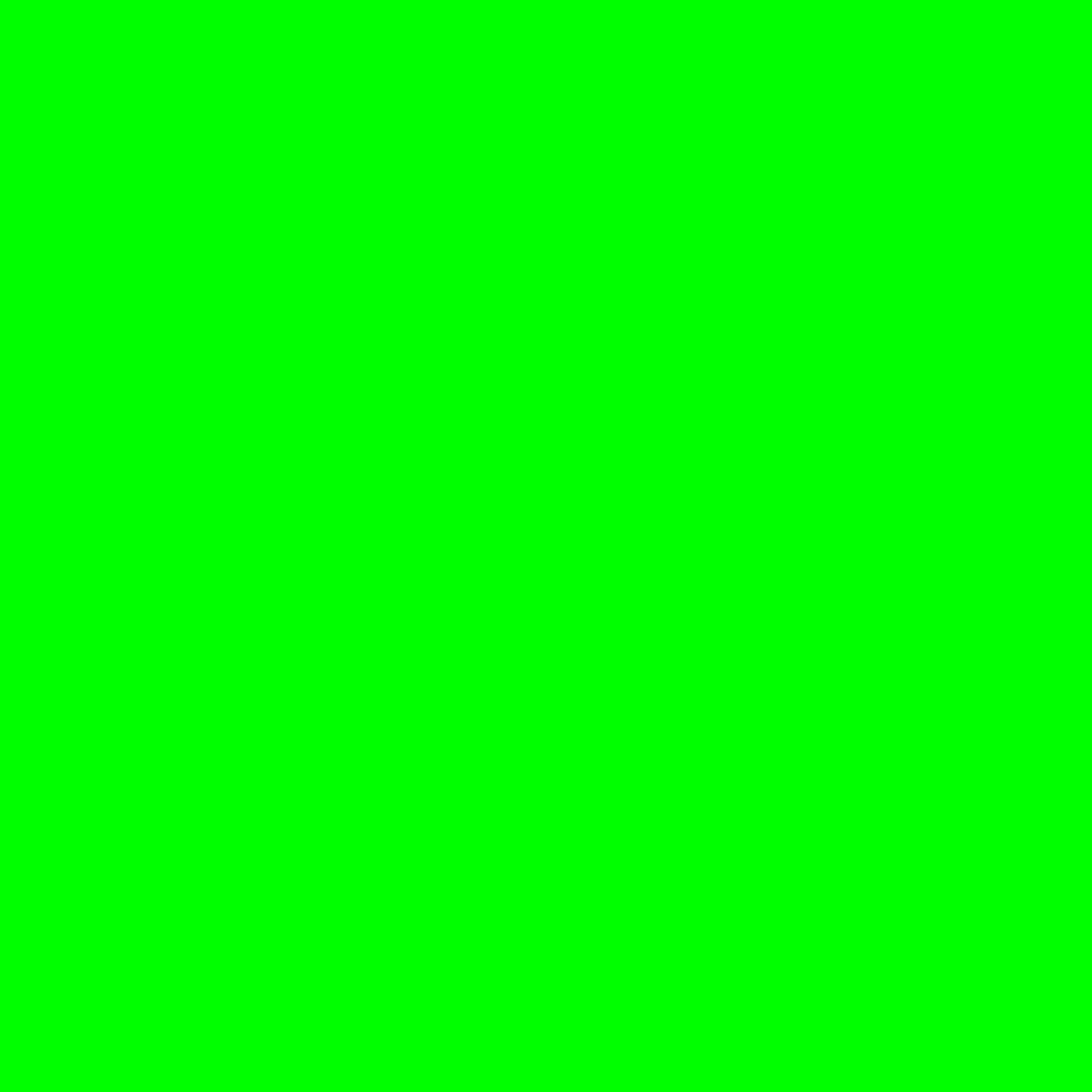} & 
   \includegraphics[width=\linewidth]{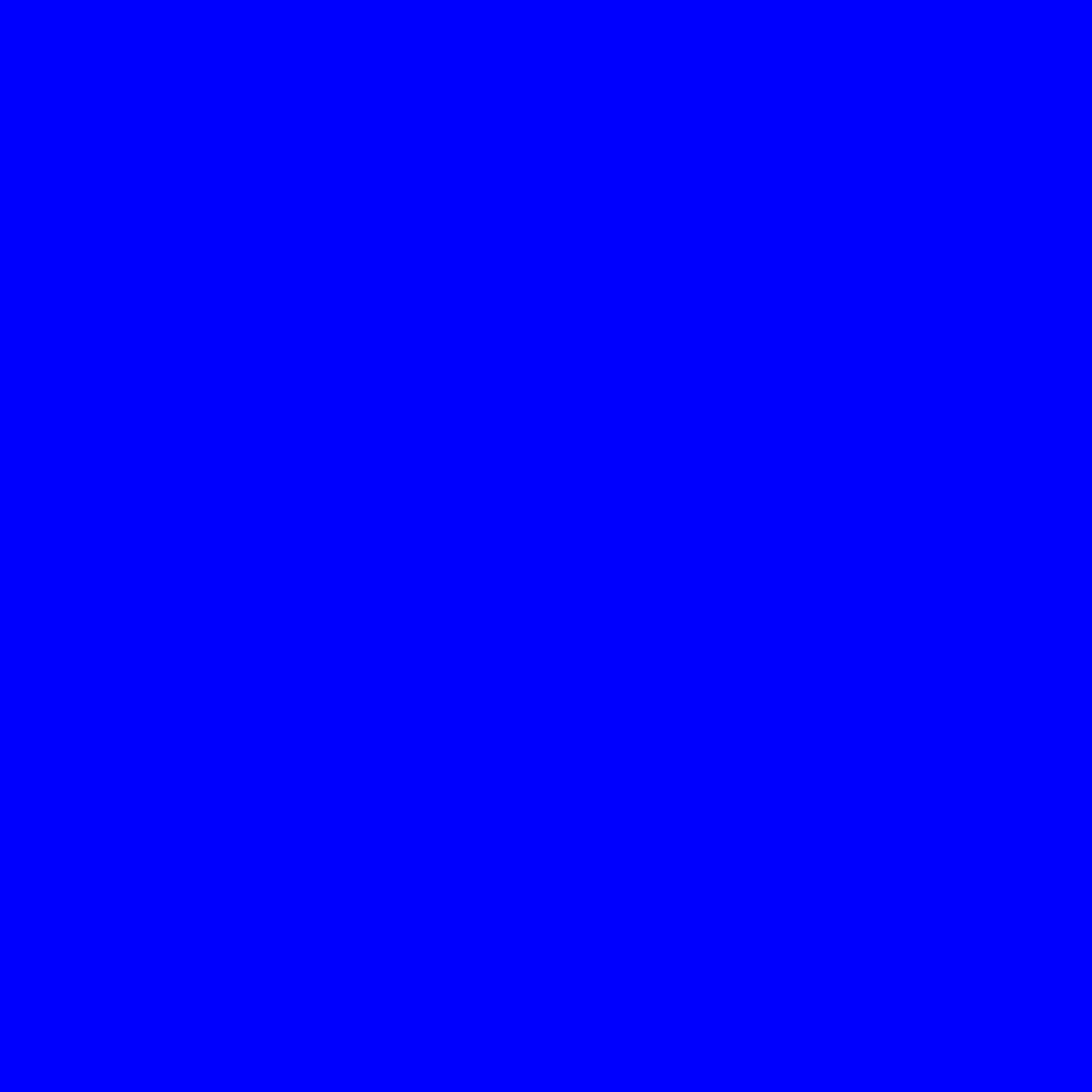} & 
   \includegraphics[width=\linewidth]{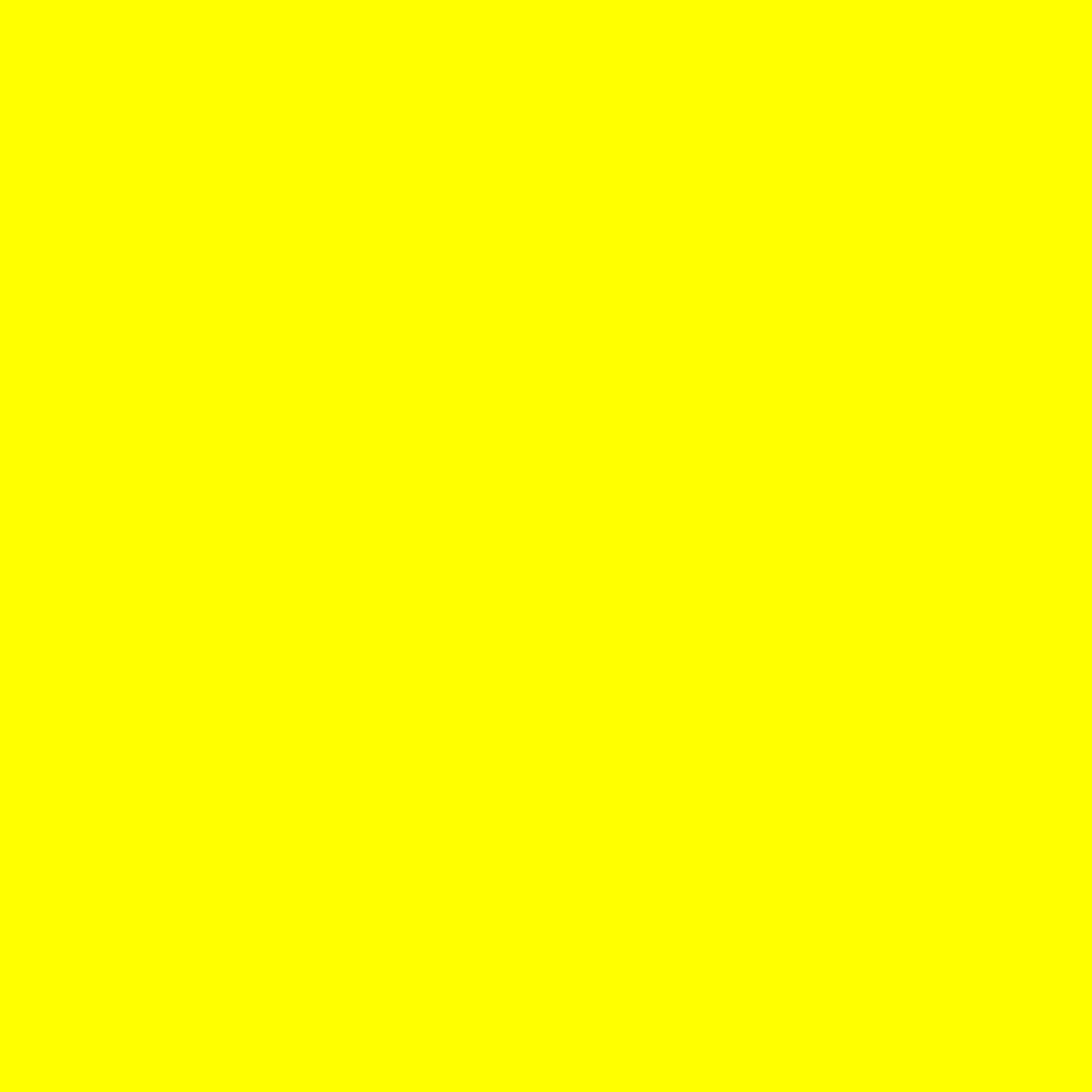} & 
   \includegraphics[width=\linewidth]{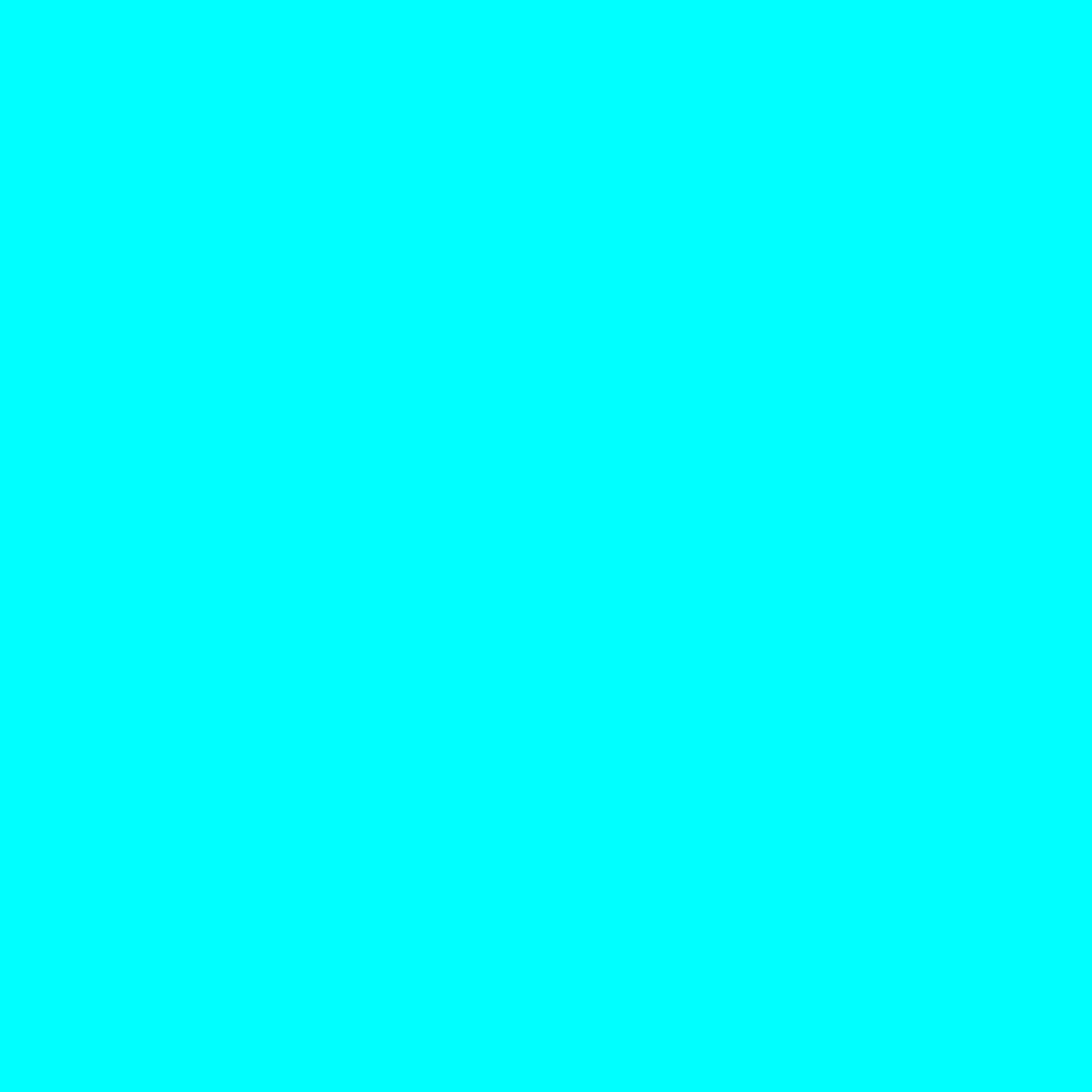} & 
   \includegraphics[width=\linewidth]{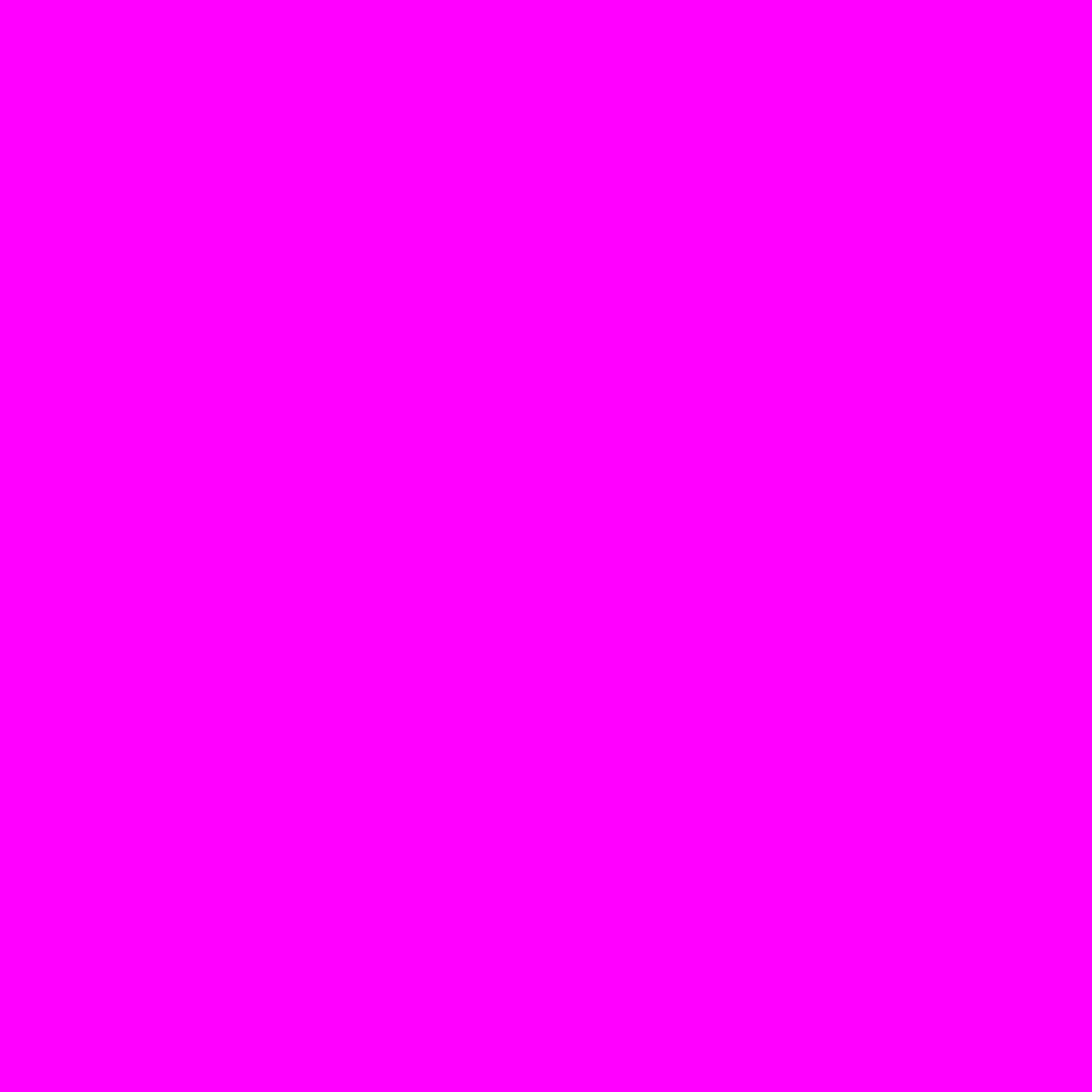} \\

    \small $\log p_\theta \textnormal{ = } \num{7.46}$  \break\
      $\mathcal{C} \textnormal{ = } \num{0.081}$
      & \small $\log p_\theta \textnormal{ = } \num{7.24}$  \break\
      $\mathcal{C} \textnormal{ = } \num{0.110}$
      & \small $\log p_\theta \textnormal{ = } \num{6.83}$  \break\
      $\mathcal{C} \textnormal{ = } \num{0.133}$
      & \small $\log p_\theta \textnormal{ = } \num{6.75}$  \break\
      $\mathcal{C} \textnormal{ = } \num{0.160}$
      &  \small $\log p_\theta \textnormal{ = } \num{6.96}$  \break\
      $\mathcal{C} \textnormal{ = } \num{0.137}$
      & \small $\log p_\theta \textnormal{ = } \num{6.67}$  \break\
      $\mathcal{C} \textnormal{ = } \num{0.163}$
      &  \small $\log p_\theta \textnormal{ = } \num{6.61}$  \break\
      $\mathcal{C} \textnormal{ = } \num{0.179}$ \\
      
   \includegraphics[width=\linewidth]{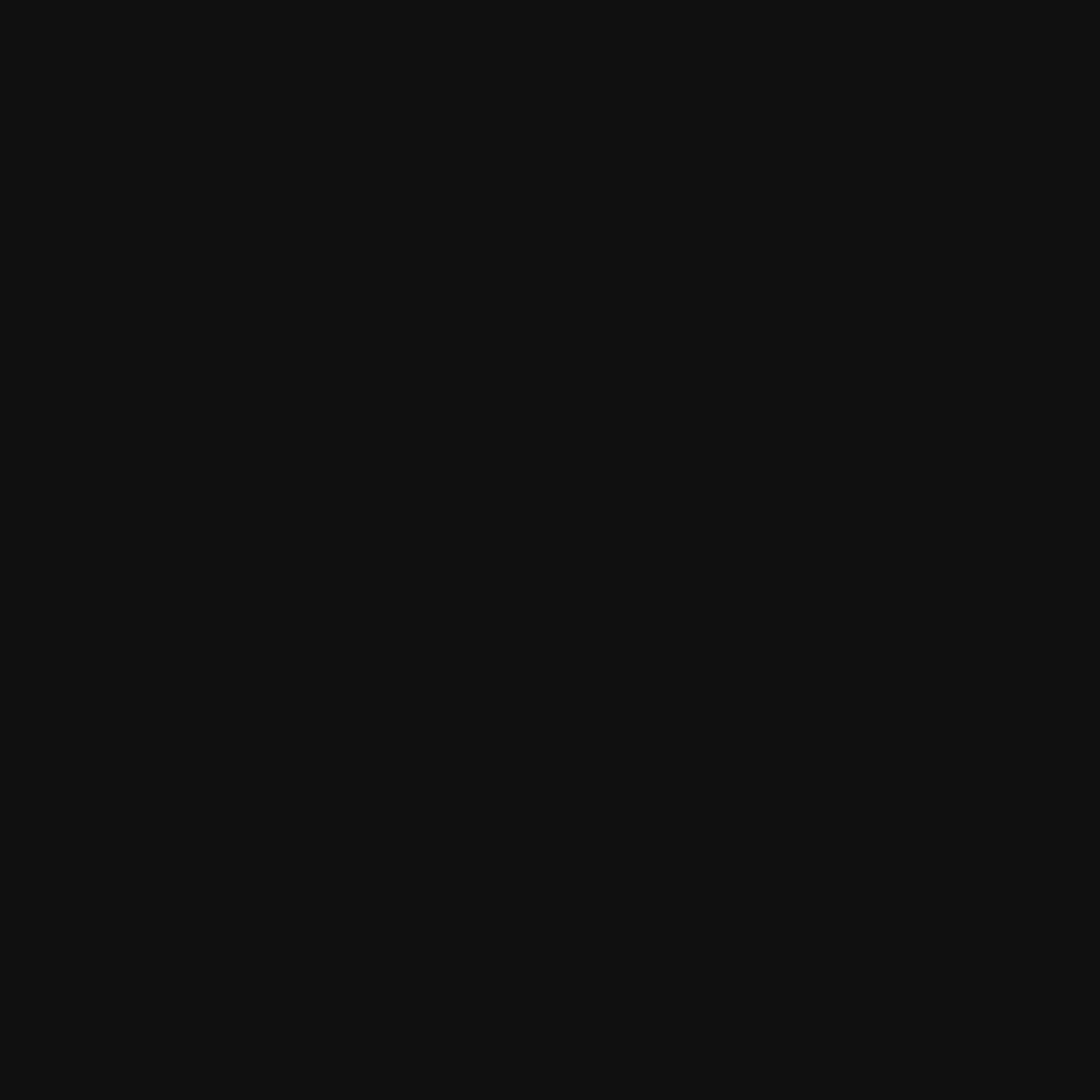} & 
   \includegraphics[width=\linewidth]{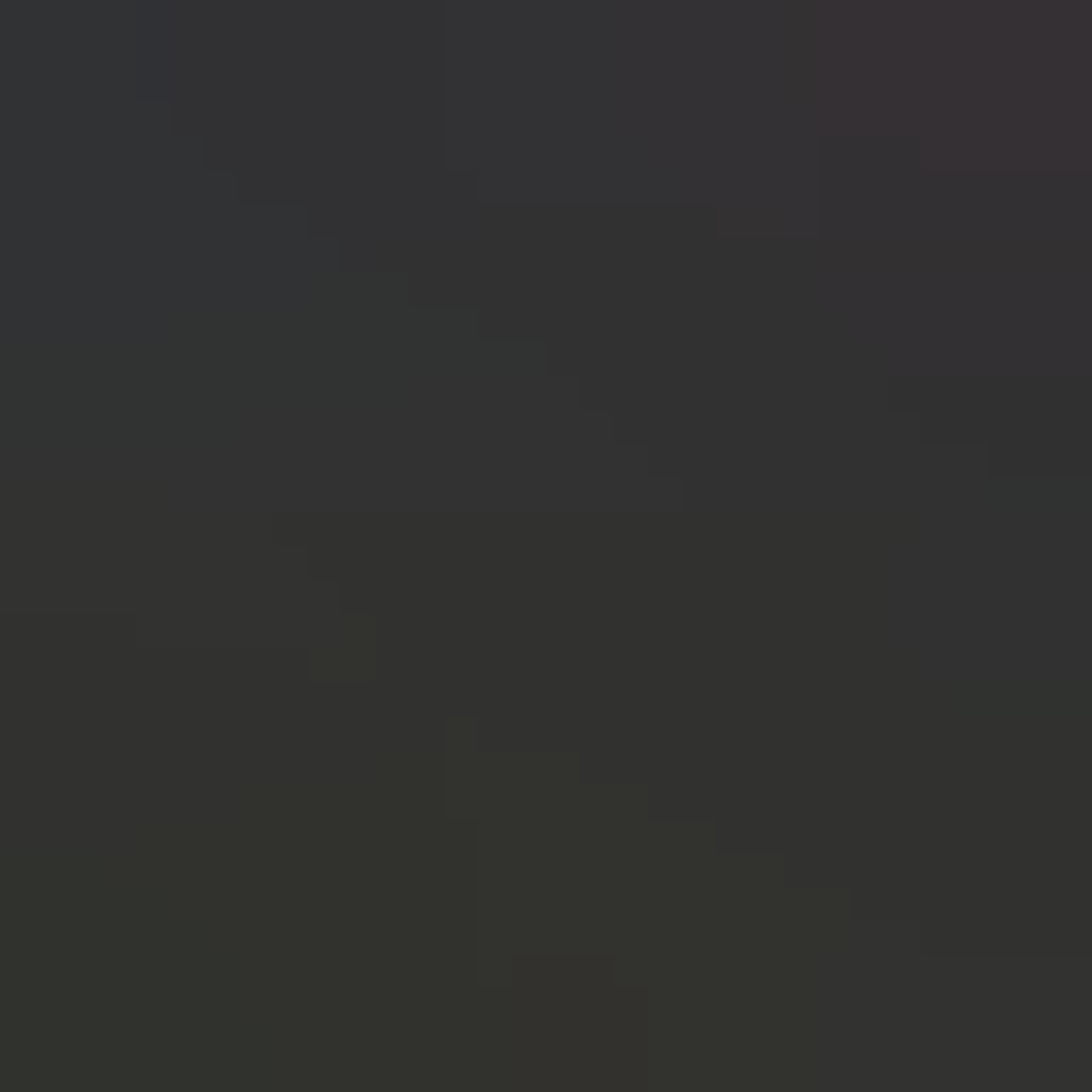} & 
   \includegraphics[width=\linewidth]{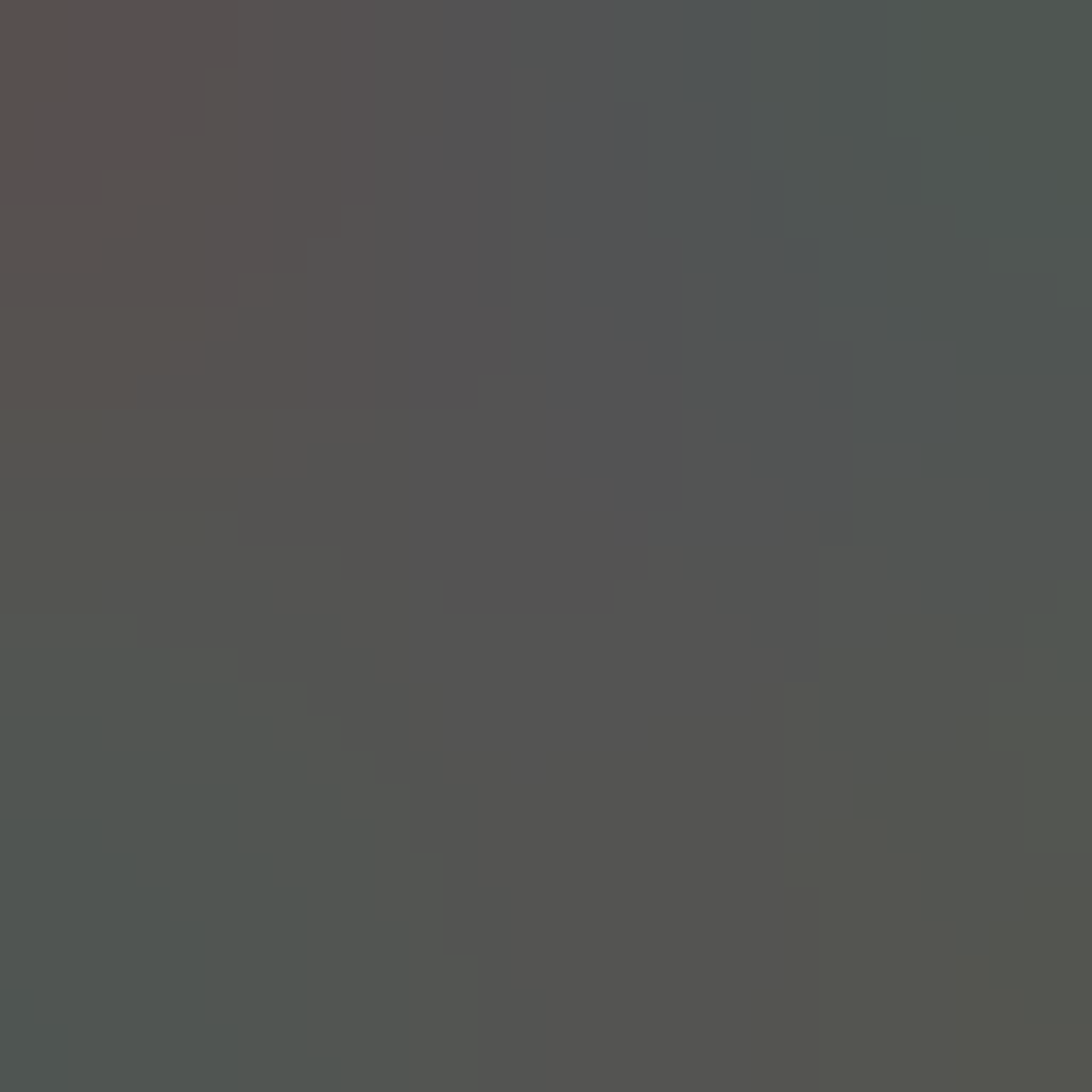} & 
   \includegraphics[width=\linewidth]{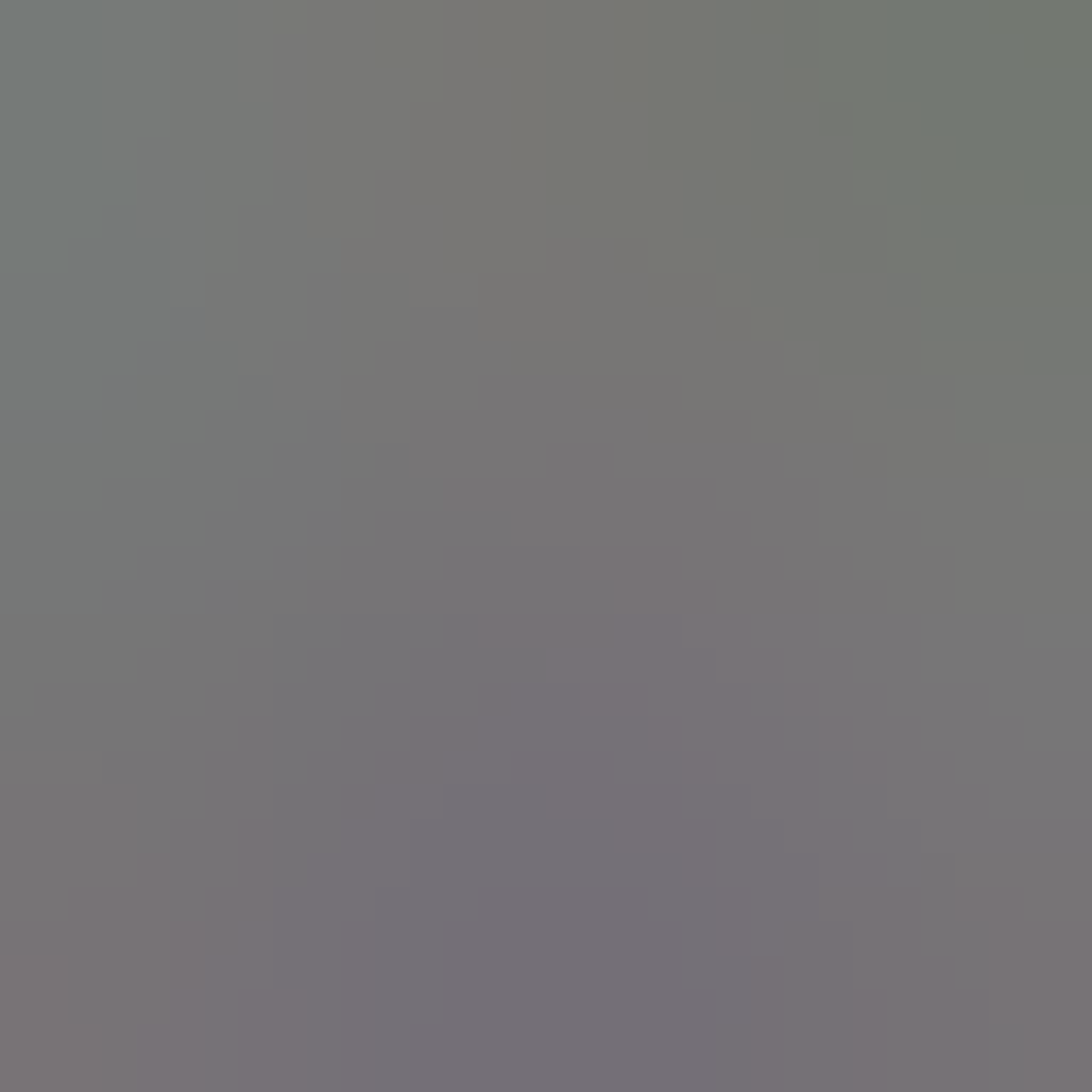} & 
   \includegraphics[width=\linewidth]{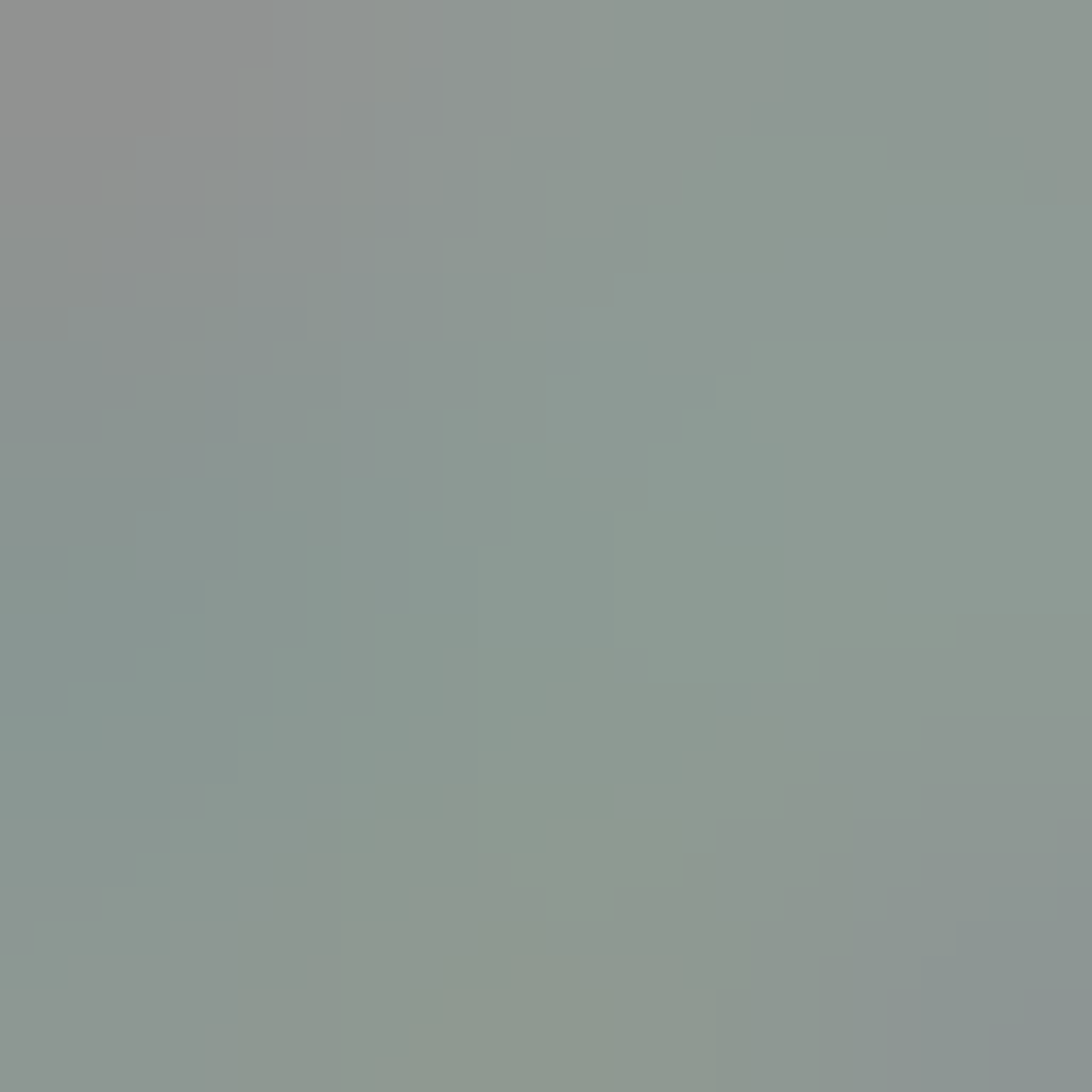} & 
   \includegraphics[width=\linewidth]{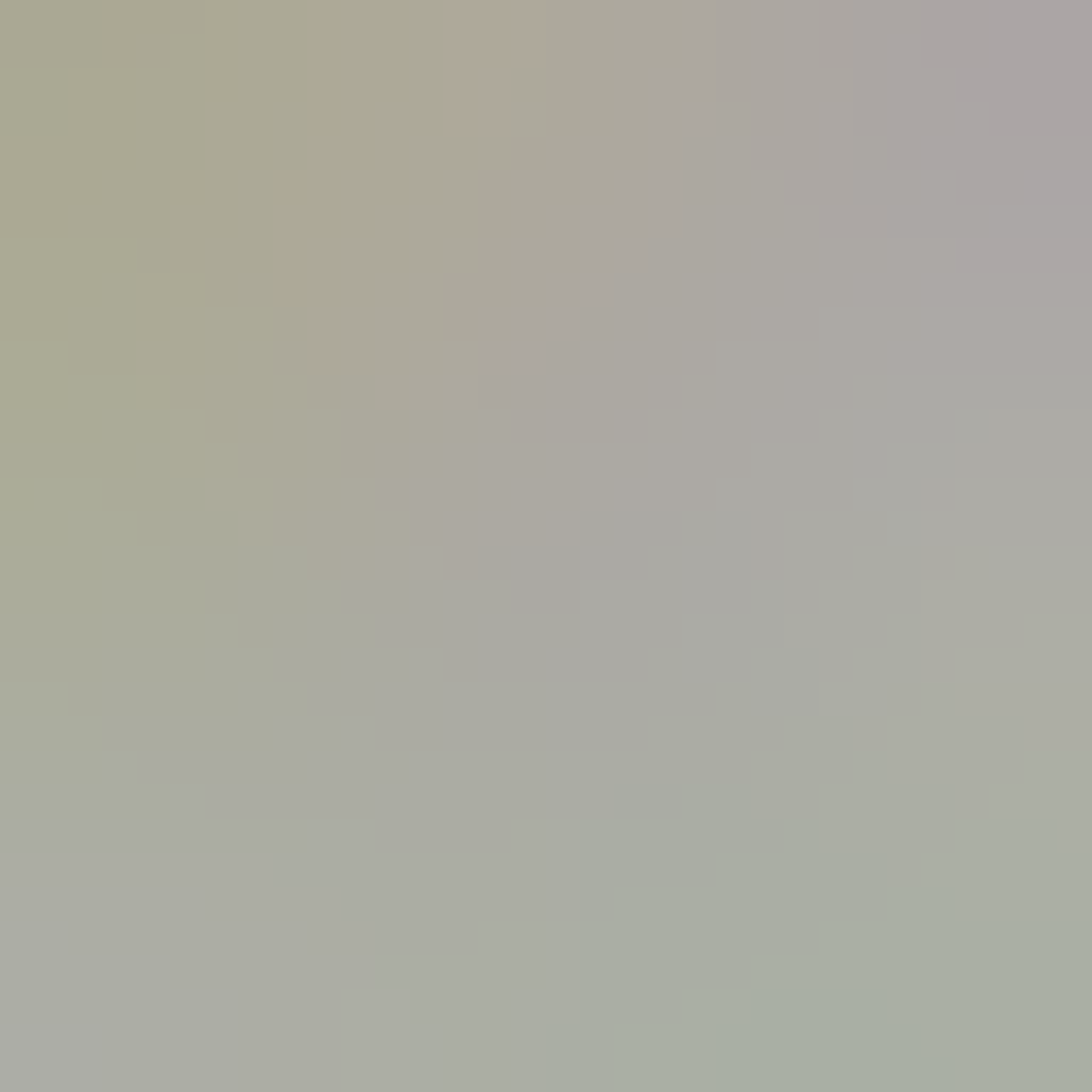} & 
   \includegraphics[width=\linewidth]{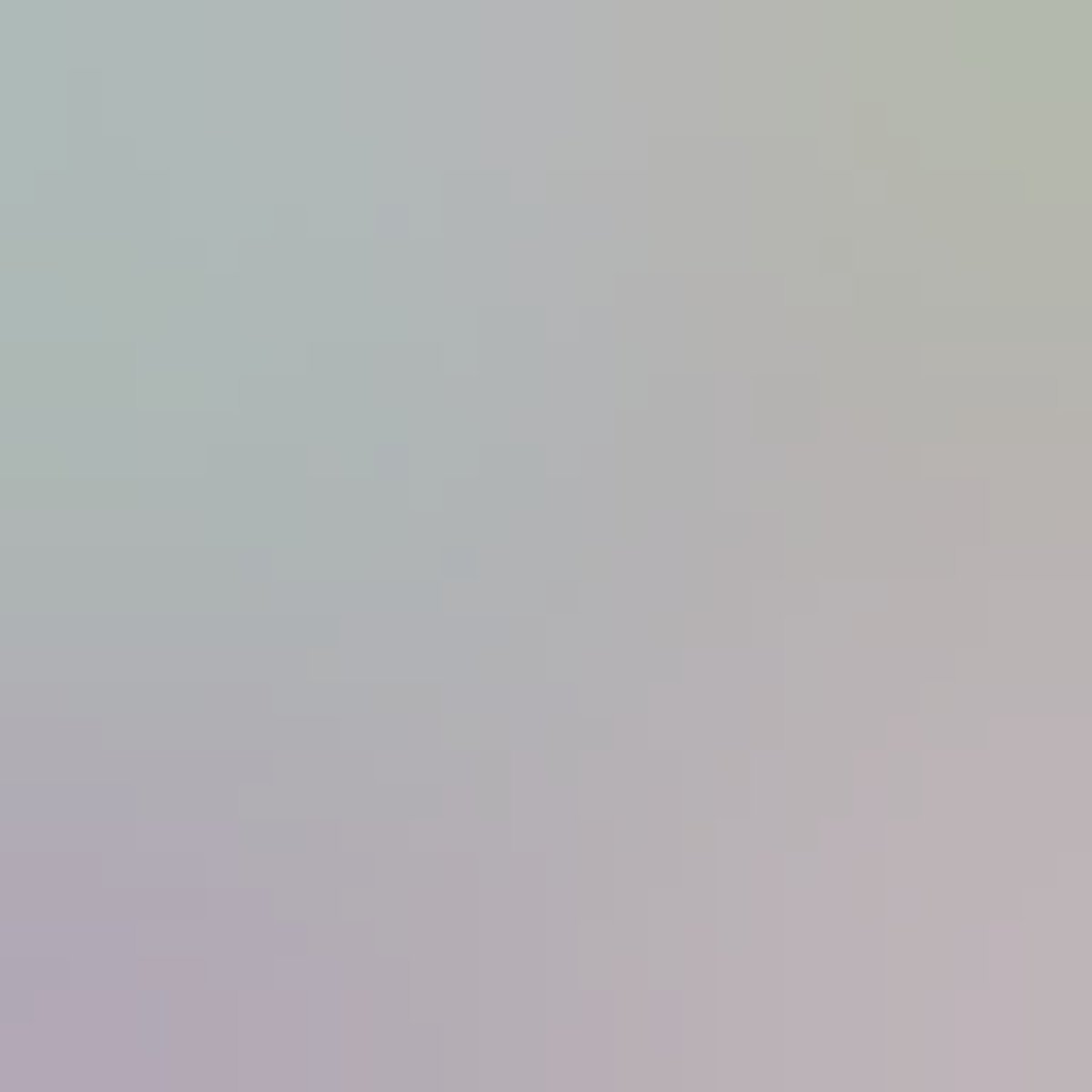} \\

   \small $\log p_\theta \textnormal{ = } \num{1.11}$  \break\
      $\mathcal{C} \textnormal{ = } \num{0.739}$
      & \small $\log p_\theta \textnormal{ = } \num{0.39}$ \break\
      $\mathcal{C} \textnormal{ = } \num{0.867}$
      & \small $\log p_\theta \textnormal{ = } \num{0.0}$ \break\
      $\mathcal{C} \textnormal{ = } \num{0.949}$
      & \small $\log p_\theta \textnormal{ = -} \num{0.4}$ \break\
      $\mathcal{C} \textnormal{ = } \num{1.004}$
      &  \small $\log p_\theta \textnormal{ = -} \num{0.7}$ \break\
      $\mathcal{C} \textnormal{ = } \num{1.023}$
      & \small $\log p_\theta \textnormal{ = -} \num{0.9}$ \break\
      $\mathcal{C} \textnormal{ = } \num{1.023}$
      &  \small $\log p_\theta \textnormal{ = -} \num{1.1}$ \break\
      $\mathcal{C} \textnormal{ = } \num{1.023}$ \\
      
   \includegraphics[width=\linewidth]{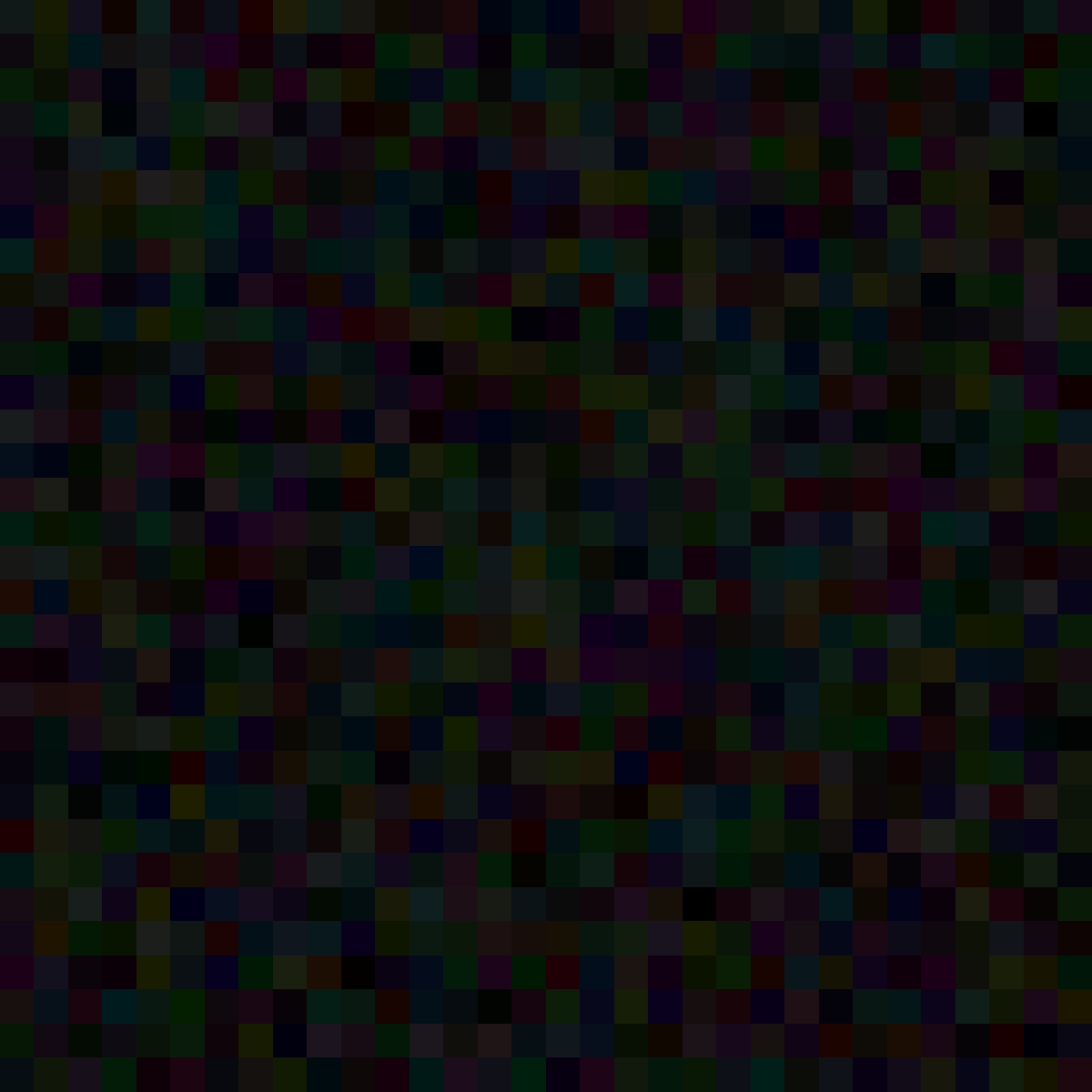} & 
   \includegraphics[width=\linewidth]{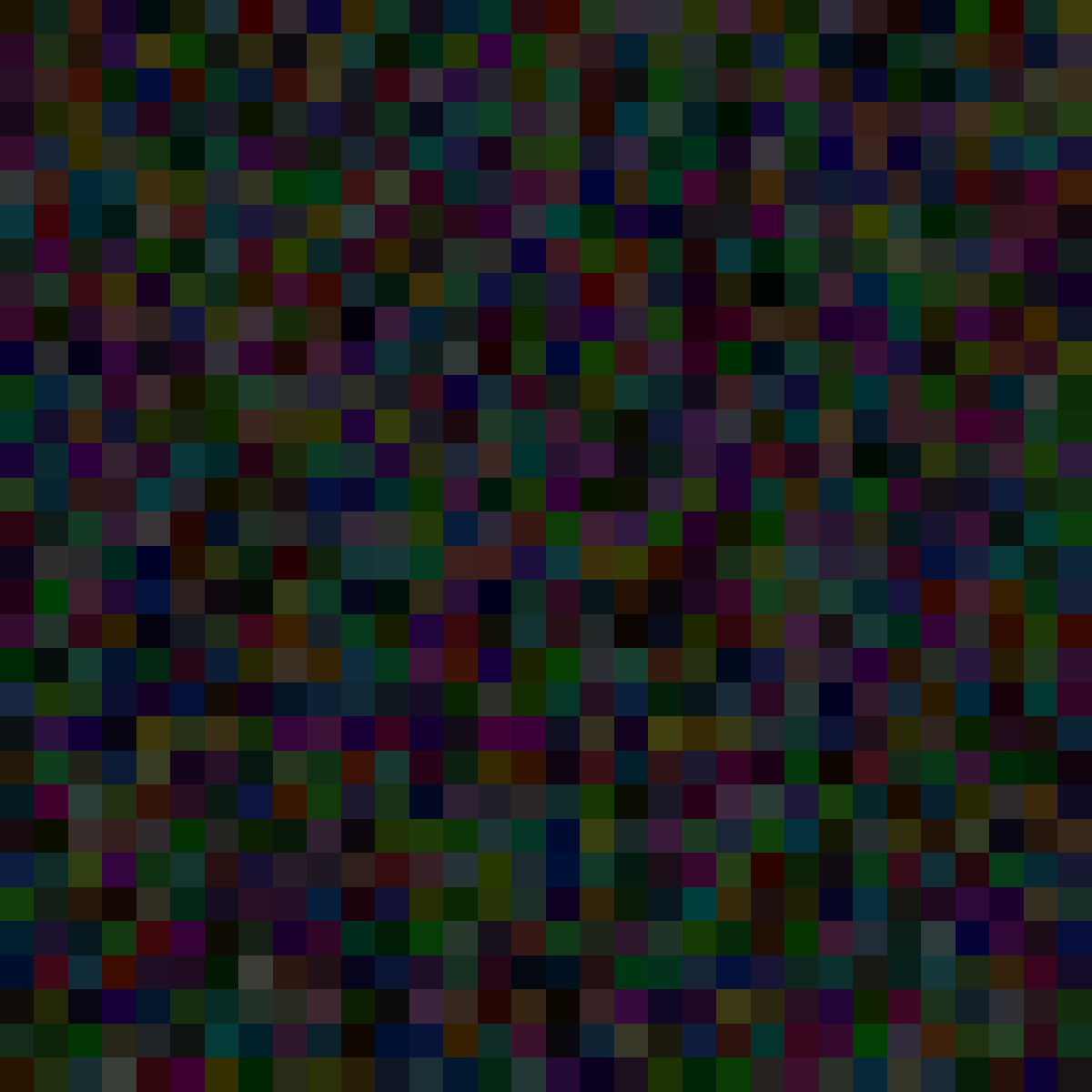} & 
   \includegraphics[width=\linewidth]{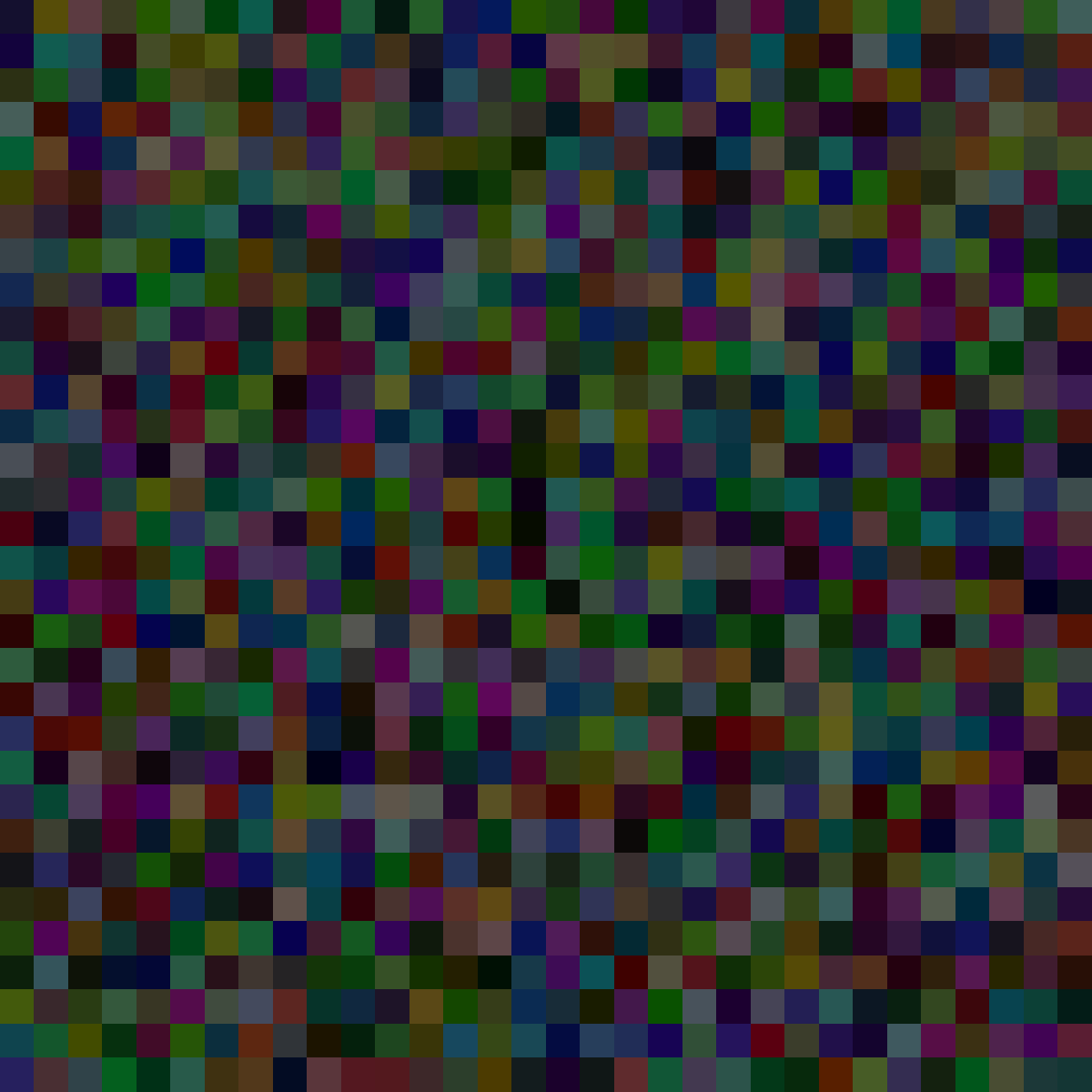} & 
   \includegraphics[width=\linewidth]{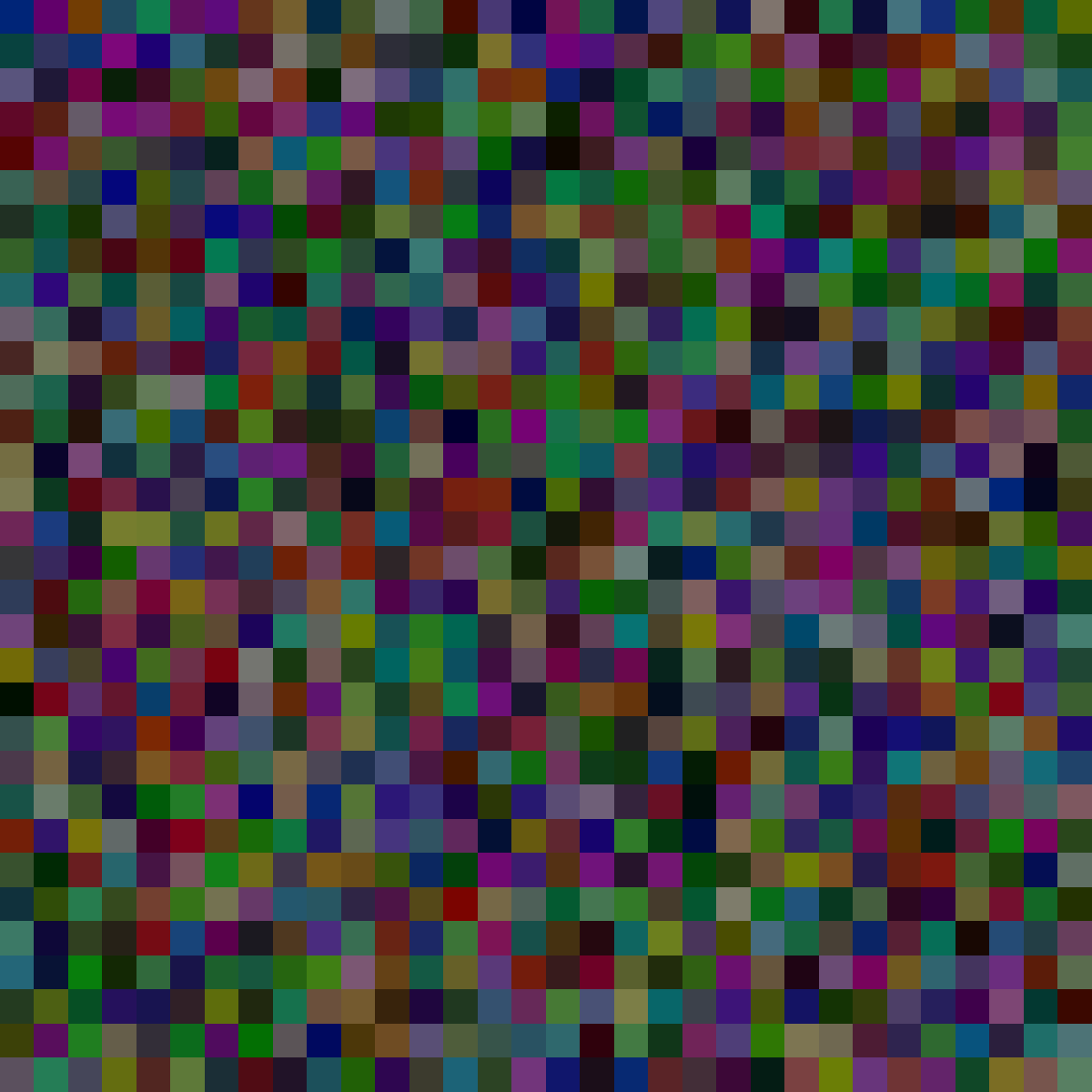} & 
   \includegraphics[width=\linewidth]{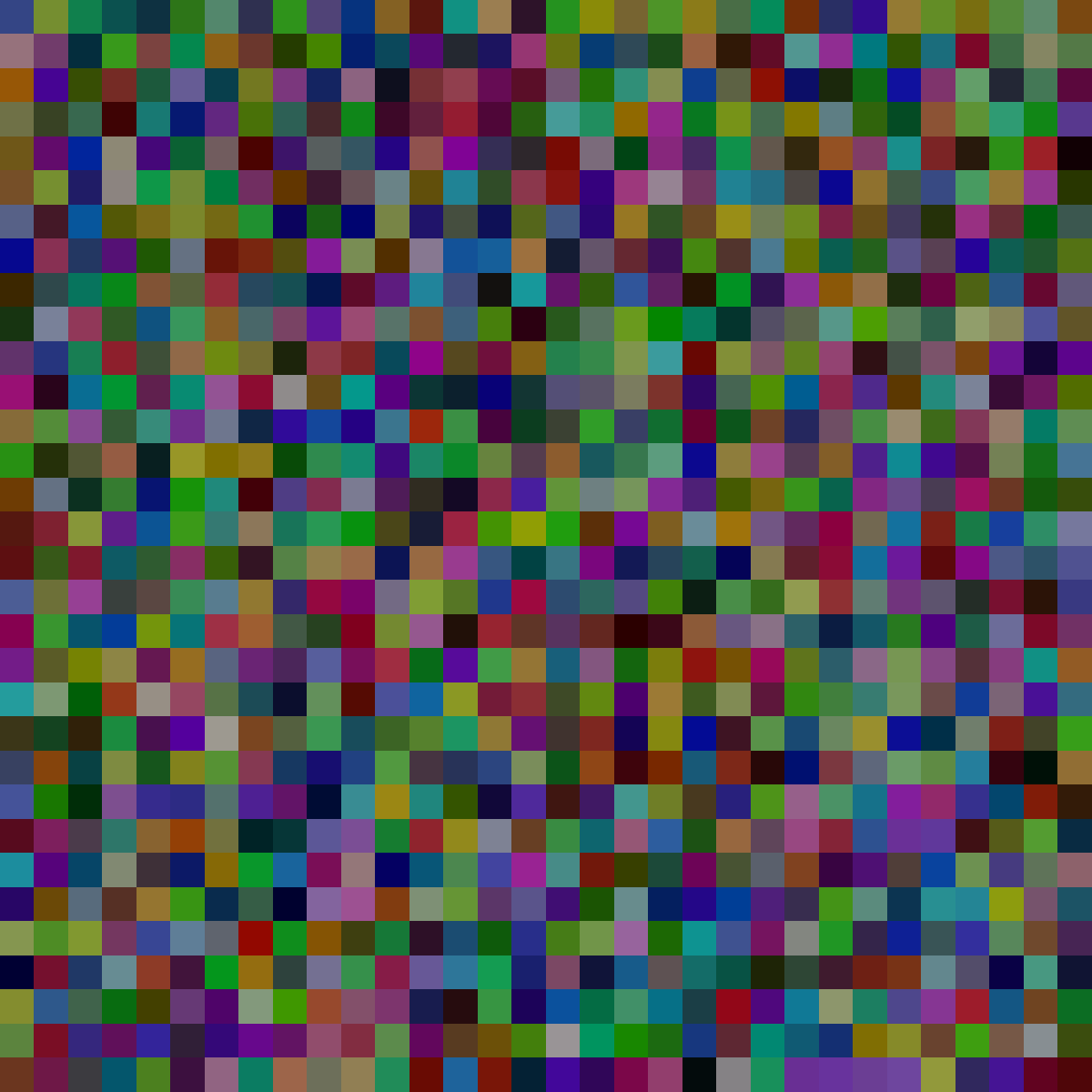} & 
   \includegraphics[width=\linewidth]{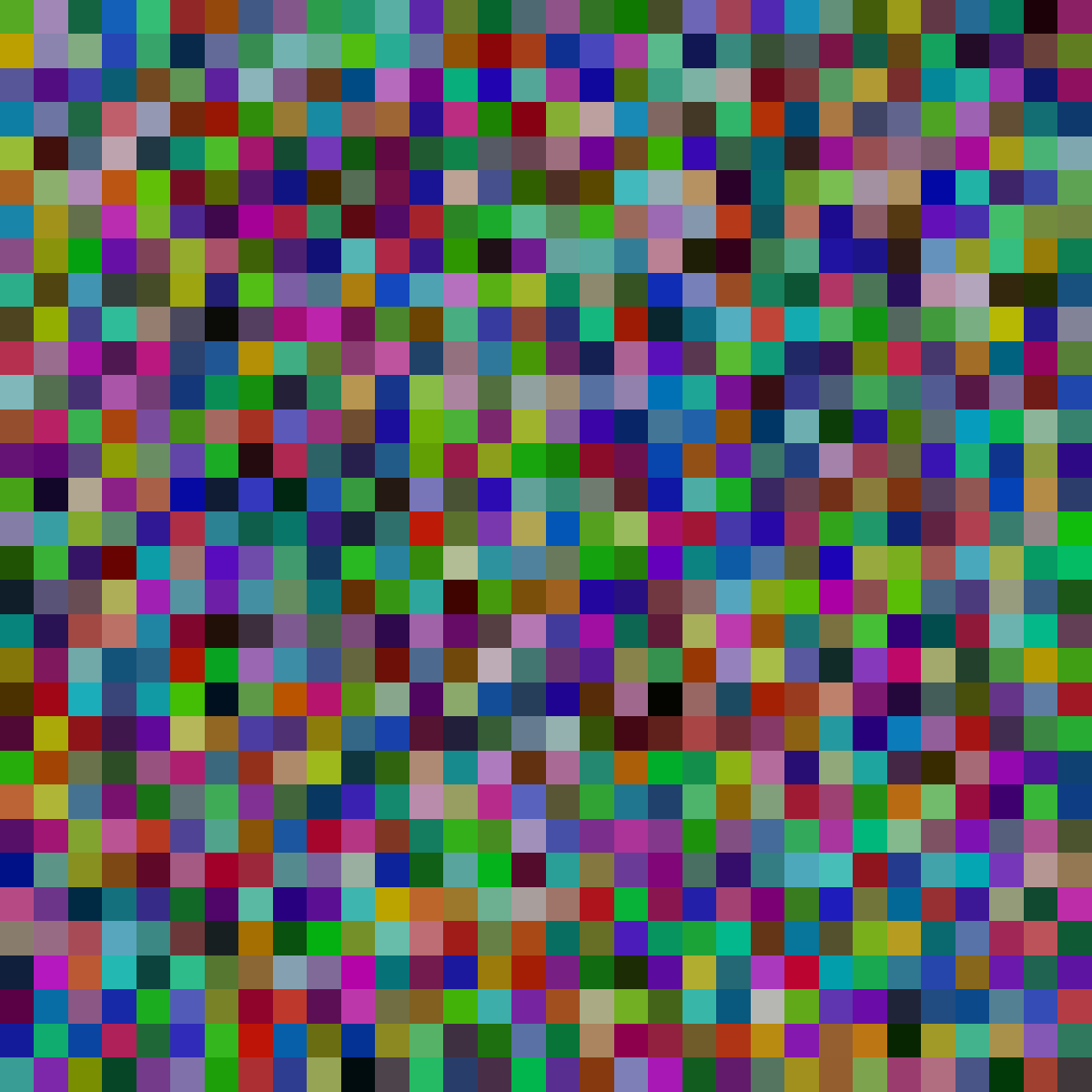} & 
   \includegraphics[width=\linewidth]{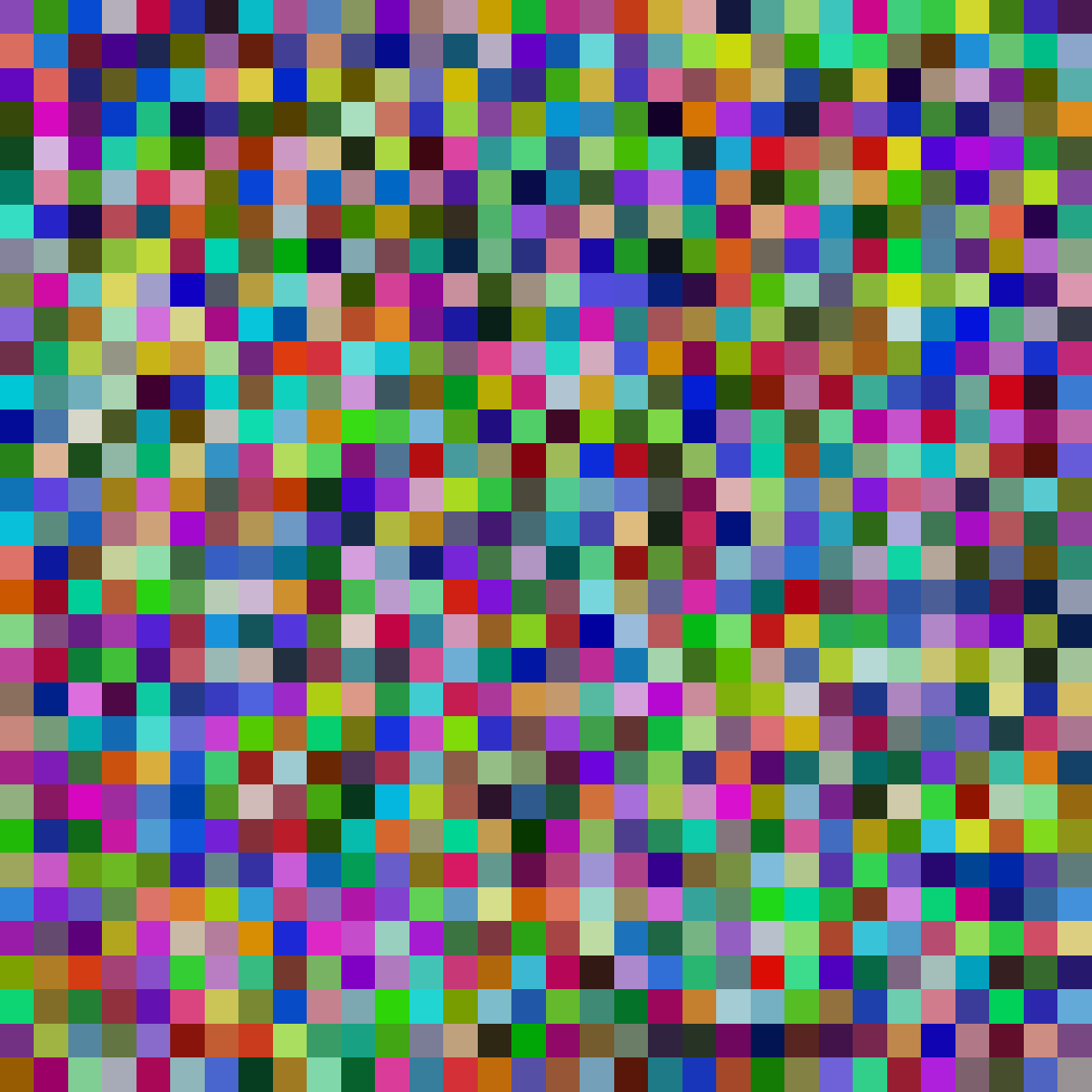} \\
   
  \end{tabularx}
  \caption{Outcomes of evaluating $\ode$ for samples with black-box (optimization free) designed high and low complexities. First row: monochrome images. Second row: Filtered uniform noise of increasing magnitude. Third row: Uniform noise of increasing magnitude.}
  \label{fig:bbox_attacks}
\end{figure}

We also perform a black-box verification of the hypothesis that $\ode$ is biased towards estimating high values for samples that have low complexity by designing a simple set of such samples. Figure~\ref{fig:bbox_attacks} shows the outcome for evaluating $\ode$ using the accurate solver for three types of samples:
\begin{itemize}
    \item Monochrome samples. These samples hit the image space boundary on at least one coordinate, and as expected, have very low complexities.
    \item Smooth perturbations of increasing magnitude applied to the sample with all values set to $-1$. We generate these perturbations by sampling uniform noise and filtering it with a two-dimensional Gaussian filter with kernel size $8$.
    \item Random uniform noise samples of increasing standard deviation. Given that these have high complexity, we expect their estimated values of $\ode$ to be very low. Note that values of $\mathcal{C}$ slightly larger than $1$ can occur due to the sub-optimality of the $\texttt{PNG}$ implementation.
\end{itemize}

Figure~\ref{fig:bbox_attacks} shows that the expected negative correlation between $\ode$ and $\mathcal{C}$ holds in this case as well: monochrome images have very high values of the estimated density, and smooth perturbations of the all-black image remain "on-manifold", as predicted by the model. Contrary, random noise images have extremely low estimated values of $\ode$ and can be easily detected when compared with the values of $\ode$ for benign, in-distribution samples.

\end{appendices}

\end{document}